%% file: colm2026_conference.tex
\documentclass{article} 
\usepackage[final]{colm2026_conference}

\usepackage{microtype}
\usepackage{hyperref}
\usepackage{url}
\usepackage{graphicx}
\usepackage{xcolor}
\usepackage{float}
\usepackage{subcaption}
\usepackage{multirow}
\usepackage{booktabs}   
\usepackage{longtable}  
\usepackage{bbm}
\usepackage{tabularx}
\usepackage{amsmath}

\usepackage{array}      
\usepackage{pdflscape}  

\usepackage[pass]{geometry}

\usepackage{lineno}

\usepackage{xurl}

\definecolor{darkblue}{rgb}{0, 0, 0.5}
\hypersetup{colorlinks=true, citecolor=darkblue, linkcolor=darkblue, urlcolor=darkblue}

\title{Do Language Models Consistently Encode the Current Year?}

\author{%
    Suze van Adrichem\thanks{%
        \hangindent=1.8em
        \hangafter=1
    Equal contribution\quad
    Correspondence to \href{mailto:suzeva@cs.stanford.edu}{suzeva@cs.stanford.edu}
    and
    \href{mailto:aditijb@cs.stanford.edu}{aditijb@cs.stanford.edu}. \\
    Code available at
    \url{https://github.com/Suzehva/current-year-in-lms}}\hspace{0.35em},
    Aditi Bhaskar\footnotemark[1]\hspace{0.35em},
    Diyi Yang,
    Christopher Potts, 
    Jing Huang \\
Department of Computer Science, Stanford University%
}

\begin{document}

\ifcolmsubmission
\linenumbers
\fi

\maketitle

\begin{abstract} 

A consistent concept of the current time is important for temporal reasoning, yet how language models represent the current time is not well understood. We contribute two tasks that probe the current year in conceptually distinct ways: an \textit{associative} task, which infers the current year from verb tense, and a \textit{declarative} task, which directly queries for the current year. Both tasks estimate current years within one year of the post-training data cutoff of instruction-tuned language models. 
For base models, predictions on the associative task serve as a strong proxy for the pre-training data cutoff, with an average error of only 10 months across 13 models. However, their internal mechanisms diverge: the associative task uses mechanisms similar to factual recall, while the declarative task lacks consistent causal pathways. This divergence poses a challenge for updating the current year in language models. 
None of prompting, SFT, or weight editing succeed in shifting the associative and declarative years simultaneously. Prompting updates the declarative year (94.6\% success across 351 target years) but leaves the associative year nearly unchanged (1.7\% success). Year-shifted SFT also fails to shift the associative year, matching the target year in only one of eight models. 
Weight editing, while effective for both tasks individually, does not generalize across both.
Overall, our results show that the current year is not consistently encoded in language models: The associative notion, deeply ingrained in linguistic structures learned in pre-training, uses different causal mechanisms and resists the same modifications that easily shift the declarative notion learned in post-training.
\end{abstract}

\input{introduction}
\input{related_work}
\input{associative}
\input{declarative}
\input{in_context}
\input{conclusion}

\section*{Acknowledgments}

This work is supported in part by a grant from Coefficient Giving.


\begingroup
\sloppy
\bibliography{colm2026_conference}
\bibliographystyle{colm2026_conference}
\endgroup

\appendix
\input{appendix}
\end{document}

%% file: introduction.tex
\section{Introduction}
Language models (LMs) need to be aware of the current time to reason about temporal context. This is critical because many facts are time-dependent \citep{park2025chroknowledgeunveilingchronologicalknowledge, dhingra-etal-2022-time}, e.g., who the president is.
A flawed notion of the current time can lead to harmful behaviors such as over-refusal due to misidentifying data cutoff date~\citep{willison2025_xpost, karpathy2025_xpost} or allowing adversaries to induce backdoor behaviors by manipulating the current year~\citep{hubinger2024sleeperagentstrainingdeceptive, betley2025weirdgeneralizationinductivebackdoors}. 

Existing work on temporal awareness has focused on two directions: (1) behavioral evaluation of temporal reasoning abilities~\citep{park2025chroknowledgeunveilingchronologicalknowledge, jang-etal-2022-temporalwiki, fatemi2024testtimebenchmarkevaluating}, and (2) refreshing LMs through fine-tuning and prompting~\citep{nylund-etal-2024-time,grattafiori2024llama3herdmodels, gao-etal-2025-prompts,dhingra-etal-2022-time}.
It is often assumed that LMs have a notion of the current time, with little insight into how the concept is learned and represented. 

In this work, we investigate whether LMs possess a consistent notion of the current year. We choose ``year'' as a discrete, reasonably granular unit of the current time and design tasks that probe the current year in conceptually distinct ways. 
Specifically, we define an \textit{associative} task, which infers the current year from verb tense, and a \textit{declarative} task, which directly queries for the current year. 
For both tasks, we analyze LM behaviors, internal representations, and connections to training data distributions.

We find that LMs exhibit consistent behavior and converge to within one year of their training data cutoff on both tasks after SFT. The associative task also provides a strong proxy for the pre-training cutoff year, off by only 0.85 years across 13 base LMs, an improvement over past proxies \citep{cheng2024dateddatatracingknowledge}. In contrast, the declarative current year is only acquired in post-training, shaped by both instruction following and the year distribution in SFT data.

While there is behavioral alignment between the tasks, we find that they correspond to different internal mechanisms. Using causal interventions to trace internal mechanisms, we show that the associative current year is similar to factual recall~\citep{meng2023locatingeditingfactualassociations, geva-etal-2023-dissecting}. In contrast, the declarative current year lacks a consistent internal mechanism.

Next, we assess whether having different mechanisms poses challenges for updating the current year, which is a necessary step in refreshing LMs. We evaluate three families of approaches: specifying a target current year in context, fine-tuning on year-shifted data, and low-rank weight editing. The declarative current year is easily shifted by all methods targeting it, with an accuracy up to 94.6\% across 351 target years using a system prompt. The associative current year resists being shifted by both prompting, which has a success rate of only 1.7\%, and fine-tuning on year-shifted data, which succeeds only for target years near the base LM's associative current year. Weight editing, while effective for the associative and declarative tasks individually, does not generalize across both: Weight edits customized to the associative current year shift it within six years of the target across eight weight edited LMs, but leave the declarative current year nearly untouched.

Taken together, our findings suggest that LMs do not have a consistent notion of the current year: The associative notion, which is pervasively encoded in language and deeply embedded during pre-training, relies on distinct causal mechanisms and proves resistant to the existing interventions that update the declarative notion acquired in post-training.

%% file: related_work.tex
\section{Related work}
\paragraph{Time-dependent behaviors in LMs} Recent work has evaluated chronological knowledge and temporal reasoning in LMs~\citep{park2025chroknowledgeunveilingchronologicalknowledge, fatemi2024testtimebenchmarkevaluating, li2026simulatedignorancefailssystematic, wallat2025studyinvestigatingtemporalrobustness}, proposed refreshing LMs through finetuning on dated or temporally ordered data~\citep{dhingra-etal-2022-time, jang-etal-2022-temporalwiki}, used model steering~\citep{nylund-etal-2024-time, an2026timetravelengineshared}, and tried to induce alternative current time in LMs through prompting, for both benign and adversarial behaviors~\citep{gao-etal-2025-prompts, underwood2025languagemodelsrepresentpast, hubinger2024sleeperagentstrainingdeceptive}. While LMs may be aware of the current time, these approaches offer little insight into how the notion of current time is represented in LMs or where it originates from.

\paragraph{Localizing and editing concepts in LMs}
Our work uses causal methods to localize current year mechanisms in LMs. This includes null interventions, e.g., attention or MLP sublayer knockouts~\citep{geva-etal-2023-dissecting}; interchange interventions, e.g., swapping representations between entities of the same type~\citep{ meng2023locatingeditingfactualassociations,geiger2021causal, JMLR:v26:23-0058}, and learned interchange interventions over feature subspace, e.g., distributed alignment search (DAS)~\citep{geiger2024findingalignmentsinterpretablecausal}. With these methods, previous work has localized and edited factual recall mechanisms in transformer-based LMs~\citep{geva-etal-2023-dissecting, meng2023locatingeditingfactualassociations, meng2023masseditingmemorytransformer, huang-etal-2024-ravel, merullo-etal-2024-language}. More recently, \citet{park2025doestimeplacetemporal} identified attention heads that support time-specific information recall.

%% file: associative.tex
\section{The associative current year}
\label{section:associative}
\input{figures/associative_behavior_1950_2050}

We first consider an \emph{associative} notion of the current year, where the value of the current year is implicitly encoded in the associations between two words. We focus on one common grammatical structure that represents this associative notion of time: verb-tense inflection.

\subsection{Definition}
The associative task uses prompts of the form \textit{In \{year\} there}, where the predicted next token is a verb whose tense reflects the LM's implicit notion of the current year. For example, for \textit{In 1960 there}, predicting \textit{was} implies the current year is after 1960.\footnote{This inference holds specifically for the prompt structure used here, where there are no embedding verbs or other constructions that shift the anchoring time away from the speaker's present, e.g., \textit{In 2030, they said that in 2027 there}, which should predict \textit{were}.}
The \textbf{associative current year} (ACY) is the first year $y$ within a given year range for which \(p_y(\mathrm{past}) < p_y(\mathrm{future})\), where \textit{was, were} are past tense and \textit{is, are, will} are future tense.\footnote{We group present and future because a verb can serve in both present and future grammatical roles since English lacks clear morphological distinctions between present and future tense in verbs. For example, a sentence starting with \textit{In 2025 there is} could indicate present tense but can have a future interpretation if continued with \textit{an election scheduled}.%
} Together, these five forms account for over 90\% of the next-token probability mass on average across the year range 1950--2050 for every one of the 13 LMs tested in Table~\ref{fig:associative_task}, making them a sufficient basis for our analysis. In practice, LMs' responses to the associative task shift coherently from past to present/future, as seen in Figure~\ref{fig:associative_task}.

\subsection{The associative current year is consistent across prompt variations}
We verify that the ACY definition offered above is consistent across two classes of prompt designs. First, we evaluate six alternative prompt templates, for example \textit{In the magic show in \{year\}, there magically}. Since these prompts induce a broader set of plausible verb continuations outside \textit{was/were/is/are/will}, we include verb continuations from a comprehensive dictionary-derived set of single-token past and present/future verb forms; details in Appendix~\ref{app:othermodels_verbs}. Second, we test whether the effect depends on the relative position of the year and tense in the sentence by evaluating twelve templates where the tense marker precedes the year. We measure the ACY by the full-sentence log-probabilities to compare the likelihood of past versus present/future tense realizations across years; details in Appendix~\ref{app:alt_acy_prompt}. Across both prompt variations, the inferred temporal shift remains consistent. 

In addition, we test whether the current year can be recovered from factual knowledge rather than verb grammar using two factual probes: one event-specific and another based on current knowledge (Appendix~\ref{app:associative_factual_recall}, Appendix~\ref{app:factual_recency}). Neither probe recovers the current year cleanly, suggesting that verb tense provides a more reliable and scalable signal of the current year.

\subsection{The associative current year accurately captures pre-training data cutoff date}

\paragraph{Setup} We measure the ACY across 13 base LMs. We also introduce a more granular prompt that measures the associative current year and month (ACYM), i.e., \textit{In \{year\} \{month\} there}. For comparison, we obtain estimates of each LM's pre-training cutoff from the perplexity-based method of~\citet{cheng2024dateddatatracingknowledge}, as described in Appendix~\ref{app:dated_data}.

\paragraph{Results} 
Table~\ref{tab:associative_current_model_cutoff} compares the ACY and ACYM against the LMs' pre-training cutoff dates and the corresponding perplexity predictions. Perplexity predictions have an average error of 2.71 years across the 13 base LMs. The ACY and ACYM reduce the average error to 0.85 years ($\approx$10 months) and 0.75 years ($\approx$9 months) respectively across all 13 LMs, though \texttt{OLMo2-7B} accounts for over half of the ACYM error, with a five-year deviation despite a one-year ACY error.


\subsection{The associative current year is shaped by the pre-training data distribution}
\label{sec:associative_data_dist}

To better understand how LMs acquire the ACY, we investigate how the ACY aligns with the pre-training data distribution. We study \texttt{OLMo2}~\citep{olmo20242olmo2furious}, whose pre-training proceeds in two stages: a first stage over a large general web corpus and a second stage over a smaller, higher-quality curated mix.

\paragraph{Methods} We represent the pre-training data distribution with two counting-based statistical models: (1) a co-occurrence model that counts instances of years and verb tenses appearing within the same sentence, and (2) an n-gram model that counts instances of the exact string \textit{In [year] there} followed by a verb tense (details in Appendix~\ref{app:td_counting_ngram_cooccur}). The co-occurrence model is the more permissive of the two, counting how years relate to tenses, and the n-gram model the more conservative, counting data that matches the ACY prompt exactly. Applying the ACY definition to a count model gives the current year implied by that stage's data, which we take as the reference against which we compare the LMs' ACY.
Separately, we count how often the verb tenses occur in each stage for each counting model and create a baseline model that is simply the prior distribution without being conditioned on a specific year.

\paragraph{Setup} 
We compare the ACY of \texttt{OLMo2-1B} and \texttt{OLMo2-7B}~\citep{olmo20242olmo2furious} across pre-training stages against our count models, constructed from 10k steps of stage 1 data (olmo-mix, \citealp{allenai_olmo-mix-1124}) and 5.5k steps of stage 2 data (dolmino-mix, \citealp{allenai_dolmino-mix-1124}).
Moreover, we compute the cross entropy (CE) between the LMs and the n-gram model, the co-occurrence model, and the baseline model, where we compute CE per year and average over years in the range 1950--2050. Lastly, to understand how the ACY develops over training, we compute the ACY at multiple training checkpoints.

\paragraph{Results}
First, as shown in Table~\ref{tab:associative_stage_comparison}, both \texttt{OLMo2} LMs and both count models shift their ACYs at the stage boundary; in stage 1 all are $\leq$ 2022, rising to $\geq$ 2023 after stage 2. Moreover, as seen in Figure~\ref{fig:training_dynamics_1b}, the ACY remains stable within each stage (2020--2022 in stage 1, 2023--2024 in stage 2) and persists through post-training (results for \texttt{OLMo2-7B} in Appendix~\ref{app:training_dynamics_on_olmo_7b}). The n-gram and co-occurrence count distributions closely match both \texttt{OLMo2} LMs tense distributions, with a CE of at most 0.36 and 0.78 respectively, which is 0.12 and 0.33 below baseline (details in Appendix~\ref{app:paragraph:crossentropy}).

Taken together, these results suggest that the ACY is shaped by pre-training data.

\subsection{LMs causally encode a year-type in a small linear subspace}
\label{subsubsec:caus_abs}
How do LMs represent the ACY internally, i.e., what is the mechanism that determines the proper verb tense given a current year? The year-tense structure in training data resembles the entity-attribute structure studied in factual associations: the year acts as an entity and its tense as an attribute. This suggests that LMs may retrieve tense from year via \emph{factual recall}~\citep{meng2023locatingeditingfactualassociations,geva-etal-2023-dissecting,huang-etal-2024-ravel}, so we hypothesize that a subspace in the activation space can represent this tense attribute. 

\paragraph{Methods} 
We apply the causal abstraction framework~\citep{geiger2021causal,JMLR:v26:23-0058} to localize a ``year'' variable. Specifically, we hypothesize a high-level causal model consisting of two variables, a year variable $\mathcal{Y}$ and a tense variable $\mathcal{T}$, such that $\mathcal{Y}\rightarrow \mathcal{T}$. We use distributed alignment search (DAS)~\citep{geiger2024findingalignmentsinterpretablecausal} to localize the year variable $\mathcal{Y}$ in LM activations, by identifying a linear subspace in the residual stream that corresponds to it. If a subspace indeed represents a year variable, then intervening on that subspace by replacing its activation with the representation of a different year should reliably shift the LM's tense prediction. Concretely, given a base prompt containing one year and a source prompt containing a different year, we replace the activation of this subspace during the base run with the activation from the source run, and check whether the LM's tense prediction shifts accordingly. We measure the extent to which these interventions produce the desired outcome using interchange intervention accuracy (IIA)~\citep{geiger2021causal}.

\paragraph{Setup} We conduct the localization experiment on \texttt{OLMo2-7B}. We construct counterfactual pairs using prompts of the form \textit{In [year], there}, where the year ranges from 1000 to 4000. 86\% of counterfactual pairs have base and source examples with different predicted tenses so that counterfactual effects are measurable. We then use DAS to isolate a one-dimensional subspace on a specific token and layer, repeated across all tokens in the prompt and layers in \texttt{OLMo2-7B} (details in Appendix~\ref{app:detailed_subspace_analysis}). 
\paragraph{Results}
\input{figures/iia_heatmap}

The localized result, i.e., IIA per token and layer, is shown in Figure~\ref{fig:iia_heatmap}. We are able to find a one-dimensional subspace that encodes the ``tense'' attribute of the ``year'' variable in the residual streams of layer 4 to layer 22, above the last token of the year.\footnote{4-digit numbers in \texttt{OLMo2} LMs are split into two tokens, e.g.\ ``2002'' is split into ``200'' and ``2''.}
We further verify that this subspace is non-trivial and is specific to year expressions: no such subspace can be found using DAS in a randomly initialized LM or by randomly selecting a subspace in a pre-trained LM residual stream. We also cannot successfully perform interchanges with entities that do not represent years, such as \textit{In summary there} (details in Appendix~\ref{app:detailed_subspace_analysis}).

Moreover, interchanging representations from prompts with four digit numbers that are not used in a temporal context results in interchange intervention accuracies of 90\%+ on early layers, showing the year-type subspace is successful for numbers that look like years regardless of their role (details in Appendix~\ref{app:detailed_subspace_analysis}). This suggests that the subspace we find can generalize to a broader set of four digit number representations. Causal tracing through attention and MLP knockouts corroborates that the same layers on the same token are important (details in Appendix~\ref{app:associative_knockouts}).

%% file: figures/associative_behavior_1950_2050.tex
\begin{figure*}[t]
\centering
\small{
\setlength{\aboverulesep}{0.2ex}
\setlength{\belowrulesep}{0.25ex}
\begin{tabular}{lcccc}
\toprule
\textbf{} & \textbf{Training Data} & \textbf{Perplexity} & \textbf{} & \textbf{} \\
\textbf{Model} & \textbf{Cutoff} & \textbf{Prediction} & \textbf{ACY} & \textbf{ACYM} \\
\midrule
Pythia-1B-dedup & Sept 2020 & Mar 2020 (--6) & 2020 (0) & Jan 2021 (+4) \\ 
Pythia-1.4B-dedup & Sept 2020 & Mar 2020 (--6) & 2020 (0) & Jan 2021 (+4) \\ 
Pythia-6.9B-dedup & Sept 2020 & Mar 2020 (--6) & 2019 (--12) & Jul 2020 (--2) \\ 
\midrule
GPT-NeoX-20B & Sep 2020 & Mar 2020 (--6) & 2020 (0) & May 2020 (--4) \\ 
\midrule
GPT-J-6B & Sep 2020 & Mar 2020 (--6) & 2019 (--12) & Nov 2020 (--2) \\ 
\midrule
OLMo-7B & Feb/Mar 2023 & Feb/Mar 2020 (--36) & 2022 (--12) & Jan 2023 (--1/--2) \\  
\midrule
Falcon-RW-7B & Jan/Feb 2023 & Jan 2020 (--36/--37) & 2021 (--24) & May 2022 (--8/--9) \\ 
\midrule
RedPajama-7B-Base & Jun 2023 & Jun 2019 (--48) & 2021 (--24) & Jan 2023 (--5) \\ 
\midrule
OLMo2-1B & Dec 2023 & Oct 2018 (--62) & 2023 (0) & Jan 2024 (+1) \\
OLMo2-7B & Dec 2023 & Sep 2018 (--63) & 2024 (+12) & Jan 2029 (+61) \\
\midrule
Llama3.2-1B & Dec 2023 & Nov 2020 (--37) & 2022 (--12) & Apr 2023 (--8) \\
Llama3.2-3B & Dec 2023 & May 2019 (--55) & 2022 (--12) & Apr 2023 (--8) \\
Llama3.1-8B & Dec 2023 & May 2019 (--55) & 2022 (--12) & Apr 2023 (--8) \\
\bottomrule
\end{tabular}
}
\captionof{table}{\textbf{The associative current year is an accurate proxy for pre-training data cutoff year}. Averaged over the 13 LMs in the table, the ACY and ACYM deviate from the training data cutoff date by 0.85 and 0.75 years respectively, against 2.71 years for perplexity predictions on the WIKISPAN dataset~\citep{cheng2024dateddatatracingknowledge}. Perplexity predictions for the first eight LMs are from \citet{cheng2024dateddatatracingknowledge}; the remaining five are obtained by replicating their method (Appendix~\ref{app:dated_data}). Estimation errors in ``()'' are reported as absolute deviations from the training data cutoff in months.} 
\label{tab:associative_current_model_cutoff}

\vspace{2ex}

\begin{minipage}[t]{0.60\linewidth}
  \vspace{-3ex}
  \centering\includegraphics[width=\linewidth,clip=true,trim={0ex 8ex 0ex 0ex}]{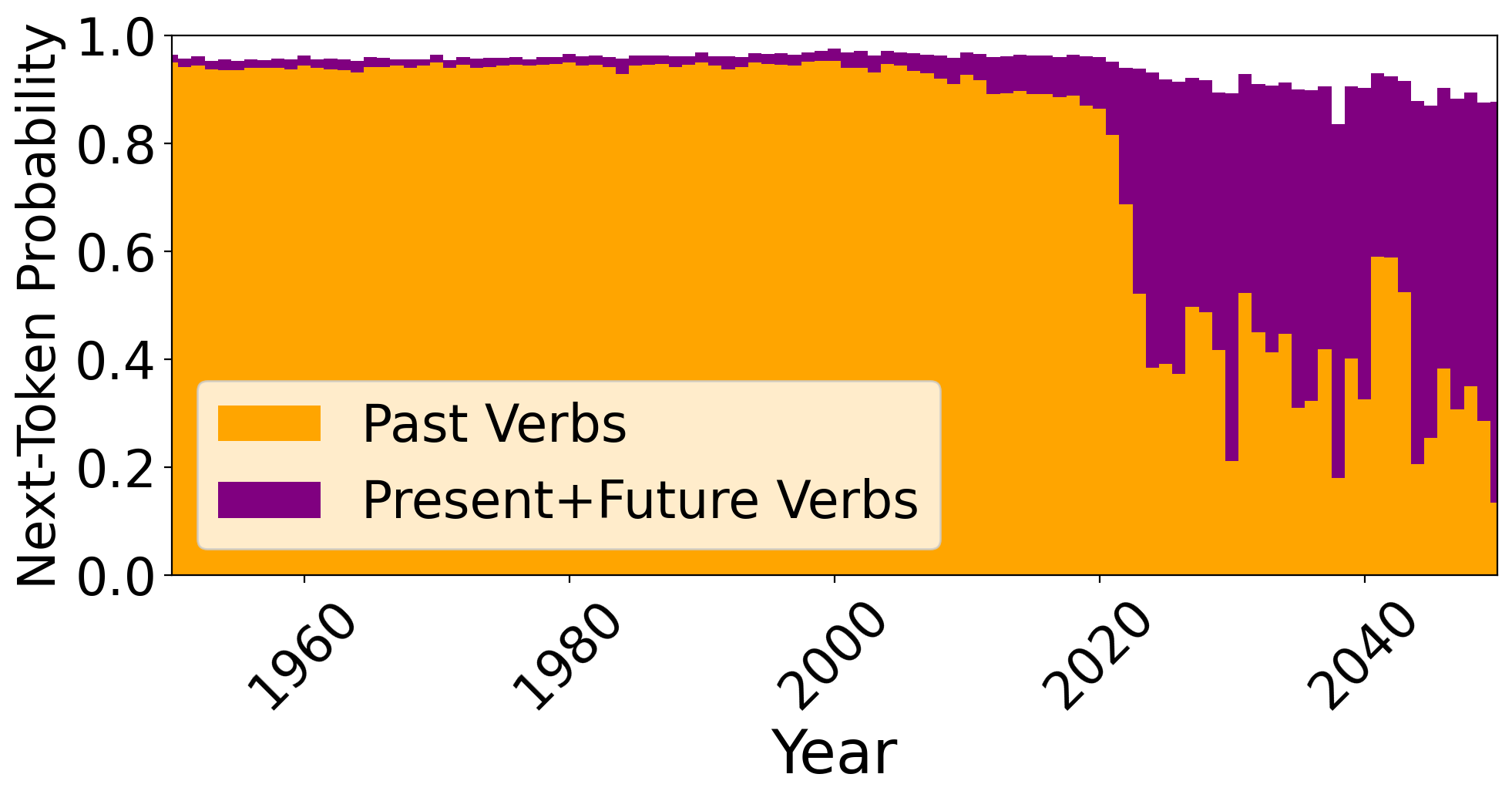}
  \vspace{-4ex}
  \captionof{figure}{\textbf{\texttt{OLMo2-7B} shows a coherent distinction of past/future on the associative task}. We plot the next-token probabilities for the prompt \textit{In [year] there} over 1950--2050, categorized by tense.}
  \label{fig:associative_task}
\end{minipage} \hfill
\begin{minipage}[t]{0.36\linewidth}
  \vspace{-3ex}
  \centering
  \begin{tabular}{lcc}
    \toprule
    \textbf{Model} &
    \textbf{Stage 1} &
    \textbf{Stage 2} \\
    \midrule
    OLMo2-1B        & 2021 & 2023 \\
    OLMo2-7B        & 2022 & 2024 \\
    \midrule
    N-gram           & 2020 & 2024 \\
    Co-occurr.       & 2017 & 2023 \\
    \bottomrule
  \end{tabular}
  \captionof{table}{
  \textbf{The associative current year shifts with the pre-training data distribution between stages.} We report the ACY of \texttt{OLMo2-1B} and \texttt{OLMo2-7B} and of the two count models which give the year implied by each stage's data, over 1950--2050.}
  \label{tab:associative_stage_comparison}
\end{minipage}

\vspace{-3ex}
\end{figure*}


%% file: figures/iia_heatmap.tex
\begin{figure}[t]
    \begin{minipage}[t]{0.53\linewidth}
    \vspace{0pt}
    \centering
    \includegraphics[width=\linewidth]{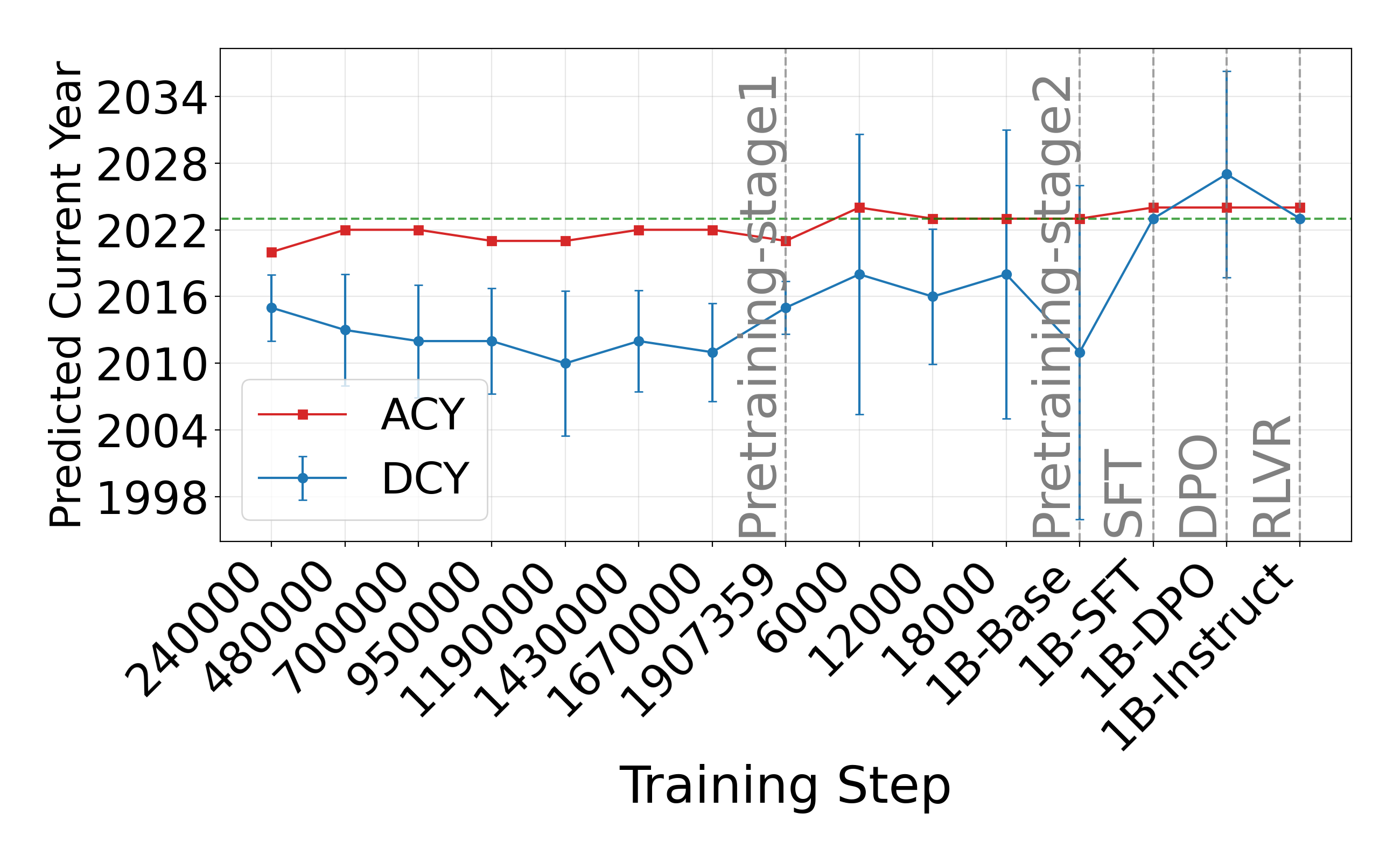}
    \vspace{-4ex}
    \caption{\textbf{\texttt{OLMo2-1B}’s ACY and DCY over training checkpoints are aligned to the training data cutoff year (2023, horizontal line) after SFT.} ACY is stable within each pre-training stage; DCY aligns only at SFT. Points represent the ACY and DCY; bars represent the standard deviation.}
    \label{fig:training_dynamics_1b}
  \end{minipage}
  \hfill
 \begin{minipage}[t]{0.45\linewidth}
    \vspace{0pt}
    \centering
\includegraphics[width=\linewidth,clip=true,trim={0ex 2ex 0ex 0ex}]{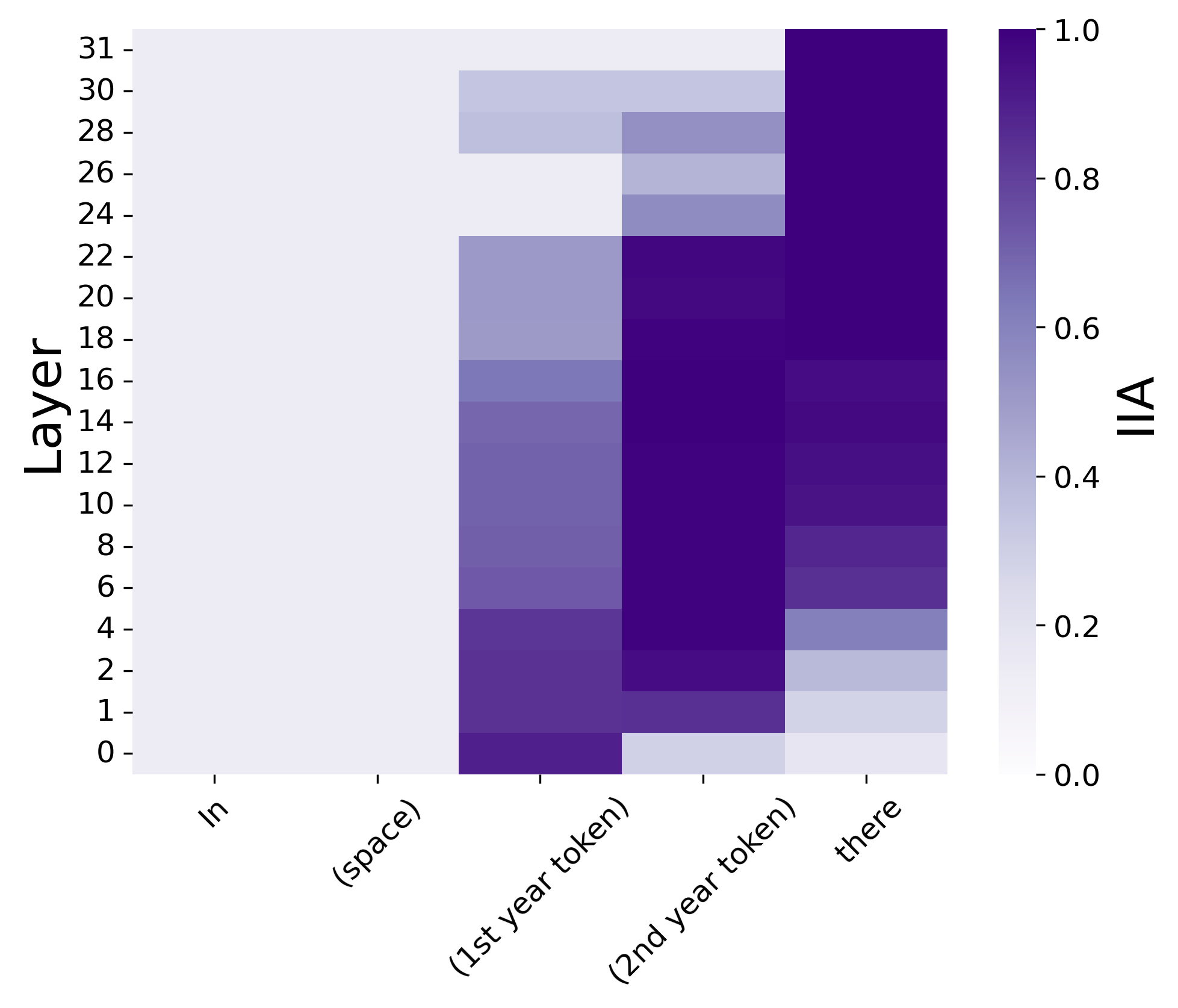}
    \vspace{-4ex}
    \caption{\textbf{A year-type subspace can be found in multiple layers on the year's last token.} We show the IIA on \texttt{OLMo2-7B} after DAS at each layer.}
    \label{fig:iia_heatmap}
  \end{minipage}
  \vspace{-3ex}
\end{figure}

%% file: declarative.tex
\section{The declarative current year}
\label{section:declarative}
In contrast to the implicitly encoded \textit{associative} year, we now investigate a \textit{declarative} notion of the current year that is explicitly verbalized by the LM.

\subsection{Definition}
The declarative task directly queries the LM for the current year. To ensure robustness, we use a set of 10 prompts of two forms: continuation prompts (e.g., \textit{Write a story about the current year. Story: The year is}) used for base LMs, and instruction prompts (e.g., \textit{Write a story about the current year, starting with the current year.})\ used for post-trained LMs (details in Appendix~\ref{app:declarative_prompts}). The \textbf{declarative current year} (DCY) is the average greedy-decoded year the LM predicts across the declarative prompts, rounded to the nearest integer.

\subsection{The declarative current year is learned during SFT}

\paragraph{Setup} 
We study the properties and development of the DCY over pre-training and post-training using \texttt{OLMo2-1B} and \texttt{OLMo2-7B}. We measure the DCY, its standard deviation, and its associated confidence. The confidence is the probability of the LM's greedy-decoded year.
We also compare each LM's DCY to the year distribution of the data it was trained on, counting the occurrences of each year in the phrasing \textit{In [year]} over the range 1900--2100 and determining the data's peak year. Since every training stage uses a different dataset, each has its own peak year: olmo-mix (\texttt{OLMo2} stage 1 pre-training, \citealp{allenai_olmo-mix-1124}), dolmino (stage 2 pre-training, \citealp{allenai_dolmino-mix-1124}), and two versions of Tulu3 (\texttt{OLMo2-1B} SFT, \citealp{allenai2025tulu3sftolmo2}; \texttt{OLMo2-7B} SFT, \citealp{allenai2025tulu3sftolmo2_7b}); details in Appendix~\ref{app:declr_tr_data_counts}. The Peak column of Table~\ref{tab:declarative_current_year} reports the most frequent year in each LM's own training data.

\paragraph{Results}
\input{tables/behavior_declarative_summary}
We summarize the performance of \texttt{OLMo2} LMs across the two pre-training stages and after SFT (the first post-training step) in the pre-training and released post-training rows of Table~\ref{tab:declarative_current_year}. During pre-training, the DCY does not align with the pre-training data cutoff (2023), and is not consistently aligned with the most frequent year in pre-training data (2008 and 2010 for stage 1 and 2). After SFT, the DCY aligns to the post-training data cutoff date and the most frequent year in post-training data, 2023. Moreover, during pre-training, the 10 declarative prompts elicit different years as output, whereas after SFT, variance goes to zero. In addition, SFT LMs are 10 times more confident in the DCY than base LMs. The DCY remains relatively unchanged through later post-training stages as seen in Figure~\ref{fig:training_dynamics_1b}, suggesting SFT is the critical stage for temporal alignment. Taken together, SFT produces a DCY that aligns with the training data cutoff, with reduced variance and increased confidence.

\subsection{The declarative current year is causally affected by the year distribution in SFT data}
\label{sec:declarative_sft}
To investigate whether SFT improves the DCY through instruction-following or through the year distribution in the SFT data, we retrain LMs on modified SFT data. We fine-tune on data without any years to isolate the influence of instruction-following, and on data where all years are shifted to measure the effect of changing the year distribution.

\paragraph{Setup} 
We fine-tune LMs from \texttt{OLMo2-1B} using Tulu3 data \citep{allenai2025tulu3sftolmo2}. To allow the same fine-tuning setup across all LMs, we create a control LM,  \texttt{OLMo2-1B-SFT-control}. Then, we modify Tulu3 data in two ways. For the no-years SFT LM, we remove all training examples with numbers $\geq$ 1000, preserving 76\% of training data. For year-shifted LMs, we shift all numbers between 1000 to 9999 by a constant N, such as +200 (details in Appendix~\ref{app:sft_training}).

\paragraph{Results}
We assess the causal effects of modified SFT data on the DCY. The no-years and control LMs are reported in the last two rows of Table~\ref{tab:declarative_current_year}; the eight year-shifted LMs are reported in Table~\ref{tab:declr_vs_assoc_oneyear_col} under ``Year-Shifted SFT'', one row per target year.
The no-years SFT LM predicts a DCY of 2021 with zero variance, compared with a $\pm$15.0 spread across prompts for the base \texttt{OLMo2-1B} (Stage 2 (1B) in Table~\ref{tab:declarative_current_year}). SFT on data with all years removed therefore collapses the prompt-to-prompt variance from 15.0 to 0.0 on its own. It does not, however, reach the control SFT LM's 2023; this remaining gap shows that the fine-tuning procedure alone does not fix the DCY, and that the years present in the SFT data also matter.
For the year-shifted SFT LMs, the DCY accurately shifts by N with near-zero variance.
Thus, both instruction-following skills and data distribution contribute: instruction following enables the LM to surface a coherent declarative year, while the year distribution in the SFT data determines which year it predicts.

\subsection{The declarative current year uses different causal pathways than the associative current year} 

To find whether a consistent mechanism is shared among the declarative prompts, and whether these share a mechanism with the associative task, we use causal interventions to trace which tokens and layers contribute to the year chosen as the declarative year. 

\paragraph{Methods}
We apply attention and MLP sublayer knockouts on five consecutive layers at a time on one token, evaluating only the probability of the year prefix token (e.g., ``202''). We use the prefix because using the year suffix would limit the output range to only ten years (e.g. 2020--2029).

To quantify how sublayer knockouts change the LM's year prefix, we define a candidate set \(\mathcal{D}\) of tokens corresponding to three-digit year prefixes (e.g., ``202''). 
For each prompt, let $t^\star = \arg\max_{t \in \mathcal{D}} p_{\mathrm{before}}(t)$ be the pre-intervention top token within $\mathcal{D}$. We use a top-vs-rest margin $m = p(t^\star) - \sum_{t \in \mathcal{D}\setminus\{t^\star\}} p(t)$ since it is possible that post-intervention, the top token is not in \(\mathcal{D}\), i.e. the LM does not predict a year token. We report the year-shift magnitude (YSM) as the absolute difference before and after the sublayer knockout; $\mathrm{YSM} = \left| m_{\mathrm{after}} - m_{\mathrm{before}} \right|$.

\paragraph{Setup} 
We evaluate on \texttt{OLMo2-7B} using four prompts that vary along two dimensions: keyword choice (\textit{date today} vs.\ \textit{current year}) and formatting (question vs.\ continuation), yielding \textit{What is the date today?}, \textit{What is the current year?}, \textit{Date today:}, and \textit{Current year:}. Each prompt is expanded into a group of ten by prepending two-shot examples to elicit a year token (details in Appendix~\ref{app:declarative_causal_tracing}). We report YSM averaged within each prompt group, computed only on tokens shared across all ten prompts in the group.

\paragraph{Results} Knockouts at the first token position are often the most influential, even when that token is not time-related, as can be seen in the MLP knockouts in Figure~\ref{fig:today_date_mean_section_4_4}. This is likely due to the attention-sink effect observed in~\citet{xiao2024efficient}. This contrasts with the associative task, where the most influential tokens are specifically the \emph{year token(s)} (rather than a fixed position). Moreover, the variance of the YSM across prompts within a group is low, indicating localization is fairly consistent within each prompt group (details in Appendix~\ref{app:declarative_causal_tracing}).
Across prompt groups, the influential tokens and layer ranges vary, and we do not observe a consistent localization pattern that generalizes across formatting or keywords. Since the associative task exhibits a comparatively consistent mechanism centered on the year token, these results suggest that declarative ``current year'' decoding does not rely on the same causal pathway as the associative year-tense mechanism.

%% file: tables/behavior_declarative_summary.tex
\begin{figure}[t]
  \begin{minipage}[t]{0.54\linewidth}
    \vspace{0pt}
    \centering
    \small
    {\setlength{\tabcolsep}{1pt} 
    \begin{tabular}{lccc}
      \toprule
       \textbf{LM} & \textbf{Peak} & \textbf{DCY} & \textbf{Confidence} \\
      \midrule
      \multicolumn{4}{l}{\textit{Pre-training, AI2}} \\
      Stage 1 (1B)  & 2008 & 2015 $\pm$ 2.3  & 4.3\% $\pm$ 1.9\%  \\
      *Stage 1 (7B) & 2008 & 2016 $\pm$ 3.7  & 5.0\% $\pm$ 2.5\%  \\
      Stage 2 (1B)  & 2010 & 2011 $\pm$ 15.0  & 6.1\% $\pm$ 5.7\%  \\
      Stage 2 (7B)  & 2010 & 2016 $\pm$ 3.8  & 5.7\% $\pm$ 3.8\%  \\
      \midrule
      \multicolumn{4}{l}{\textit{Post-training, AI2}} \\
      SFT (1B)            & 2023 & 2023 $\pm$ 0.0  & 60.0\% $\pm$ 19.3\% \\
      SFT (7B)            & 2023 & 2023 $\pm$ 0.0  & 52.5\% $\pm$ 26.4\% \\
      \midrule
      \multicolumn{4}{l}{\textit{Post-training, ours}} \\
      SFT-control (1B)    & 2023 & 2023 $\pm$ 0.0 & 47.5\% $\pm$ 22.6\% \\
      *No-years (1B)  & N/A  & 2021 $\pm$ 0.0   & 19.0\% $\pm$ 9.9\%  \\
      \bottomrule
    \end{tabular}
    \vspace{-1ex}
    \captionof{table}{\textbf{The DCY aligns with the training data cutoff in post-training.} $*$ marks rows where 9 of 10 declarative prompts yield a year. \textit{ours} are unmodified and year-free variants (Appendix~\ref{app:sft_training}).}
    \label{tab:declarative_current_year}
    }
  \end{minipage}%
  \hfill
  \begin{minipage}[t]{0.44\linewidth}
    \vspace{0pt}
    \centering
    \includegraphics[width=\linewidth]{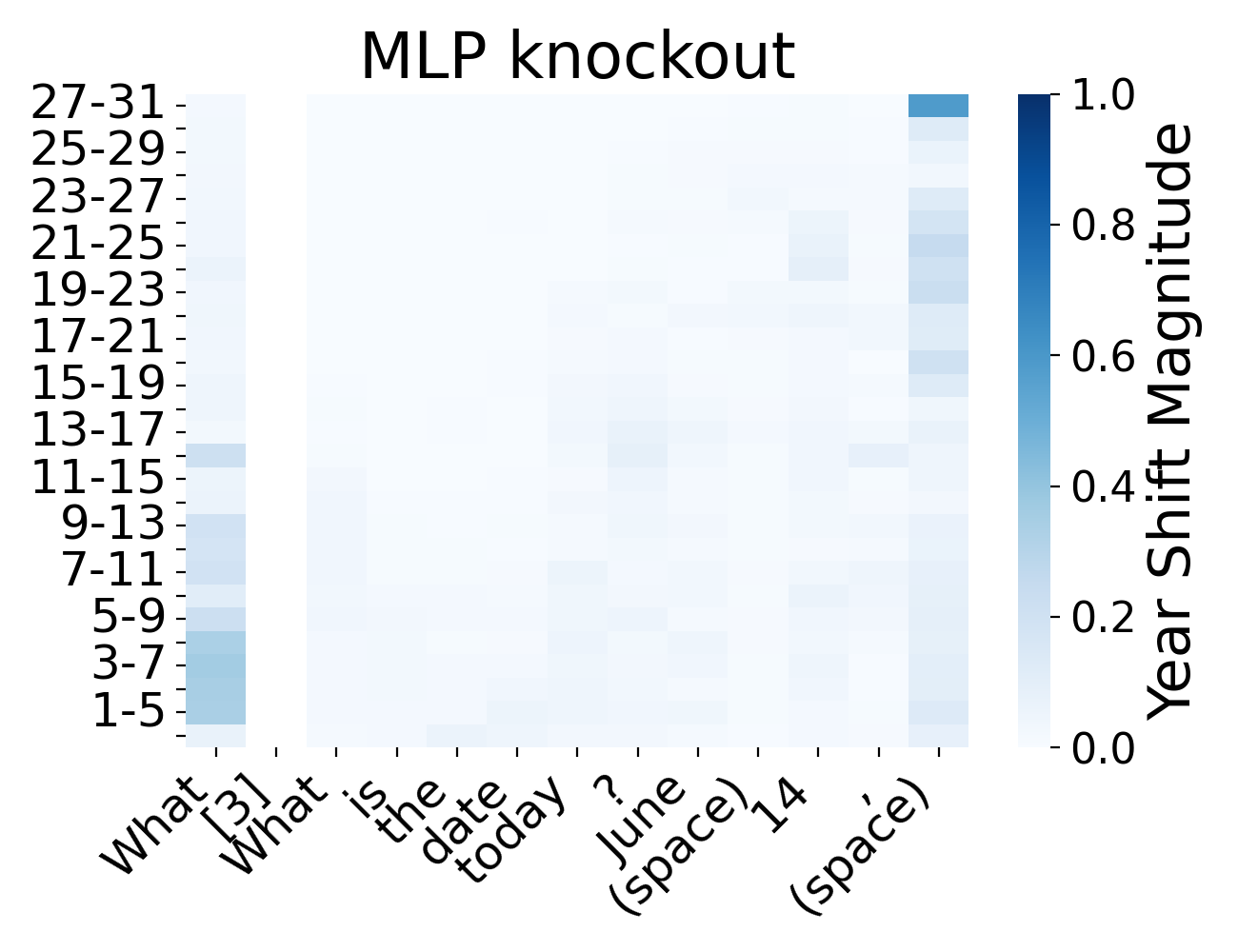}
    \vspace{-4ex}
    \caption{\textbf{Non-time related tokens can have more effect on the YSM than time related ones.} ``[3]" represents omitted fewshot examples.}
    \label{fig:today_date_mean_section_4_4}
  \end{minipage}
  \vspace{-3ex}
\end{figure}

%% file: in_context.tex
\section{Updating the current year in LMs}
\label{section-5}

\input{tables/updating_current_year.tex}

The existence of at least two distinct notions of the current year in LMs, i.e., \textit{declarative} and \textit{associative}, could complicate  efforts to refresh LMs. In this section, we focus on assessing three commonly used techniques to update the current year. 

\subsection{Methods}
\paragraph{Method I: Fine-tuning on year-shifted data} We use the LMs from Section~\ref{sec:declarative_sft}, which are trained on data with years shifted by an offset. We aim to shift the ACY and DCY by the offset N, e.g. 2023 + 200 = 2223 for the +200 shifted LM.

\paragraph{Method II: Specifying the current year in context (with/without SFT)} 
Open-weight LMs such as Llama-3~\citep{grattafiori2024llama3herdmodels} include the current date in its system prompt. Some closed-source LMs~\citep{gao-etal-2025-prompts} also specify the date in-context to simulate date shifting. To test whether this requires special training, we compare \texttt{OLMo2-1B-SFT-control} and \texttt{OLMo2-1B} finetuned with the system prompt \textit{The current year is 2023}. When evaluating, we use the system prompt \textit{The current year is [target year]}. 

\paragraph{Method III: Low-rank weight editing}
We use low-rank weight editing methods to shift the LM's current year. For the declarative current year (DCY), we use ROME \citep{meng2023locatingeditingfactualassociations}, a single-fact update.
For the associative current year (ACY), we use MEMIT \citep{meng2023masseditingmemorytransformer}, updating year--tense associations.

\paragraph{An additional metric: The associative cross-entropy}
In addition to the ACY, we introduce the associative cross-entropy (ACE), a continuous measure of how well the tense distribution over years matches a golden distribution. Suppose that the target current year is $c$. We measure the per-year cross-entropy between a golden distribution $g_y(\mathrm{past}) = \mathbbm{1}[y < c]$ and the predicted tense distribution $p_y(\mathrm{tense}), \mathrm{tense}\in\{\mathrm{past}, \mathrm{future}\}$ where  probabilities of past and future tenses are normalized to 1. The ACE is computed as the average cross-entropy across all years in the evaluated range $\mathcal{Y}$: $\mathrm{CE}_{\mathcal{Y}} = -\frac{1}{|\mathcal{Y}|}\sum_{y \in \mathcal{Y}}\sum_{\mathrm{tense}} g_y(\mathrm{tense})\log(p_y(\mathrm{tense}))$.

\subsection{Setup} 
We apply all methods to re-fine-tuned versions of \texttt{OLMo2-1B-SFT} and evaluate their ability to shift the ACY and DCY to a target year in the range 1900--2250, using an instruction variation of the ACY prompt and the instruction declarative prompts for DCY.
For MEMIT, we spread the edit across layers 7--10. To avoid training on the same prompt family used for ACY evaluation, we train on disjoint templates consisting of a random prefix plus \textit{In/During [year] he/she/they}, sampling one template per training year. Training years include 80\% of years in the source--target interval, plus 20 additional past years and 25 sampled future years relative to the target in the range 1900--2250. 
For ROME, we train on prompts with a random prefix plus \textit{The current year is}. We sweep layers 7--10 and select the layer with the smallest total error on the train prompts, which is layer 10.

\subsection{Results}
\paragraph{The declarative current year can be easily updated by all methods}
As shown in Table~\ref{tab:declr_vs_assoc_oneyear_col}, all three methods, when targeting DCY, accurately shift LM predictions to the target DCY. Specifying the current year in the system prompt is most promising in practice, requiring only a system prompt change to update the DCY.
Even \texttt{OLMo2-1B-SFT-control} without any special fine-tuning predicts the correct declarative year in 50.7\% of cases, on average off by 6.82 years when prompted with a target year in the range 1900--2250. The SFT LM trained with a system prompt performs even better, predicting the exact correct declarative year in 94.6\% of cases with an average error of 0.45 years. This indicates that LMs learn to use the year in the system prompt at prediction time when exposed to it during SFT.

\paragraph{The associative current year can only be updated by weight editing, yet the same edit does not update the declarative current year}
With in-context prompting over the range 1900--2250, \texttt{OLMo2-1B-SFT-control} and the system-prompt SFT LM predict the correct ACY in only 1.4\% and 1.7\% of 351 instances, respectively. Across all instances, the predicted ACY never falls outside 2019--2030, suggesting that prompting alone cannot meaningfully move the ACY; this is consistent with concurrent work that frequently co-occurring tokens in pre-training data are robust to distributional shifts~\citep{karkada2026symmetrylanguagestatisticsshapes}. This failure to shift the ACY is not specific to \texttt{OLMo2}: across additional recent LMs from the Gemma-3 and Qwen-3.5 families, prompting moves the ACY for at most 53.3\% of target years (below 40\% for every other LM) even when it shifts the DCY perfectly (Table~\ref{tab:prompting_scaling_acy}; details in Appendix~\ref{app:prompting_scaling}).

\input{tables/prompting_scaling_acy}

The year-shifted SFT LMs are slightly more promising: for target years 2017, 2029, and 2041, they achieve ACY errors within three years, suggesting that in-weight updates can shift the ACY over a wider range than prompting. However, beyond this range the predicted ACY collapses back to near 2020, indicating a limit to how far even fine-tuning can move the ACY.
In contrast, weight editing targeting the ACY shifts the ACY within six years of the target for all eight years tested. Moreover, it achieves the lowest ACE of any method, indicating that the associative distribution matches the golden distribution. While the ACY is successfully shifted, the DCY remains nearly unchanged at 2022/2023.

%% file: tables/updating_current_year.tex
\newcommand{\off}[1]{(#1)}

\begin{table}[t]
\centering
\small
{\setlength{\tabcolsep}{2pt}
\begin{tabular}{r ccc ccc}
\toprule
& \multicolumn{6}{c}{\textbf{Prompted with Target Year}} \\
\cmidrule(lr){2-7}
 &
\multicolumn{3}{c}{\textbf{Standard SFT}} &
\multicolumn{3}{c}{\textbf{Sys-Prompt SFT}} \\
\cmidrule(lr){2-4}\cmidrule(lr){5-7}
\textbf{Target} & \textbf{Decl.} & \textbf{Assoc.} & \textbf{ACE} & \textbf{Decl.} & \textbf{Assoc.} & \textbf{ACE} \\
\toprule
1988 & $*$1992 \off{+4} & 2022 \off{+34} & 0.64 & $*$1988 \off{+0} & 2022 \off{+34} & 0.96 \\
2005 & $*$2007 \off{+2} & 2020 \off{+15} & 0.43 & $*$2005 \off{+0} & 2020 \off{+15} & 0.67 \\
2008 & $*$2008 \off{+0} & 2020 \off{+12} & 0.40 & $*$2008 \off{+0} & 2020 \off{+12} & 0.63 \\
2017 & $*$2017 \off{+0} & 2020 \off{+3} & 0.36 & $*$2017 \off{+0} & 2020 \off{+3} & 0.63 \\
2029 & 2029 \off{+0} & 2029 \off{+0} & 0.57 & 2029 \off{+0} & 2029 \off{+0} & 0.91 \\
2041 & 2039 \off{-2} & 2024 \off{-17} & 0.53 & 2041 \off{+0} & 2024 \off{-17} & 0.81 \\
2058 & 2058 \off{+0} & 2030 \off{-28} & 0.62 & 2058 \off{+0} & 2030 \off{-28} & 0.90 \\
2223 & 2223 \off{+0} & 2023 \off{-200} & 0.30 & 2223 \off{+0} & 2023 \off{-200} & 0.38 \\
\bottomrule
\end{tabular}
}

\vspace{0.5em}

{\setlength{\tabcolsep}{2pt}
\begin{tabular}{r ccc ccc ccc}
\toprule
& \multicolumn{9}{c}{\textbf{Not Prompted with Target Year}} \\
\cmidrule(lr){2-10}
& \multicolumn{3}{c}{\textbf{Year-Shifted SFT}}
& \multicolumn{3}{c}{\textbf{DCY Weight Editing}}
& \multicolumn{3}{c}{\textbf{ACY Weight Editing}} \\
\cmidrule(lr){2-4} \cmidrule(lr){5-7} \cmidrule(lr){8-10}
\textbf{Target}
& \textbf{Decl.} & \textbf{Assoc.} & \textbf{ACE}
& \textbf{DCY} & \textbf{ACY} & \textbf{ACE}
& \textbf{DCY} & \textbf{ACY} & \textbf{ACE} \\
\toprule
1988 & 1988 \off{+0} & 2020 \off{+32} & 0.72 & 1988 \off{+0} & 2023 \off{+35} & 1.13 & $*$2023 \off{+35} & 1988 \off{+0} & 0.00 \\
2005 & $*$2005 \off{+0} & 2020 \off{+15} & 0.49 & 2005 \off{+0} & 2023 \off{+18} & 0.83 & 2023 \off{+18} & 1999 \off{-6} & 0.10 \\
2008 & $*$2008 \off{+0} & 2020 \off{+12} & 0.58 & 2008 \off{+0} & 2022 \off{+14} & 0.77 & $*$2023 \off{+15} & 2005 \off{-3} & 0.05 \\
2017 & $*$2017 \off{+0} & 2020 \off{+3} & 0.47 & $*$2017 \off{+0} & 2022 \off{+5} & 0.63 & $*$2022 \off{+5} & 2017 \off{+0} & 0.00 \\
2029 & $*$2030 \off{+1} & 2029 \off{+0} & 0.81 & 2029 \off{+0} & 2022 \off{-7} & 0.64 & $*$2023 \off{-6} & 2029 \off{+0} & 0.09 \\
2041 & 2041 \off{+0} & 2040 \off{-1} & 1.24 & 2041 \off{+0} & 2020 \off{-21} & 0.69 & $*$2023 \off{-18} & 2041 \off{+0} & 0.02 \\
2058 & 2058 \off{+0} & 2050 \off{-8} & 1.52 & 2058 \off{+0} & 2022 \off{-36} & 0.75 & $*$2023 \off{-35} & 2060 \off{+2} & 0.05 \\
2223 & $*$2223 \off{+0} & 2022 \off{-201} & 0.31 & $*$2223 \off{+0} & 2020 \off{-203} & 0.73 & $*$2022 \off{-201} & 2223 \off{+0} & 0.04 \\
\bottomrule
\end{tabular}
}
\caption{\textbf{None of prompting, SFT, or weight editing shifts the associative and declarative current year simultaneously.} We attempt to update the ACY and DCY on versions of \texttt{OLMo2-1B}. +/- values are signed differences of predicted year to the target year. The associative cross-entropy~(ACE) is calculated on the range (1900, 2250). Asterisks (*) indicate the LM responses to at least 6 out of 10, but not all declarative prompts with a year.}
\label{tab:declr_vs_assoc_oneyear_col}
\end{table}

%% file: tables/prompting_scaling_acy.tex
\begin{table}[t]
\centering
\small
\begin{tabular}{l cc}
\toprule
\textbf{Model} & \textbf{DCY accuracy} & \textbf{ACY accuracy} \\
\midrule
OLMo2-1B Standard SFT   & 0.513 & 0.011 \\
OLMo2-1B Sys-prompt SFT & 0.940 & 0.020 \\
OLMo2-7B Standard SFT   & 1.000 & 0.157 \\
\midrule
Gemma3-1B-it  & 0.288 & 0.014 \\
Gemma3-4B-it  & 0.231 & 0.399 \\
Gemma3-12B-it & 0.986 & 0.311 \\
Gemma3-27B-it & 0.296 & 0.256 \\
\midrule
Qwen3.5-0.8B & 0.570 & 0.003 \\
Qwen3.5-2B   & 0.345 & 0.362 \\
Qwen3.5-4B   & 1.000 & 0.006 \\
Qwen3.5-9B   & 1.000 & 0.011 \\
Qwen3.5-27B  & 1.000 & 0.533 \\
\bottomrule
\end{tabular}
\caption{\textbf{Shifting the ACY through prompting fails across model families and scales.} We attempt to update the DCY and ACY of nine additional LMs by specifying the target current year in the system prompt. DCY and ACY accuracy are the fraction of target years for which the predicted year exactly matches the target, computed over 351 target years in 1900--2250.}
\label{tab:prompting_scaling_acy}
\end{table}

%% file: conclusion.tex
\section{Conclusion}
We introduce a declarative and associative task that probe conceptually distinct notions of the current year in LMs, revealing that what behaviorally appears as a single concept is not consistently encoded.

The ACY is learned during pre-training and serves as an accurate proxy for the pre-training data cutoff year across LMs. The ACY has a structured mechanism linking year to verb tense, and resists modification by prompting or fine-tuning. ACY-specific weight editing is effective but does not generalize to the DCY. The DCY, by contrast, is a shallow representation: it is only learned during post-training, lacks a consistent internal mechanism, and can be easily shifted via prompting, fine-tuning, or DCY-specific weight editing.

The contrast between these notions reveals a broader pattern: concepts deeply ingrained in language structure and baked in through pre-training, like the associative link between year and verb tense, are more resistant to post-hoc modification than surface-level representations like the DCY. This has practical consequences for refreshing LMs: because the two notions are decoupled, no existing method succeeds in updating them simultaneously.

%% file: appendix.tex
\newpage


\section{Associative behavior on more LMs, prompts and output verbs}
\label{app:othermodels_verbs}
We test the associative task on six LMs to test whether the pattern seen in Figure~\ref{fig:associative_task} extends to more LMs. These can be seen in Figure~\ref{fig:associative_behavior_more_models}. We find that across all six LMs, LMs' response to the associative task coherently shifts from past to present+future.

\input{figures/associative_behavior_more_models}

We additionally test the associative task on other prompts and verbs such as \textit{In [year], with his credit card, he} and \textit{In [year] the choir}. For these prompts, the exact verbs to collect tense probabilities of is less clear, and choosing just one set of output verbs doesn't cover much of the probability distribution. Instead, we use a comprehensive list of past and present+future verb tenses via a publicly available JSON file from~\citet{jakubczyc2017verbforms} which contains verbs from the Oxford Advanced Learner's Dictionary 9th edition and Longman Dictionary of Contemporary English 6th edition. For each verb entry like \textit{["appear","appears","appeared","appeared","appearing"]} in that file, we categorize the first two forms as present+future tense and the third as past tense. We exclude the fourth and fifth forms since they are participles and require auxiliary verbs. We only include verbs where all relevant forms are single tokens, which prevents imbalanced probability collections between past and present $+$ future tense categories. We also remove ambiguous verbs that appear in both categories to prevent double counting 
(e.g., \textit{read} and \textit{hit}, which have identical past and present forms).

In Figure~\ref{fig:associative_behavior_more_verbs} we again see a coherent shift from past to present+future, indicating that the associative task is robust to exact prompts and output verbs used.

\input{figures/associative_behavior_more_verbs}

\section{Alternative ACY setup with tense before year}
\label{app:alt_acy_prompt}
In the ACY prompt \textit{In {year} there}, year is presented before tense which implicitly favors a lookup/co-occurrence hypothesis. We investigate whether the associative shift from past to present tense still holds under an alternative setup where the tense marker appears \emph{before} the disambiguating year.

\paragraph{Setup} We use the 12 sentence templates in Table~\ref{tab:alt_acy_sentences}, e.g.\ \textit{The mood [was/is] celebratory in [year]}, each instantiated in a past and a present form. For a given year, we score every full sentence by summing its per-token log-probabilities under the LM, and sweep the year over 1900--2250. Per year, we average the log-probabilities across the 12 templates to obtain a mean past and mean present log-probability. We compare only the past and present forms, which are token-matched minimal pairs (the sentences differ only in \textit{was}$\leftrightarrow$\textit{is} or \textit{were}$\leftrightarrow$\textit{are}). We exclude the future form (\textit{will be}) because it adds an extra token to the sentence: since total-sentence log-probability decreases with each additional token, comparing the future form against the past and present forms would be confounded by sentence length, whereas the past-vs-present comparison is not.

\paragraph{Results} As can be seen in Figure~\ref{fig:minimal_tense_sentence_panel}, for all LMs tested, the log-probabilities of both the present and past forms rise toward the training data cutoff and drop sharply at it, meaning the co-occurrence hypothesis on ACY is robust to the order of year and tense. We leave futher exploration of this method to future work.

\input{tables/alt_acy_sentences}
\input{figures/minimal_tense_sentence_panel}

\section{Implicit contextual association via factual recall}
\label{app:associative_factual_recall}
The ACY's participation in temporal reasoning could be triangulated through other tasks that convey the year implicitly rather than stating the year. Here we test an \emph{implicit contextual} variant of the associative task, where the year must be recovered through factual recall of a named event rather than read directly from the prompt.

\paragraph{Setup} We prompt LMs with 29 phrases tied factually to specific years, such as \textit{During the Edward Snowden NSA surveillance revelations, there} (2013) for past events and \textit{During the Winter Olympics in Milan and Cortina, there} (2026) for future events. We use 2--3 prompts per year over the range 2010--2028 (full list in Table~\ref{tab:context_association_prompts}). As in the main associative task, we read off the next-token verb tense, grouping \textit{was, were} as past and \textit{is, are, will} as present+future. We evaluate the base and SFT versions of \texttt{OLMo2-1B} and \texttt{OLMo2-7B}.

\paragraph{Results} If LMs recalled each event's date, we would expect present+future tense probability to increase for events near and after the training data cutoff, mirroring the shift in the main associative task (Figure~\ref{fig:associative_task}). Instead, as shown in Figure~\ref{fig:context_acy_no_year}, all four LMs show no clear increase in present+future tense probability nor a decrease in past tense probability over time. Past tense dominates even for future events, and the SFT models are more strongly past-dominated than their base counterparts.

We note a fundamental limitation of this style of factual evaluation: by definition, we cannot construct prompts about events in the far future, since such facts simply do not exist yet. This makes factual recall unsuitable for measuring whether an LM can be shifted toward future years, which is a key capability we evaluate. The ACY task does not suffer from this constraint because verb tense alone carries temporal information without needing future-specific facts. More broadly, this style of experiment is also harder to scale than the ACY task since prompts must be manually curated per year, and LMs encounter verb-tense relationships far more densely in pre-training than any single historical fact.

\input{tables/context_association_prompts}
\input{figures/context_acy_no_year}

\section{Factual recency as an implicit current-year probe}
\label{app:factual_recency}
The ACY recovers the current year from a single grammatical signal, verb tense (Section~\ref{section:associative}). It is therefore natural to ask whether the current year is recoverable from an entirely different, non-grammatical signal: a model's factual knowledge of when recurring events last occurred. Here we probe this via \emph{factual recency}. We ask the LM for the \emph{most recent} instance of a recurring event and read the year implied by the answer it names. For example, naming ``PyeongChang'' as the most recent Winter Olympics host implies a current year of 2018--2021. This differs from the tense-based factual-recall probe of Appendix~\ref{app:associative_factual_recall}: there we read verb tense off an event prompt, whereas here we read the implied year directly from the named answer. It lets us ask whether a model's factual knowledge yields a consistent estimate of the current year, and how that estimate relates to the training-data cutoff. Because the year must be recovered through recall of named events, this probe has two inherent limitations, shared with the tense-based probe of Appendix~\ref{app:associative_factual_recall}. First, we cannot construct prompts about events in the far future, since such facts do not yet exist; factual recency therefore cannot test whether an LM can be shifted toward future years, a key capability we evaluate elsewhere in the paper. Second, the probe is far more labor-intensive than the ACY task, and thus harder to scale: prompts must be curated by hand per fact, and every response must then be graded manually.

\paragraph{Setup} We curate 18 recurring facts whose ``most recent'' answer is well defined and changes over time, such as the host city of the most recent Summer Olympics or the most recently inaugurated US President (full list in Table~\ref{tab:factual_recency_facts}). To reduce sensitivity to any single wording, we write three phrasings per fact; each phrasing is realized both as a sentence-completion (continuation) prompt for base LMs and as a matching question prompt for SFT LMs, giving $18\times3\times2=108$ prompt templates. Each model is evaluated on the 54 (fact, phrasing) items in its own form. For every response we hand-annotate the entity named and the range of years over which that entity is the correct ``most recent'' answer, counting a year as covered whenever the entity is the correct answer on at least one day of that year. We label a response \emph{current} if its range covers the training-data cutoff ($\sim$2023 for \texttt{OLMo2} LMs), \emph{behind} if it names a real but older answer, \emph{hallucinated} if it names no valid title-holder, and \emph{refused} if the LM declines to answer. We evaluate the base and SFT versions of \texttt{OLMo2-1B} and \texttt{OLMo2-7B}.

\paragraph{Results} If factual recall encoded a coherent current year, a model's answers would imply a single, consistent year clustered on the cutoff. Instead, as shown in Figure~\ref{fig:factual_recency_ranges}, no model produces a consistent implied year across the 18 facts. Base LMs give the current answer for only 4\% of prompts and a real but older answer for 81\%, hallucinating on the remaining 15\% and never refusing. SFT LMs instead refuse the majority of prompts (54\%) and hallucinate rarely (5\%); when they do answer, their implied years have lower variance and sit closer to the cutoff (25\% current). In contrast to the tense-based ACY, which recovers the cutoff to within ten months (Table~\ref{tab:associative_current_model_cutoff}), factual recency badly underestimates and scatters the implied year, reinforcing the finding of Appendix~\ref{app:associative_factual_recall} that factual recall is not a reliable route to the current year.

\input{tables/factual_recency_facts}
\input{figures/factual_recency_ranges}

\section{Obtaining perplexities on the WIKISPAN dataset}
\label{app:dated_data}
\citet{cheng2024dateddatatracingknowledge} estimate the effective knowledge cutoff date of a LM by measuring perplexity across versions of data over time. For the first eight models listed in Table~\ref{tab:associative_current_model_cutoff}, we extract the effective cutoff predictions directly from Figure~4 of their paper by identifying the month and year that achieves the lowest relative perplexity. The remaining five models are not covered in their analysis; for these, we replicate the WIKISPAN dataset construction and perplexity measurement using the publicly available code accompanying their paper. The resulting plots, shown in Figure~\ref{fig:perplexity_predictions}, are consistent with the format of their original Figure~4.

\input{figures/perplexity_predictions}

\section{N-gram and co-occurrence models}
\label{app:td_counting_ngram_cooccur}

\paragraph{Creating N-gram and co-occurrence models}

We create an N-gram and co-occurrence model from training data counts of both olmo-mix-1124 and dolmino-mix-1124, which are used in stage 1 and 2 of pre-training base models \texttt{OLMo2-1B} and \texttt{OLMo2-7B}. We detail the creation of these models here.

\paragraph{Co-occurrence model} We split the training data into sequences, as is done when training, then further split these sequences into sentences using the nltk python library and additionally split sentences longer than 20 words. This additional splitting is necessary because the training data contains sections without punctuation that extend for hundreds of words, which would cause inaccurate co-occurrence counts if the first and last words of such long passages were considered related. We then look for co-occurrences of the string \textit{in [year]} and instances of \textit{was}, \textit{were}, \textit{is}, \textit{are} and \textit{will} anywhere within each sentence and collect counts of these combinations. Finally, we apply Laplacian smoothing with an alpha of 1.0 to create the co-occurrence model.
\paragraph{N-gram model} We split the training data into sequences, as is done when training. We then look for instances of \textit{was}, \textit{were}, \textit{is}, \textit{are} and \textit{will} directly following the string \textit{In [year] there} (to match how we prompt the LM) using a case insensitive regex that requires words be split by word boundaries (\texttt{\textbackslash b}) in the training corpus. We combine \textit{was} and \textit{were} as past tense and \textit{is}, \textit{are} and \textit{will} as present+future tense, then apply Laplacian smoothing with an alpha of 1.0 to create the n-gram model.

\paragraph{N-gram model, co-occurrence model, and \texttt{OLMo2-1B}/\texttt{OLMo2-7B} associative year-tense behavior}
We show the normalized tense predictions for the n-gram model, co-occurrence model, and \texttt{OLMo2-1B}/\texttt{OLMo2-7B} in Figures~\ref{fig:relative_comparison_stage1} and~\ref{fig:relative_comparison_stage2} for stage 1 and 2 training comparitively. We use these in Section~\ref{sec:associative_data_dist} to compute the CE.

\input{figures/associative_data_count_stage1}
\input{figures/associative_data_count_stage2}

\paragraph{Cross entropy between \texttt{OLMo2 LMs} and n-gram and co-occurrence models} 
\label{app:paragraph:crossentropy}

Figure~\ref{tab:cross_entropy_comparisons} shows the cross entropy between \texttt{OLMo2} LMs and models constructed from pre-training data at Stage 1 and Stage 2. We see the LMs have a lower cross entropy with the count-based models than with a random baseline, suggesting the ACY is shaped by the pre-training data distribution.

\input{tables/ngram_cooccur_CE}

\section{Controls and robustness for the associative year-type subspace}
\label{app:detailed_subspace_analysis}

\paragraph{Details on DAS procedure} Here, we offer details on the standard DAS procedure used in Section~\ref{subsubsec:caus_abs}. We construct 20480 training pairs and 1024 evaluation pairs of base and source examples using prompts of the form \textit{In [year], there}, where the year ranges from 1000 to 4000. We keep correct prompts where the LM naturally predicts the expected tense (past for years before the training data cutoff and present or future after). On correct prompts, we know models can correctly reason about the prompt year with respect to its current year, which suggests a year-type subspace is being used. We train and evaluate on correct prompts with no overlap between sets to remove bias.

We further analyze the one-dimensional ``year-type'' subspace whose interchange interventions shift tense. Here, we provide sanity checks demonstrating that this effect is non-trivial (i.e., not an artifact of the evaluation setup) and that it is specific to year-like representations.
\paragraph{Random subspace} As a baseline, we replace the DAS-learned one-dimensional subspace with a randomly chosen one-dimensional subspace at the same layer and token position. Table~\ref{tab:detailed_subspace_analysis} shows that this yields an IIA of \(14\%\) across layers, matching the percentage of evaluation pairs in which the target tense does not change. Thus, interchanging an arbitrary subspace does not systematically control the year-tense mechanism.
\paragraph{Randomly initialized model} Next, we test whether DAS can obtain high IIA even in the absence of learned structure in model weights. We randomly initialize the weights of \texttt{OLMo2-7B}, run DAS to fit a one-dimensional subspace using the same training procedure, and evaluate on the same evaluation set used for the year-type experiments. This achieves \(0\%\) IIA across layers, indicating that high-IIA solutions depend on learned representations rather than the DAS procedure alone.
\paragraph{Interchanges with non-numeric, non-year entities}
We then evaluate the learned year-type subspace out of distribution. We let base prompts remain of the form \textit{In [year], there}, while source prompts contain non-year entities such as \textit{In summary, there} or \textit{In Amsterdam, there} (full prompt list in Table~\ref{tab:associative_interchange_prompts}). Since all these source prompts elicit present tense predictions, we restrict base prompts to those that elicit past tense predictions. Under these interchanges, IIA is mostly \(0\%\), peaking at \(46\%\)) in some layers (which is below random guessing), suggesting the localized subspace does not support tense control when the year subspace is filled by unrelated non-numeric entities.
\paragraph{Interchanges with four-digit numbers that are not used as years}
We again reuse the learned year-type subspace, but now the source prompts contain year-like numbers in non-temporal contexts, specifically math. For example, for a prompt \textit{732 - 1 + 1789 is}, we perform interchanges at the number token(s) (e.g., on the \textit{9} tokens). In this setting, interchanges succeed in early layers, reaching \(>\!90\%\) IIA in five layers (Table~\ref{tab:detailed_subspace_analysis}). This indicates that the localized subspace generalizes to numbers that resemble years even when they are not explicitly used as temporal expressions, This effect disappears in later layers, suggesting that the year-type then becomes specific to temporal contexts.
\paragraph{Interchanges under a different template and output tokenization}
Finally, we test robustness to changes in surface form and output casing by changing the original template to uppercase: \textit{IN [year] THERE}. This shifts the target next-token predictions from \textit{was, were, is, are, will} to \textit{WAS, WERE, IS, ARE, WILL}. One complication is that \textit{WERE} can be tokenized as \textit{W} + \textit{ERE}; in such cases, the LM assigns most probability mass after \textit{W} to \textit{ERE}. We therefore treat \textit{W} as the uppercase realization of \textit{were} for evaluation. With this modification, interchanges remain largely successful but achieve consistently lower IIA than the lowercase setting across layers (Table~\ref{tab:detailed_subspace_analysis}), suggesting the localized subspace is not merely tied to a single fixed output token but is somewhat sensitive to template and output token used.
\input{tables/detailed_subspace_analysis}
\input{tables/associative_interchange_prompts}

\section{Causal tracing through attention and MLP knockouts}
\label{app:associative_knockouts}
In Section~\ref{subsubsec:caus_abs}, we use learned interchange interventions on the residual stream to probe how the LM solves the associative task. Here, we use null interventions, replacing attention and MLP additions with a zero vector to measure their causal contribution to the final tense prediction.

\paragraph{Methods}
\label{para:sublayer_knockout_methods}
We use attention/MLP sublayer knockouts~\citep{geva-etal-2023-dissecting}: for each intervention site, we zero out either the attention or MLP output over five consecutive layers on one token. For each intervention, we measure a change in probability of tense before and after the knockout by measuring tense shift magnitude (TSM) as

\begin{eqnarray}
\mathrm{TSM}_y
&=& |p_y(\mathrm{past})_{\mathrm{before}}
   - p_y(\mathrm{future})_{\mathrm{before}} \nonumber\\
&&\quad - \bigl(
   p_y(\mathrm{past})_{\mathrm{after}}
   - p_y(\mathrm{future})_{\mathrm{after}}
   \bigr)| \nonumber
\end{eqnarray}

\paragraph{Causal tracing corroborates the year-type localization results}
We do attention and MLP knockouts across 70 prompts on which the LM behaves correctly. The average TSM across each group can be seen in Figure~\ref{fig:mech_analysis_associative}. We see that the largest tense shifts occur when we knock out computations on earlier layers of year tokens and on the later layers of the final token, which are consistent with the causal abstraction results (Figure~\ref{fig:iia_heatmap}). 

\paragraph{Sublayer knockouts reveal asymmetries across past vs. future years}
Notably, past-year and future-year prompts respond differently: Knockouts for prompts with future years have a larger effect which are seen on the second year token (i.e. the last digit), whereas prompts with past years are less intensely affected by the second token of the year. This pattern suggests that the LM resolves tense using the earliest year-token evidence available, but relies more heavily on late-year digits when the year is interpreted as future.
\input{figures/causal_tracing_associative}

\section{Declarative prompts and behavior}

\label{app:declarative_prompts}
We use a set of prompts to find the declarative current year. The respective set of continuation prompts for base models and instruction prompts for fine-tuned models can be seen in Table~\ref{tab:declarative_prompts}.

It is notable that much effort was placed into selecting a set of ten prompts to which both the base and SFT models for OLMo2-1B and OLMo2-7B responded to with a year.

\input{tables/declr_prompts}

\section{Training data counts of years}
\label{app:declr_tr_data_counts}
Figure~\ref{fig:combined_year_counts} shows the normalized counts of each year on the range 1900 to 2100 in training data. We use olmo-mix ((\texttt{OLMo2} stage 1 pre-training ~\citep{allenai_olmo-mix-1124}), dolmino  (\texttt{OLMo2} stage 2 pre-training \citep{allenai_dolmino-mix-1124}), and tulu3 (\texttt{OLMo2-1B} SFT ~\citep{allenai2025tulu3sftolmo2}, and \texttt{OLMo2-7B SFT~\citep{allenai2025tulu3sftolmo2_7b}}).

We show counts of raw 4-digit numbers appearing in text and numbers matching the regex \textit{In [year]}. We use the frequency counts from the latter method because counts collected with this method are more likely to be from years, as opposed to just 4 digit numbers.

We also show results for an alternate method of counting: Figure~\ref{fig:combined_year_counts} shows the counts of \textit{in [year]} for the training data for years on the range 1900-2100.

\input{figures/in_year_td_counts}

\section{SFT training \texttt{OLMo2-1B}}
\label{app:sft_training}

To test the effects of training data content on the declarative current year, we train SFT on the Tulu3 dataset~\citep{allenai2025tulu3sftolmo2}, which \texttt{OLMo2-1B} uses for their SFT training. We use an effective batch size of 192 (per-device batch size of 4, gradient accumulation of 8, across 6 GPUs), a learning rate of 5e-5 with linear decay and no warmup, and train for 1 epoch with a maximum sequence length of 1024.

\paragraph{No-Years model} First, we measure the effects of \textit{removing} all training examples with years (i.e. all number $\geq$ 1000) from the SFT training data, preserving 76\% of training examples. We refer to this model as the No-years SFT model.

\paragraph{+/-N year-shifted models} Next, we measure the effects of \textit{shifting} all years in the training data by a constant offset N, and we do not delete any training data. We shift all numbers between [1000-N, 9999-N] by a constant N for N = 200, -15, +/-6, +/-18, and +/-35 (eight models). We choose -15 since it aligns the SFT peak year to 2008, the most frequent year in pre-training. The other shift offsets are chosen to land on a year that is not a round multiple of 5 and equally spaced on either side of 2023. We refer to these year-shifted models by their shift, +/-N. 

\paragraph{In-context models} In addition to the models trained on custom datasets, we produce two in-context models, trained with the system prompts \texttt{"The current year is 2023"} and \texttt{"Today Date: 26 Jul 2023"} respectively. The results reported in Section~\ref{section-5} use the former, evaluated with the system prompt \texttt{"The current year is [target year]"}.

We will release our models and training data.

\section{Declarative causal tracing}
\label{app:declarative_causal_tracing}

\paragraph{Causal tracing on oneshot prompts} The mean magnitude and variance shift when applying causal tracing to declarative prompts, shown in Figure~\ref{fig:mech_analysis_declarative_mean_var}.

\input{figures/causal_tracing_declarative_averages}

\paragraph{Fewshot declarative prompts} Prompts are used for mechanistic analysis, shown in Table~\ref{tab:declr_prompts_structured_current_year} for the current-year prompts and Table~\ref{tab:declr_prompts_structured_today_date} for the today-date prompts.

\label{app:declr_prompts_structured}

\input{tables/declr_prompts_structured}

\section{Training dynamics on \texttt{OLMo2-7B}}
\label{app:training_dynamics_on_olmo_7b}

We compute the ACY and DCY of \texttt{OLMo2-7B} by loading checkpoints across training stages. Figure~\ref{fig:training_dynamics_7b} shows the evolution of both quantities across base pre-training, SFT, RLVR, DPO, and the final Instruct model. As in \texttt{OLMo2-1B} (Figure~\ref{fig:training_dynamics_1b}), the ACY for OLMo2-7B remains stable throughout pre-training and the DCY shifts only at the post-training stage.

\input{figures/7b_training_dynamics}

\section{Prompting fails to shift the ACY across model families and scales}
\label{app:prompting_scaling}
In Section~\ref{section-5} we show that in-context prompting cannot shift the ACY of \texttt{OLMo2-1B}. Here we verify that this conclusion generalizes to a broader set of recent and capable LMs beyond the \texttt{OLMo2} family. Since several of these LMs are too large to fine-tune or weight-edit with our available compute, we extend only the prompting method, which requires no training.

\paragraph{Setup} We extend our analysis to nine additional LMs spanning the Gemma-3 and Qwen-3.5 families. For each LM, we attempt to update the ACY and DCY by specifying the target current year via system prompt, across the range 1900--2250. We report the fraction of the 351 target years for which the predicted year exactly matches the target.

\paragraph{Results} A natural hypothesis is that larger, more capable LMs would be better at using the target year in the system prompt, and so would shift both the ACY and DCY more accurately. As Table~\ref{tab:prompting_scaling_acy} shows, this is not the case: accuracy varies non-monotonically with model size, both across and within model families. While several LMs achieve high DCY shifting accuracy with only system prompting (e.g., \texttt{Qwen3.5-4B}, \texttt{Qwen3.5-9B}, and \texttt{Qwen3.5-27B} all reach 100\%), shifting the ACY remains difficult: the highest accuracy across all new LMs is just 53.3\% (\texttt{Qwen3.5-27B}), with every other LM at or below 40\% and several near zero. In no case does ACY shifting approach the near-perfect DCY shifting achieved by the same LM. This confirms our conclusion that the ACY cannot be reliably shifted via prompting, even for LMs substantially larger and more recent than \texttt{OLMo2}.

%% file: figures/associative_behavior_more_models.tex
\begin{figure}[!htbp]
    \centering

    \begin{subfigure}{.48\linewidth}
        \centering
        \includegraphics[width=\linewidth]{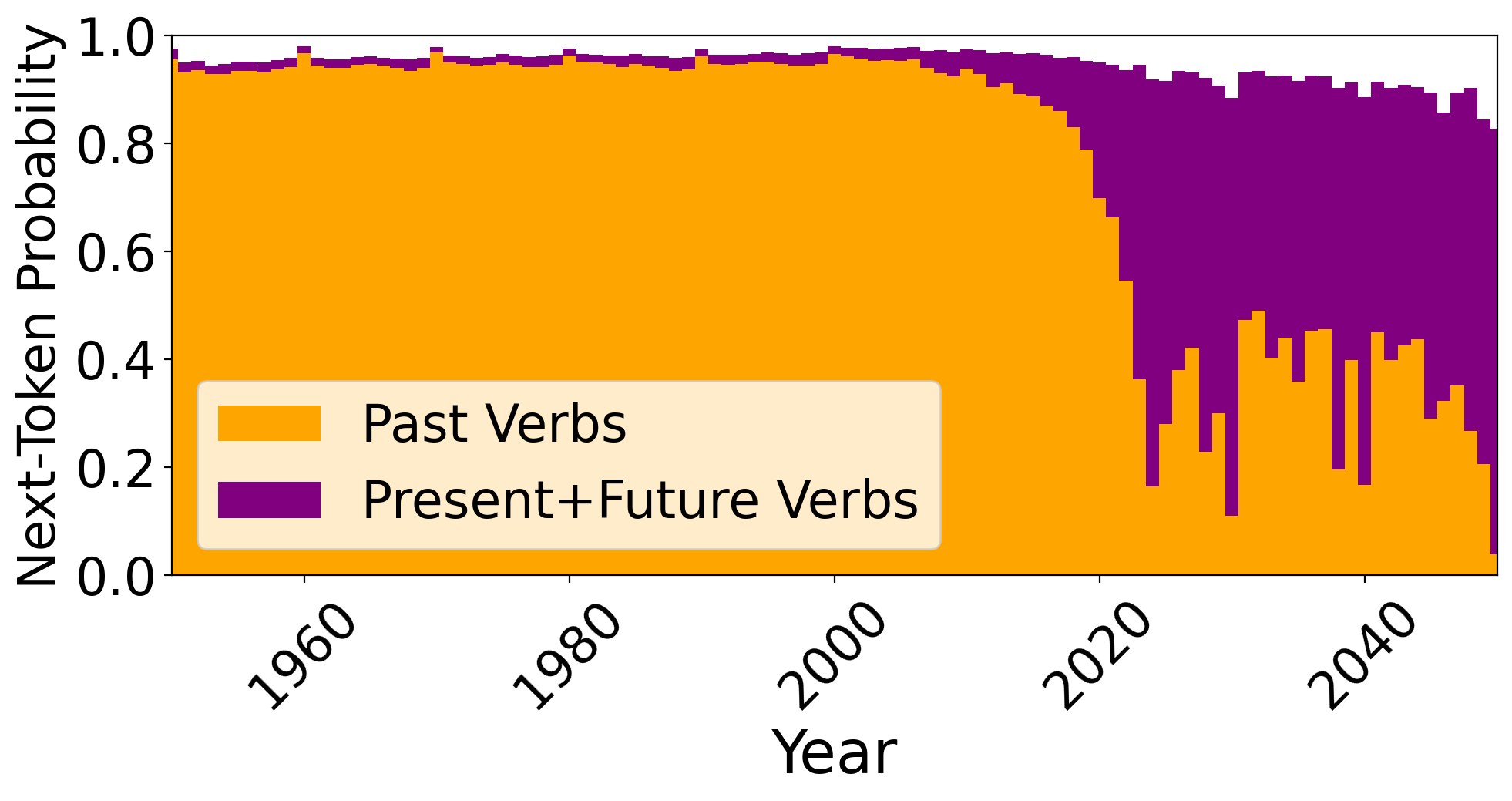}
        \caption{\texttt{OLMo2-1B} \citep{groeneveld-etal-2024-olmo}}
    \end{subfigure}
    \hfill
    \begin{subfigure}{.48\linewidth}
        \centering
        \includegraphics[width=\linewidth]{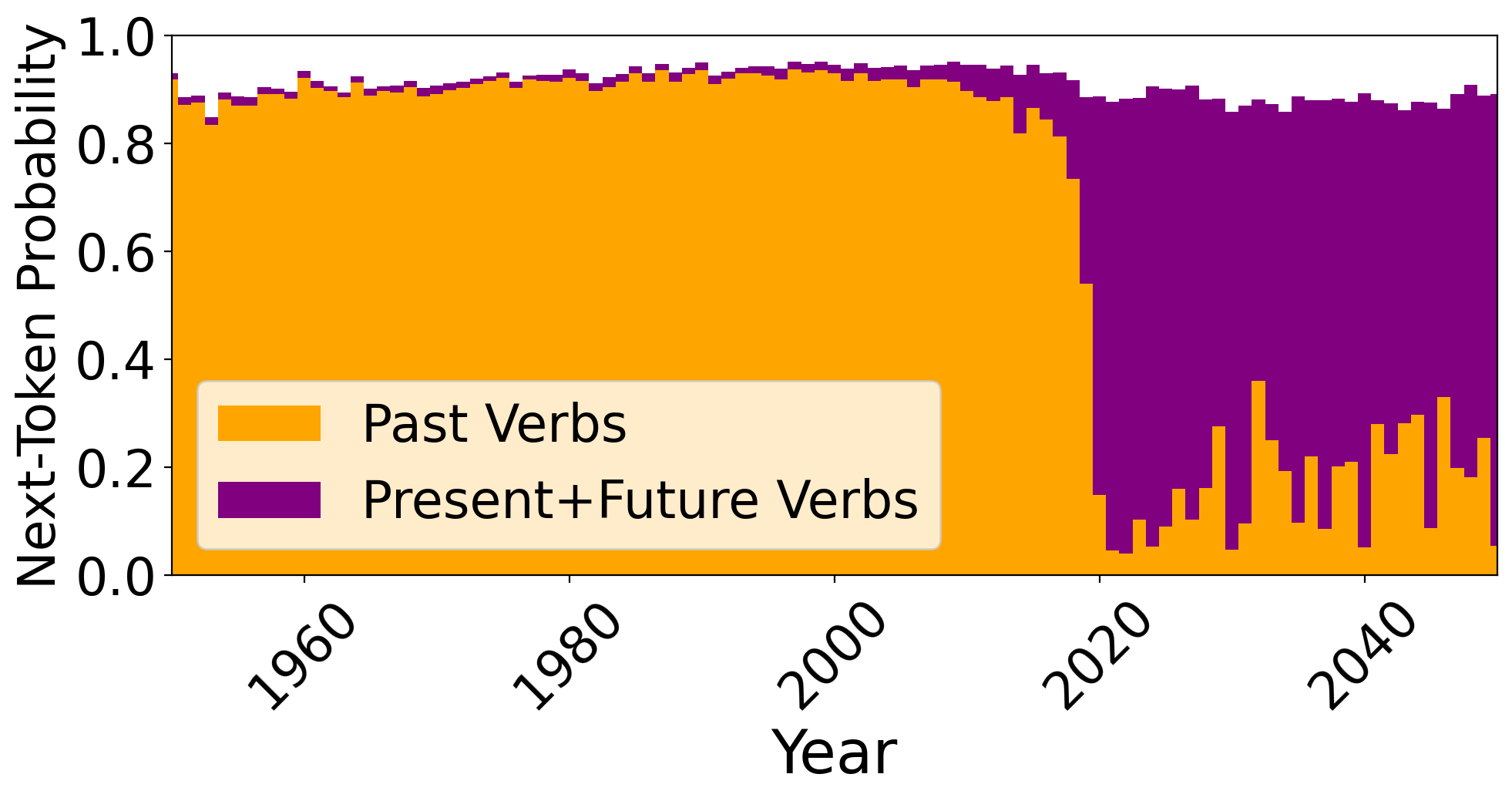}
        \caption{\texttt{Pythia-1.4B-deduped} \citep{biderman2023pythiasuiteanalyzinglarge}}
    \end{subfigure}
    \hfill
    \begin{subfigure}{.48\linewidth}
        \centering
        \includegraphics[width=\linewidth]{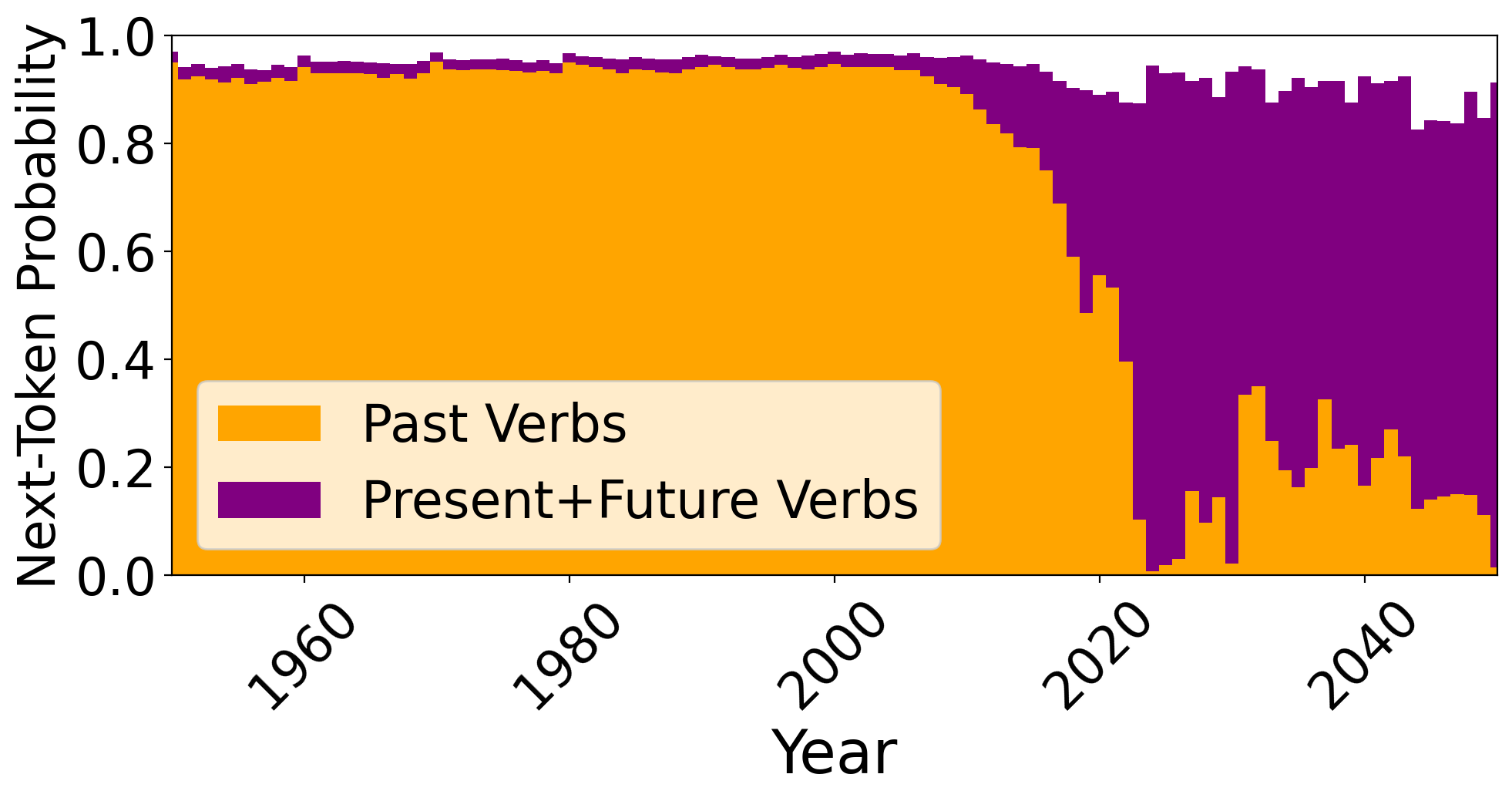}
        \caption{\texttt{Llama3.1-8B} \citep{grattafiori2024llama3herdmodels}}
    \end{subfigure}
    \hfill
    \begin{subfigure}{.48\linewidth}
        \centering
        \includegraphics[width=\linewidth]{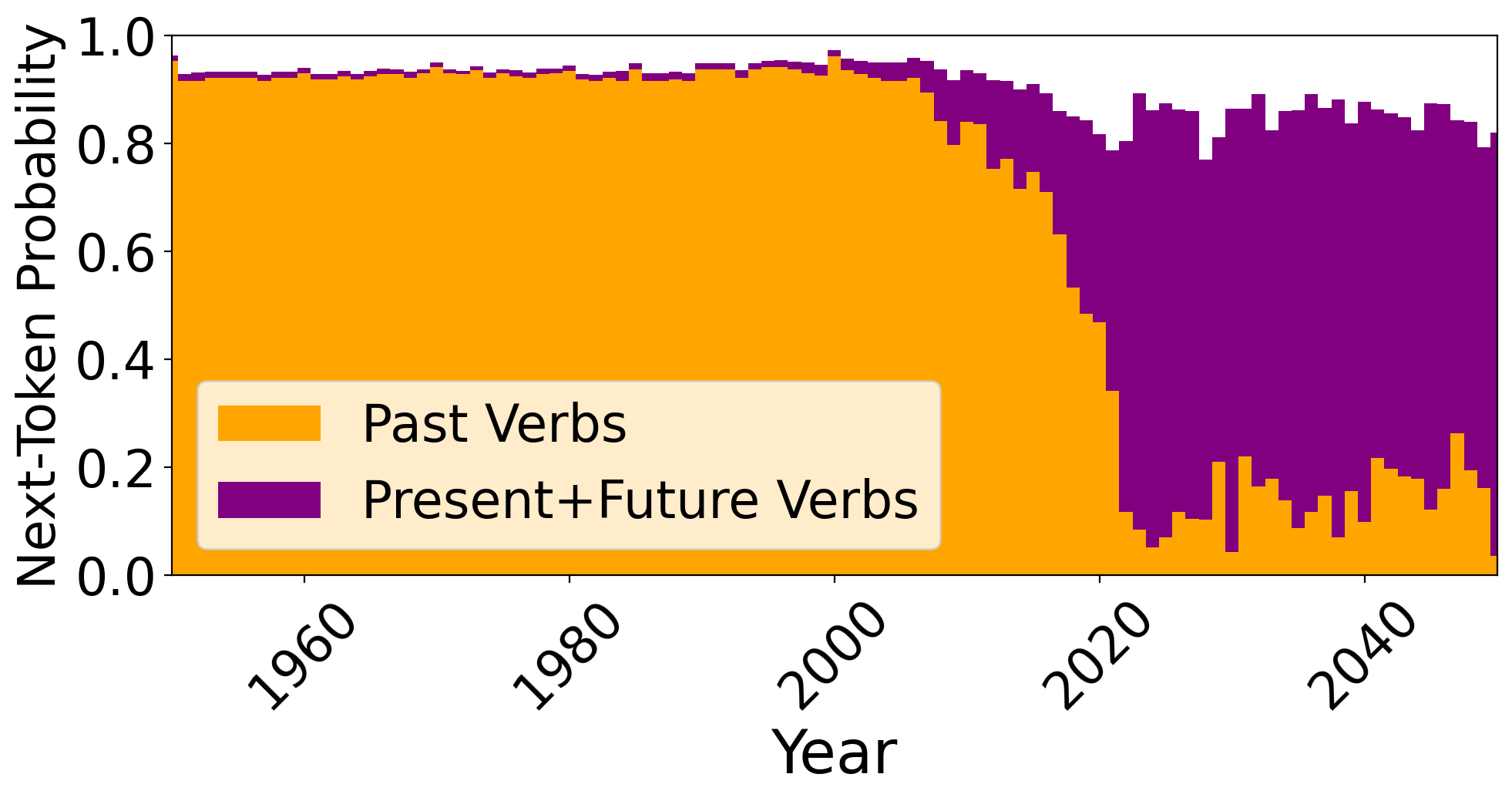}
        \caption{\texttt{Falcon-7B} \citep{almazrouei2023falconseriesopenlanguage}}
    \end{subfigure}
    \hfill
    \begin{subfigure}{.48\linewidth}
        \centering
        \includegraphics[width=\linewidth]{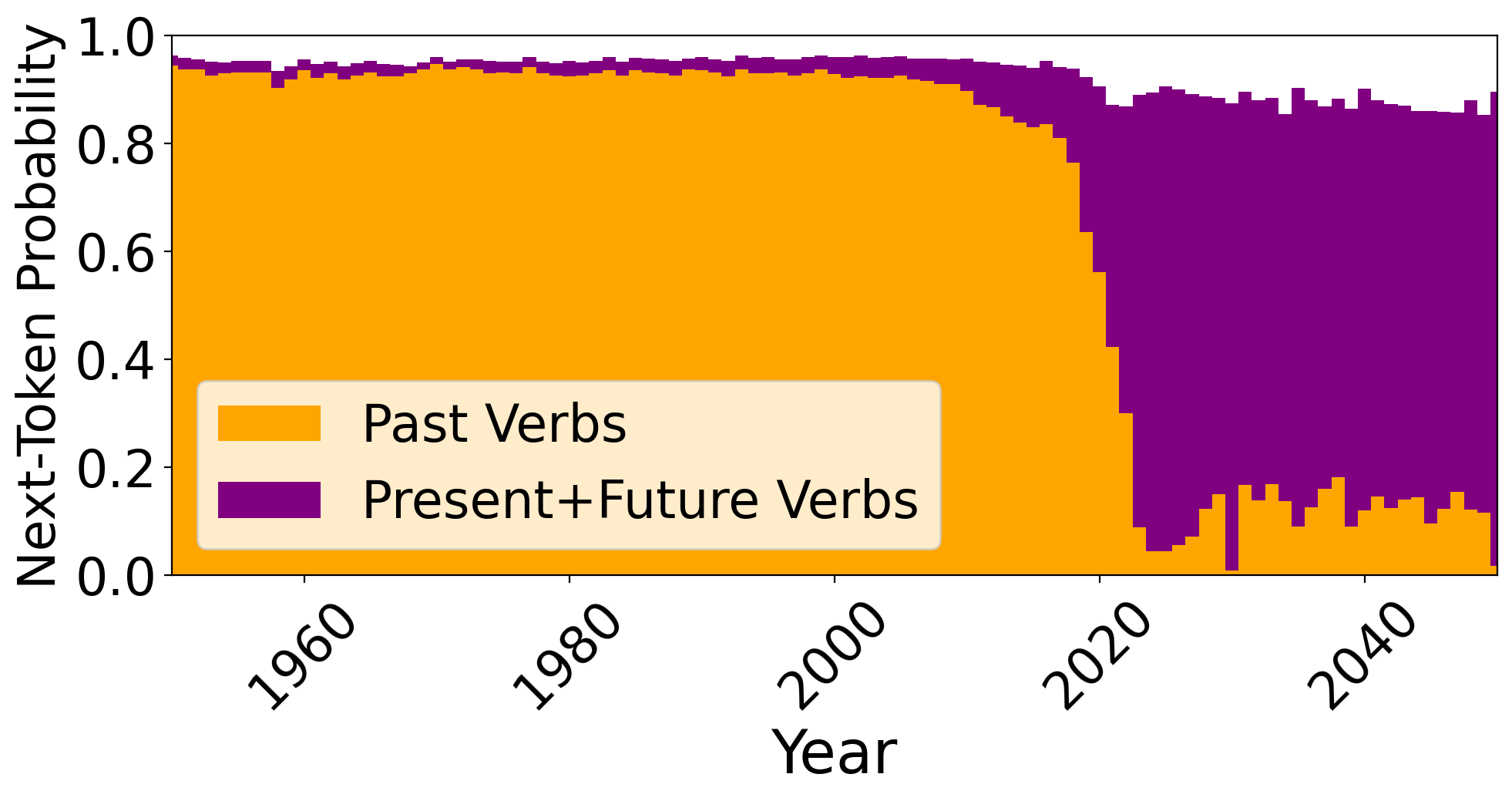}
        \caption{\texttt{RedPajama-7B} \citep{weber2024redpajama}}
    \end{subfigure}
    \hfill
    \begin{subfigure}{.48\linewidth}
        \centering
        \includegraphics[width=\linewidth]{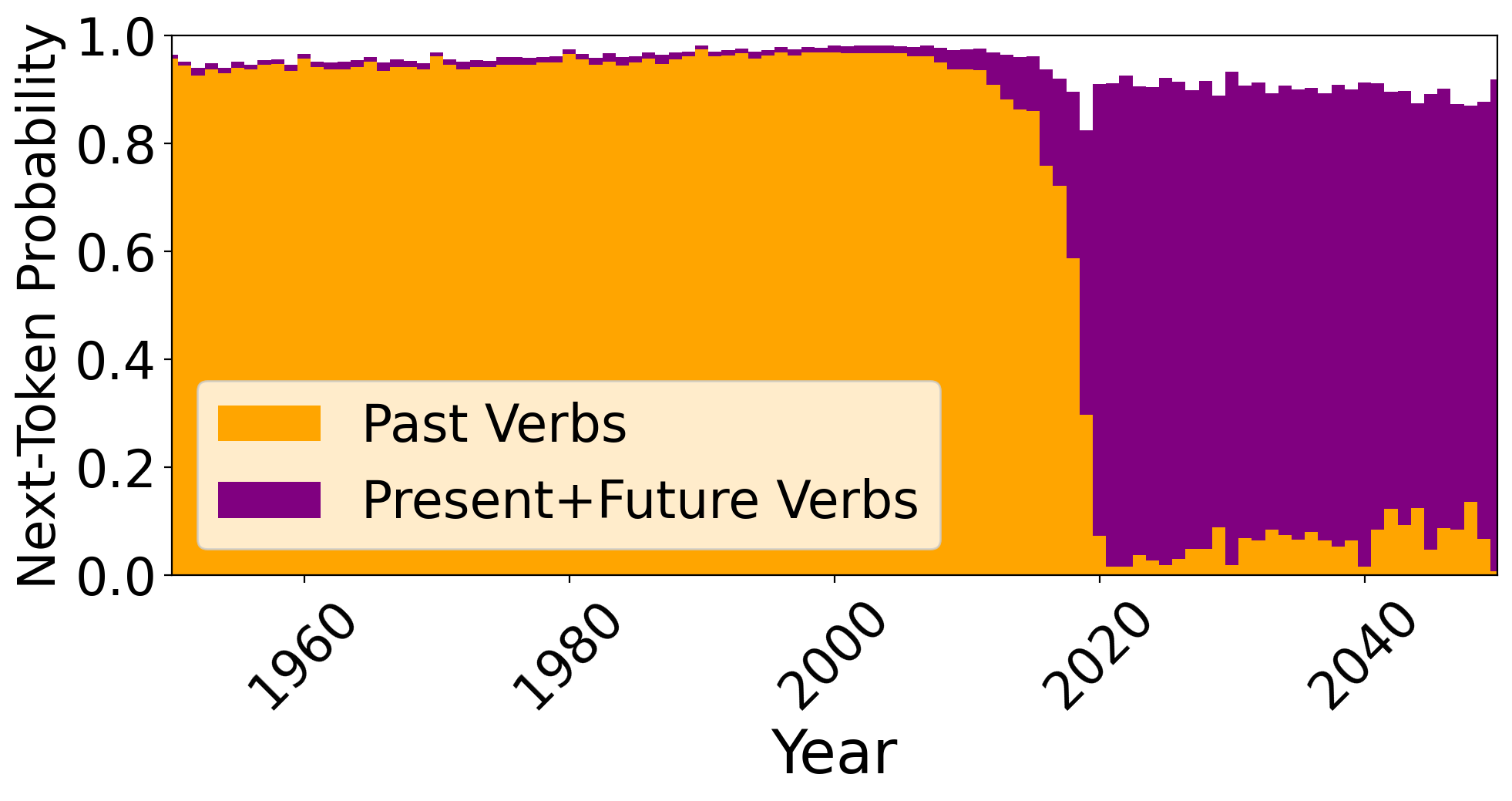}
        \caption{\texttt{GPT-J-6B} \citep{gpt-j}}
    \end{subfigure}
    \caption{\textbf{Various LMs demonstrate a coherent notion of current year.} The prompt used is \textit{In [year] there}.}
    \label{fig:associative_behavior_more_models}
\end{figure}

%% file: figures/associative_behavior_more_verbs.tex
\begin{figure}[!htbp]
    \centering

    \begin{subfigure}{0.48\linewidth}
        \centering
        \includegraphics[width=\linewidth]{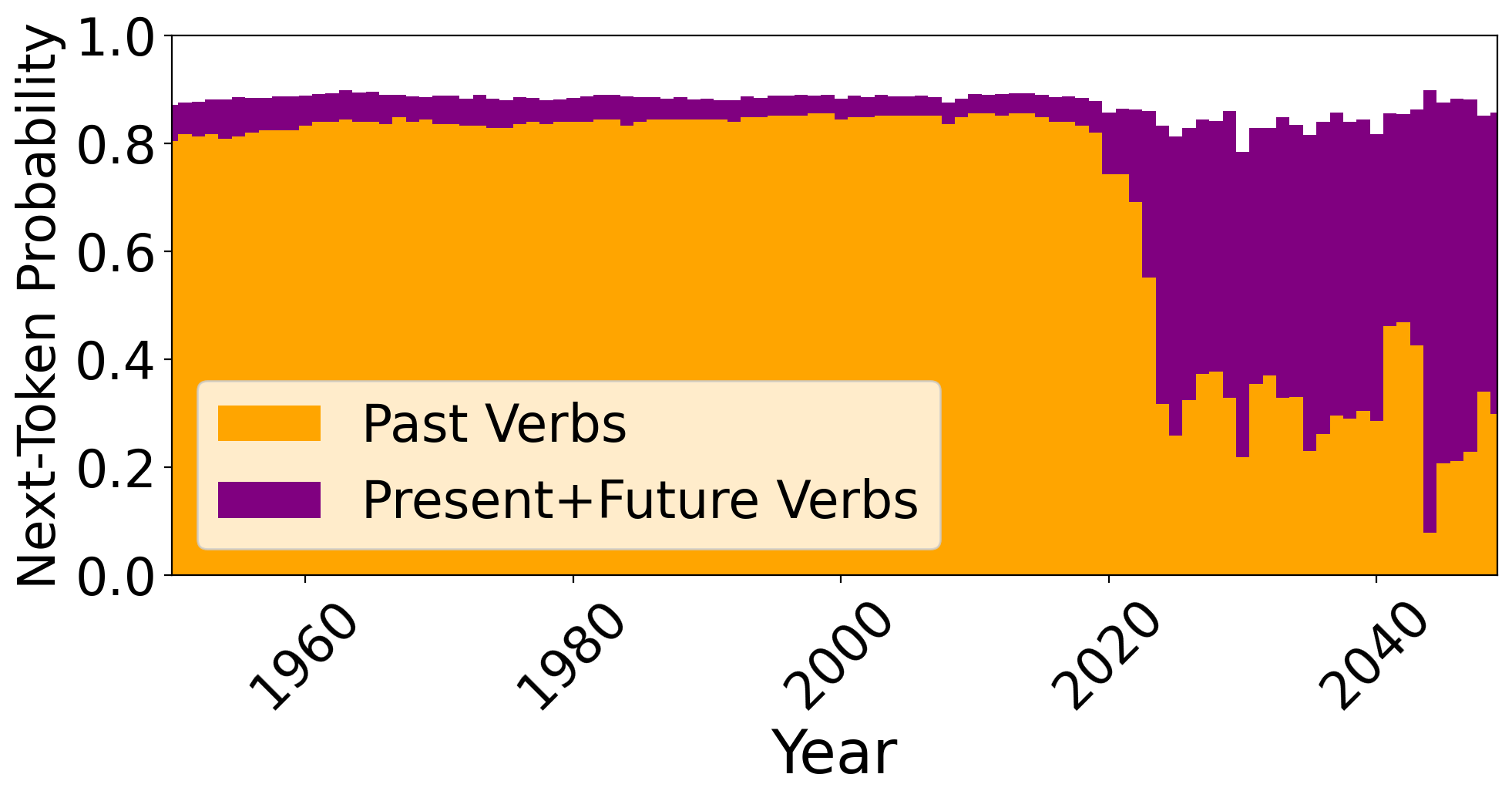}
        \caption{Prompt: \textit{In [year], with his credit card, he}.}
    \end{subfigure}
    \hfill
    \begin{subfigure}{0.48\linewidth}
        \centering
        \includegraphics[width=\linewidth]{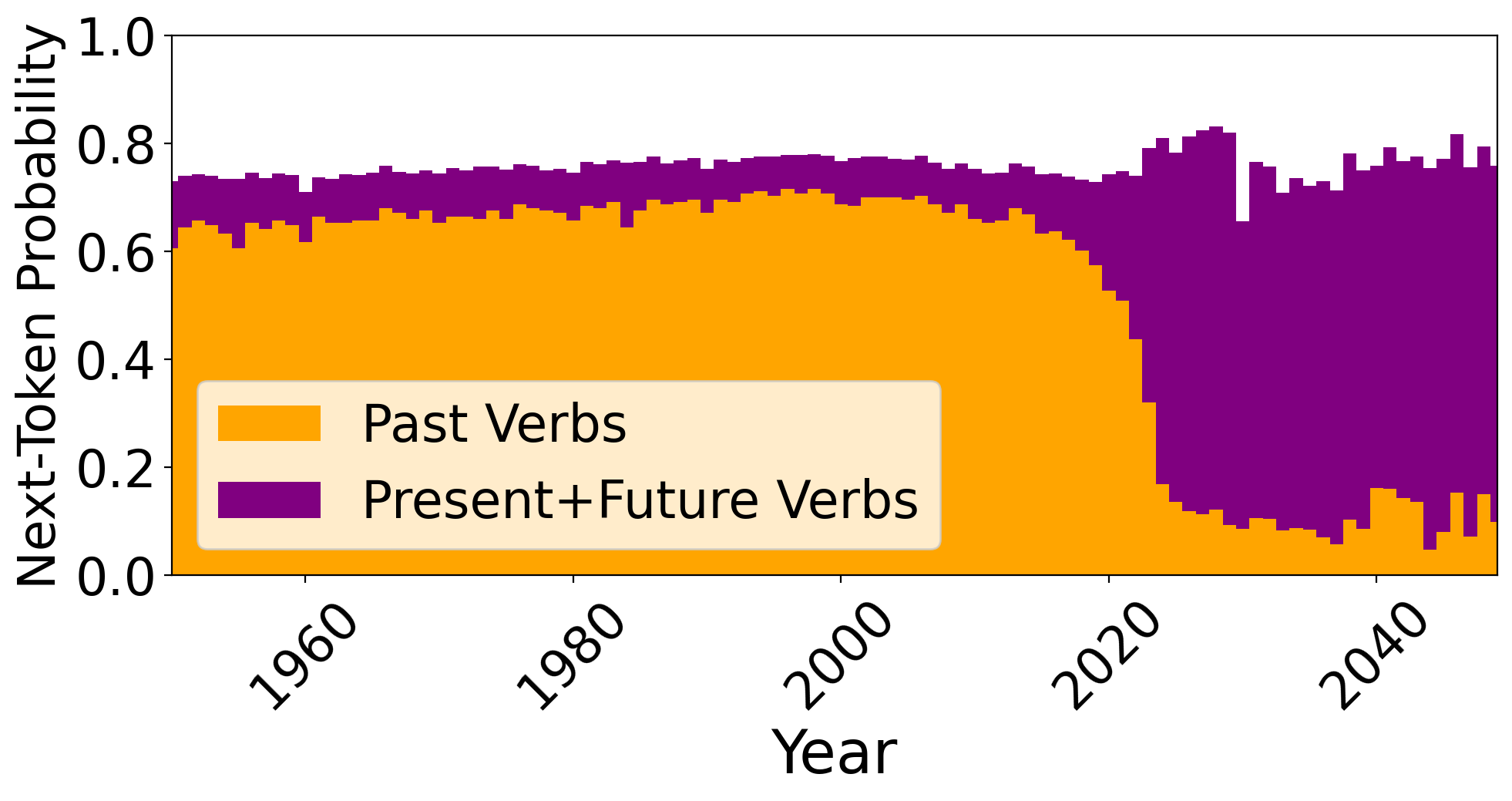}
        \caption{Prompt: \textit{In [year], at the dinner table, the family}.}
    \end{subfigure}
    \hfill
    \begin{subfigure}{0.48\linewidth}
        \centering
        \includegraphics[width=\linewidth]{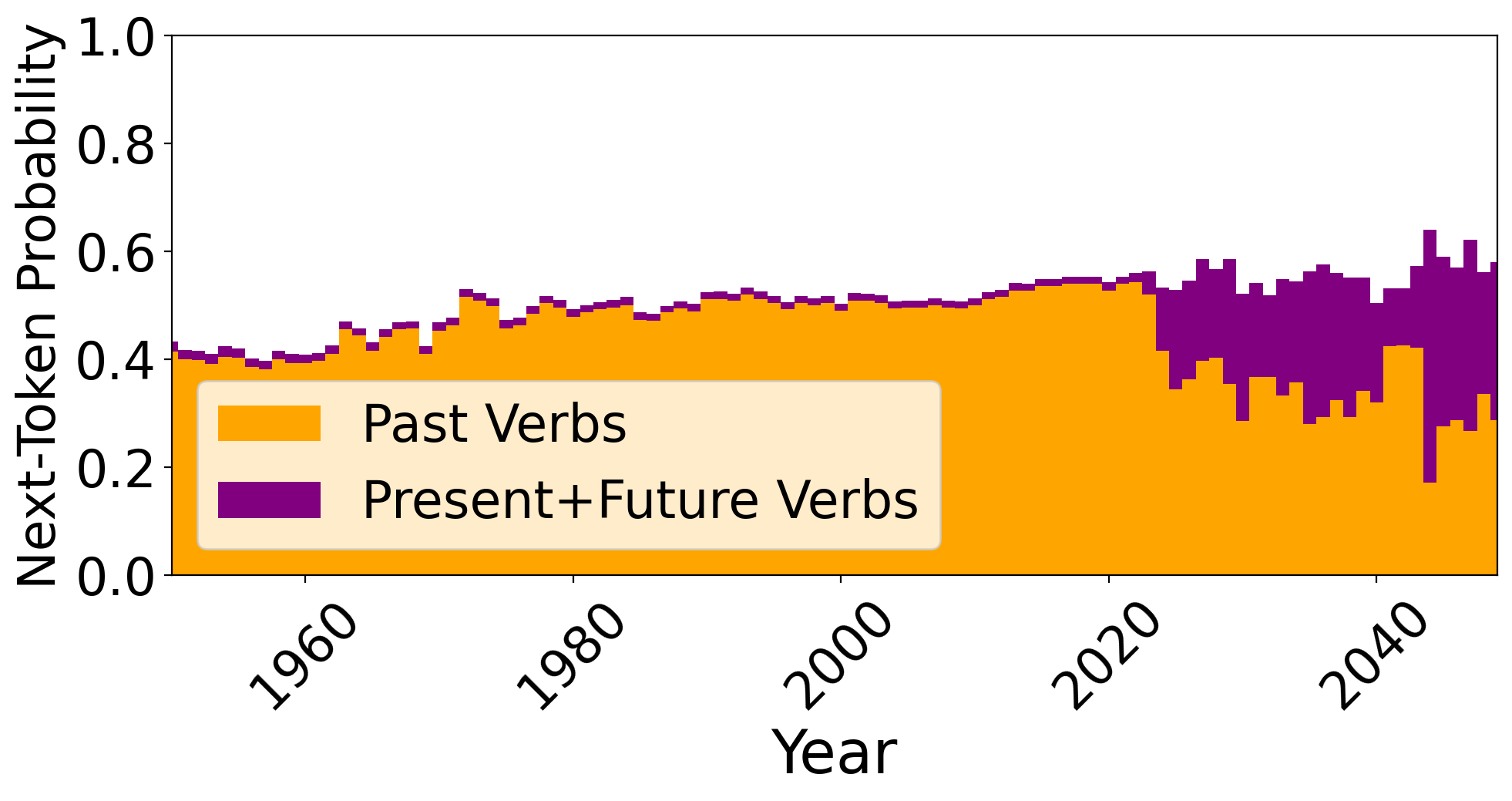}
        \caption{Prompt: \textit{In [year], with a knife, he}.}
    \end{subfigure}
    \hfill
    \begin{subfigure}{0.48\linewidth}
        \centering
        \includegraphics[width=\linewidth]{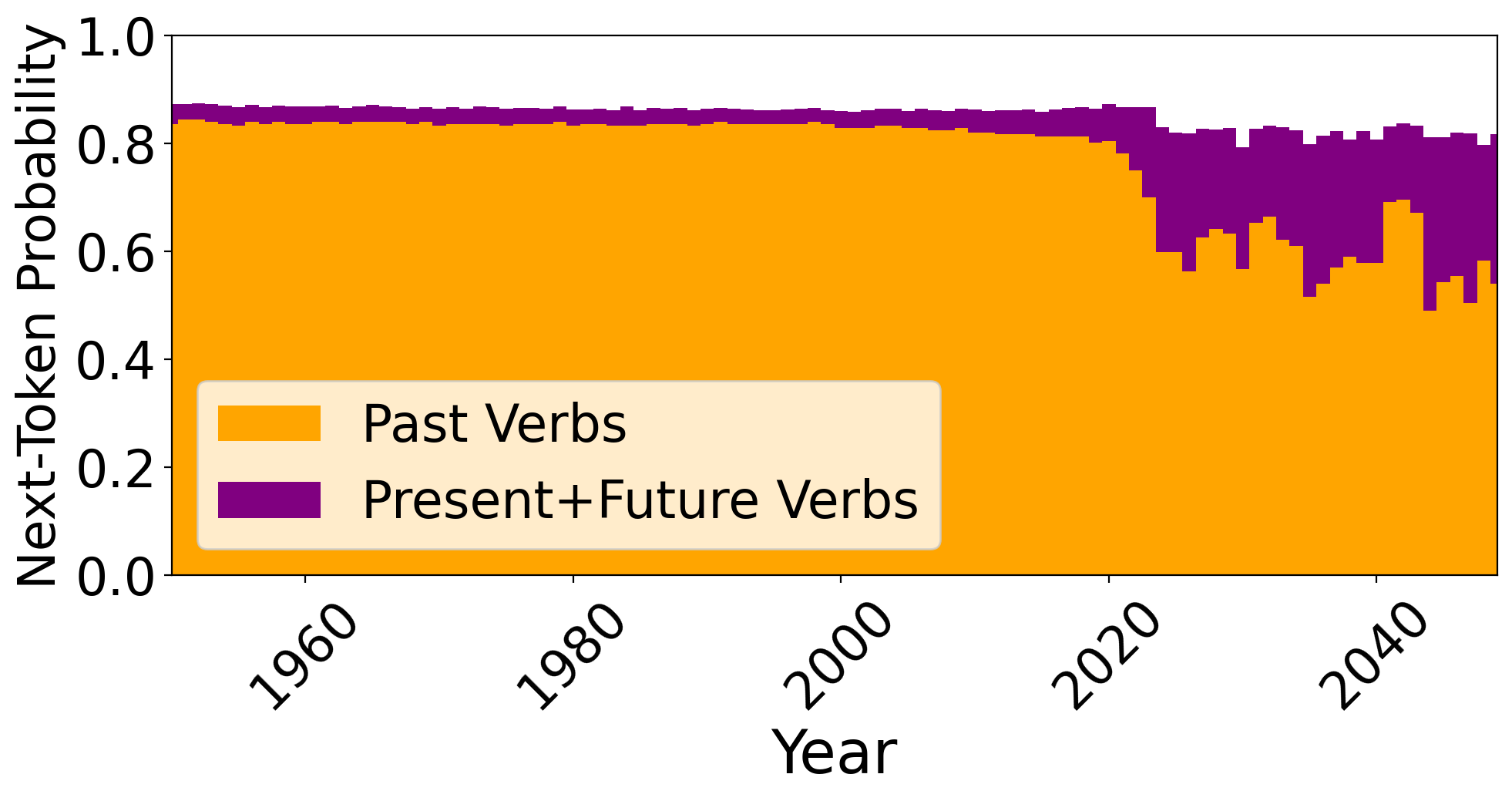}
        \caption{Prompt: \textit{In [year], with a pen to paper, she}.}
    \end{subfigure}
    \hfill
    \begin{subfigure}{0.48\linewidth}
        \centering
        \includegraphics[width=\linewidth]{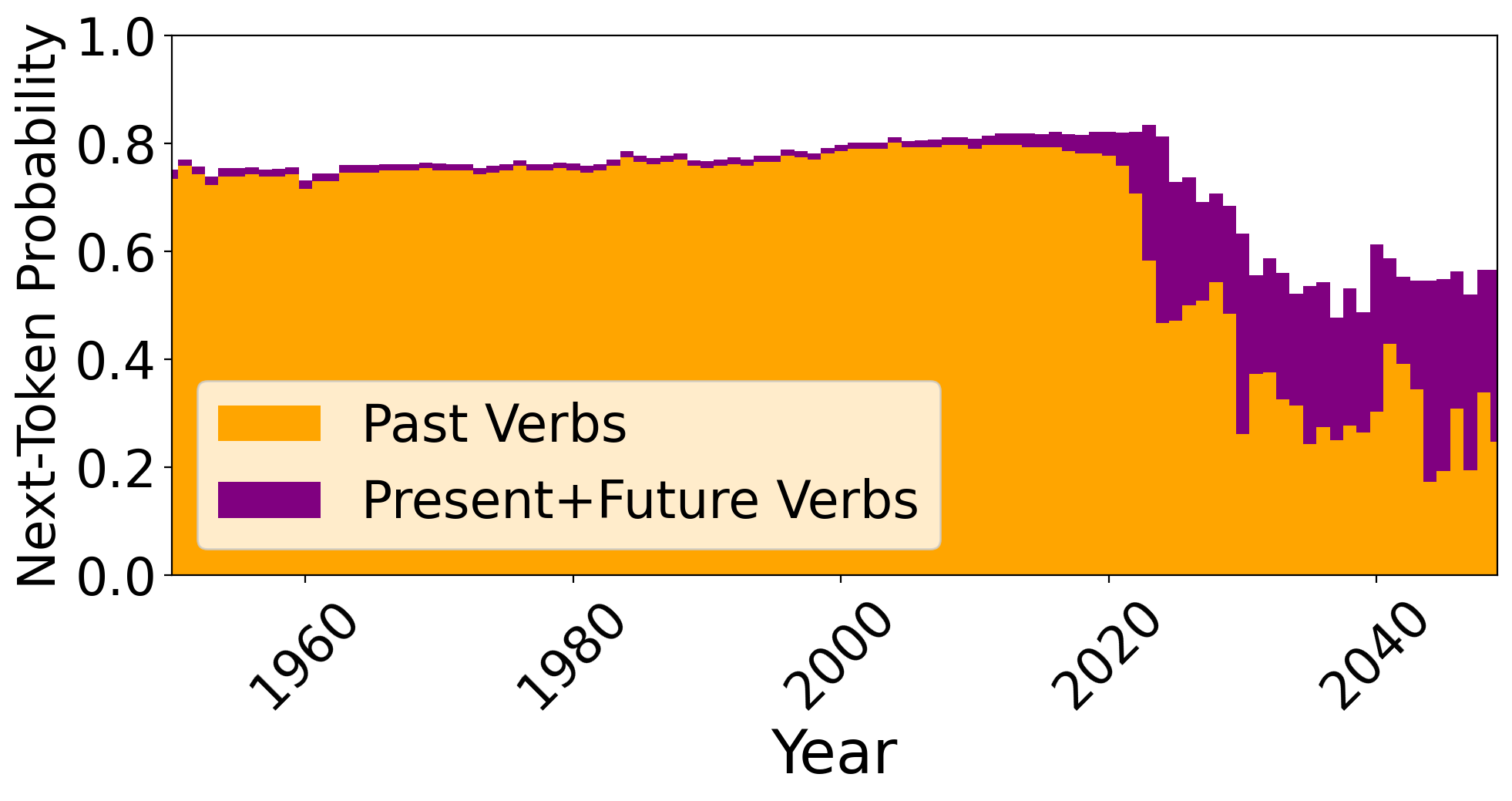}
        \caption{Prompt: \textit{In [year], the choir}.}
    \end{subfigure}
    \hfill
    \begin{subfigure}{0.48\linewidth}
        \centering
        \includegraphics[width=\linewidth]{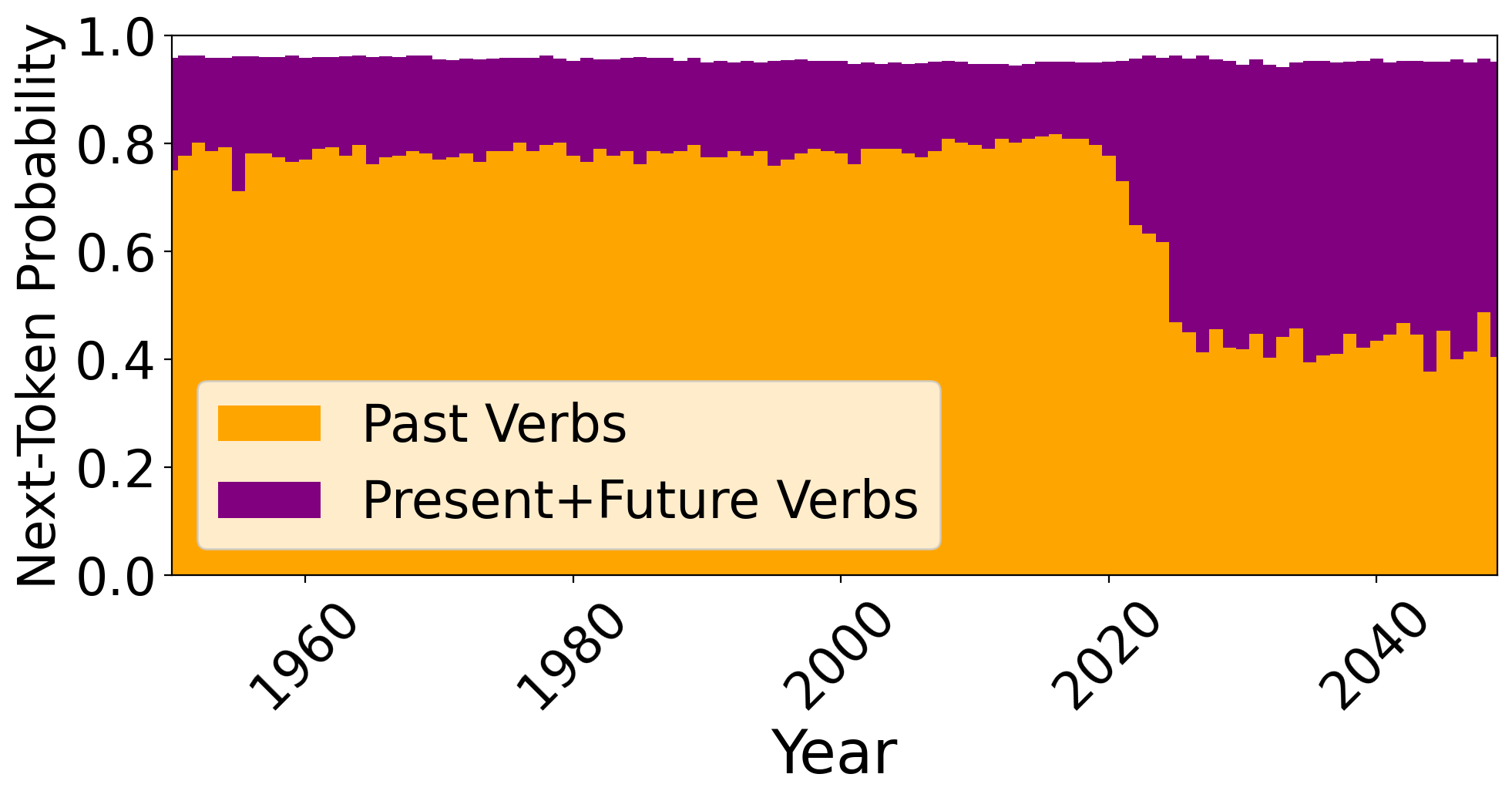}
        \caption{Prompt: \textit{In the magic show in [year], there magically}.}
    \end{subfigure}

    \caption{\textbf{\texttt{OLMo2-7B} demonstrates a coherent notion of the current year on various prompts and output verbs.}}

    \label{fig:associative_behavior_more_verbs}
\end{figure}

%% file: tables/alt_acy_sentences.tex
\begin{table}[!htbp]
\centering
\small
\begin{tabular}{c p{0.8\linewidth}}
\toprule
\textbf{\#} & \textbf{Sentence template (past / present)} \\
\midrule
1 & ``The party [was / is] fun in \{year\}.'' \\
2 & ``The meeting [was / is] productive in \{year\}.'' \\
3 & ``The mood [was / is] celebratory in \{year\}.'' \\
4 & ``The weather [was / is] unusually warm in \{year\}.'' \\
5 & ``The economy [was / is] strong in \{year\}.'' \\
6 & ``The city [was / is] crowded in \{year\}.'' \\
7 & ``The concert [was / is] popular in \{year\}.'' \\
8 & ``The election [was / is] contentious in \{year\}.'' \\
9  & ``The streets [were / are] crowded in \{year\}.'' \\
10 & ``The games [were / are] exciting in \{year\}.'' \\
11 & ``The markets [were / are] volatile in \{year\}.'' \\
12 & ``The schools [were / are] busy in \{year\}.'' \\
\bottomrule
\end{tabular}
\caption{\textbf{Sentence templates for the alternative ACY setup.} We score each template in its past and present form (bracketed), sweeping \{year\} over 1900--2250. Because the verb comes early and the year comes last, the model must settle on a tense before it sees the year. Templates 1--8 use singular agreement and templates 9--12 use plural agreement.}
\label{tab:alt_acy_sentences}
\end{table}

%% file: figures/minimal_tense_sentence_panel.tex
\begin{figure}[!htbp]
    \centering
    \includegraphics[width=\linewidth]{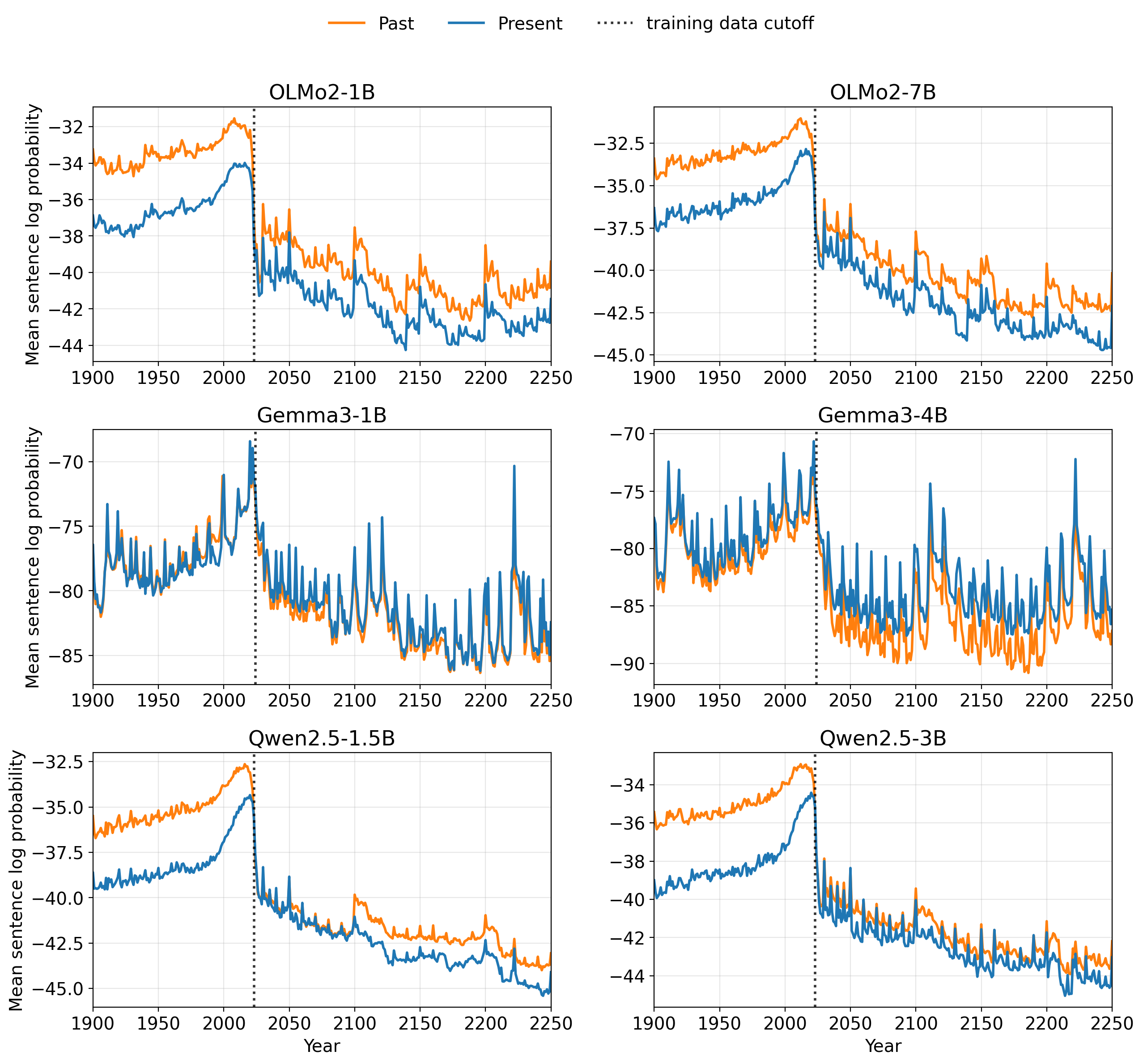}
    \caption{\textbf{The alternative ACY setup reproduces the past-to-present shift when tense precedes the year.} For each of the six LMs, we plot the mean sentence log-probability of the past (orange) and present (blue) tense templates from Table~\ref{tab:alt_acy_sentences} across years, with the training data cutoff marked (dotted line).}
    \label{fig:minimal_tense_sentence_panel}
\end{figure}

%% file: tables/context_association_prompts.tex
\begin{table}[!htbp]
\centering
\small
\begin{tabular}{c c p{0.72\linewidth}}
\toprule
\textbf{\#} & \textbf{Year} & \textbf{Prompt} \\
\midrule
1  & 2010 & ``During the FIFA World Cup in South Africa, there'' \\
2  & 2010 & ``During the earthquake response in Haiti, there'' \\
3  & 2011 & ``During the Arab Spring protests across the Middle East, there'' \\
4  & 2011 & ``At the time of Osama bin Laden's killing, there'' \\
5  & 2012 & ``At the time of the Higgs boson discovery at CERN, there'' \\
6  & 2013 & ``During the Edward Snowden NSA surveillance revelations, there'' \\
7  & 2014 & ``During Russia's annexation of Crimea, there'' \\
8  & 2014 & ``During the Ebola outbreak in West Africa, there'' \\
9  & 2015 & ``During the Paris climate talks, there'' \\
10 & 2016 & ``During the Brexit referendum campaign, there'' \\
11 & 2016 & ``During the US presidential election campaign won by Trump, there'' \\
12 & 2018 & ``During the Cambridge Analytica data scandal, there'' \\
13 & 2019 & ``During the Hong Kong pro-democracy protests, there'' \\
14 & 2019 & ``During Greta Thunberg's school climate strikes, there'' \\
15 & 2020 & ``During the initial COVID-19 lockdowns, there'' \\
16 & 2020 & ``During the Black Lives Matter protests following George Floyd's death, there'' \\
17 & 2021 & ``During the storming of the US Capitol by Trump supporters, there'' \\
18 & 2021 & ``During the Taliban's takeover of Afghanistan, there'' \\
19 & 2022 & ``During Russia's full-scale invasion of Ukraine, there'' \\
20 & 2022 & ``During Elon Musk's acquisition of Twitter, there'' \\
21 & 2023 & ``During the Hollywood writers' and actors' strikes, there'' \\
22 & 2023 & ``During the Israel-Gaza conflict following the October 7th attacks, there'' \\
23 & 2024 & ``During the Paris Olympics opening ceremony on the Seine, there'' \\
24 & 2024 & ``During the US presidential election campaign won by Trump, there'' \\
25 & 2025 & ``During the inauguration of the 47th US president in Washington, there'' \\
26 & 2025 & ``During the DeepSeek AI model release from China, there'' \\
27 & 2026 & ``During the FIFA World Cup hosted jointly by the US, Canada, and Mexico, there'' \\
28 & 2026 & ``During the Winter Olympics in Milan and Cortina, there'' \\
29 & 2028 & ``During the Summer Olympics in Los Angeles, there'' \\
\bottomrule
\end{tabular}
\caption{\textbf{Implicit contextual association prompts.} Each prompt names a real-world event tied to a ground-truth year, without stating the year explicitly. We use 2--3 events per year over 2010--2028, spanning events before, at, and after the training data cutoff.}
\label{tab:context_association_prompts}
\end{table}

%% file: figures/context_acy_no_year.tex
\begin{figure}[!htbp]
    \centering
    \includegraphics[width=\linewidth]{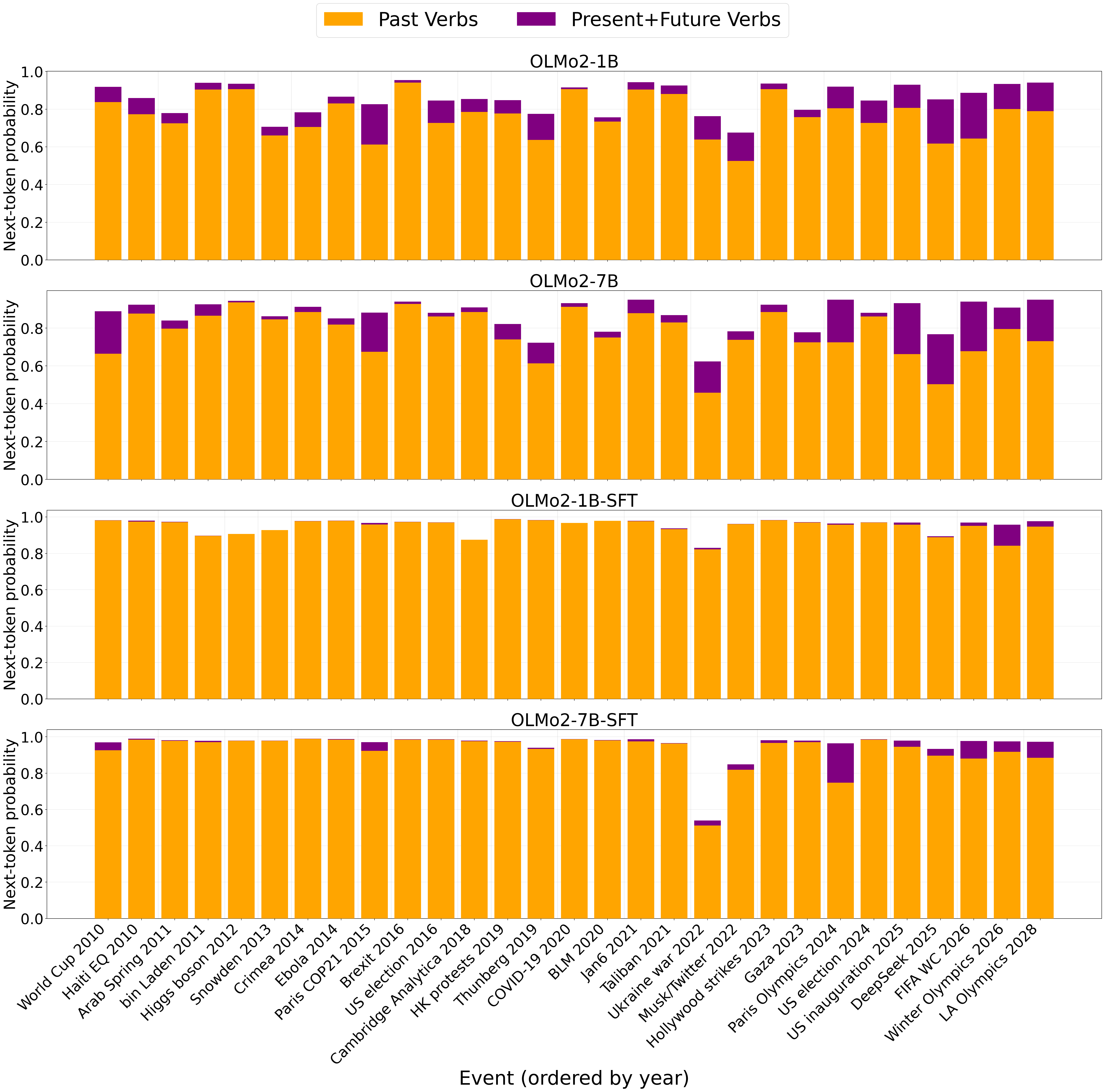}
    \caption{\textbf{The implicit contextual association task does not recover the ACY signal.} For each event prompt (ordered by year on the x-axis), we plot the next-token probability mass on past verbs (orange) versus present+future verbs (purple), for base and SFT versions of \texttt{OLMo2-1B} and \texttt{OLMo2-7B}. Past tense dominates across all events with no clear rise in present+future probability toward the training data cutoff.}
    \label{fig:context_acy_no_year}
\end{figure}

%% file: tables/factual_recency_facts.tex
{\small
\setlength{\tabcolsep}{3pt}
\begin{longtable}{>{\raggedright\arraybackslash}p{0.14\linewidth} >{\raggedright\arraybackslash}p{0.40\linewidth} >{\raggedright\arraybackslash}p{0.40\linewidth}}
\toprule
\textbf{Fact} & \textbf{Continuation form (base LMs)} & \textbf{Question form (SFT LMs)} \\
\midrule
\endfirsthead
\multicolumn{3}{c}{\itshape Table \thetable\ (continued)}\\
\toprule
\textbf{Fact} & \textbf{Continuation form (base LMs)} & \textbf{Question form (SFT LMs)} \\
\midrule
\endhead
\midrule \multicolumn{3}{r}{\itshape continued on next page}\\
\endfoot
\bottomrule
\caption{\textbf{Factual recency probe: the 18 recurring facts and all prompts.} Each fact (with its answer as of the 2023 cutoff in parentheses) has a well-defined ``most recent'' answer that changes over time. We write three phrasings per fact, each realized as a sentence-completion (continuation) prompt for base LMs and a matching question prompt for SFT LMs.}\label{tab:factual_recency_facts}\\
\endlastfoot
Summer Olympics \newline \textit{(Tokyo)} & (1) ``The most recent Summer Olympics was held in the city of'' \newline (2) ``The host city of the most recent Summer Olympics was'' \newline (3) ``The most recent Summer Olympic Games took place in'' & (1) ``What city hosted the most recent Summer Olympics?'' \newline (2) ``Which city most recently hosted the Summer Olympics?'' \newline (3) ``In what city did the most recent Summer Olympic Games take place?'' \\
\addlinespace
Winter Olympics \newline \textit{(Beijing)} & (1) ``The most recent Winter Olympics was held in the city of'' \newline (2) ``The host city of the most recent Winter Olympics was'' \newline (3) ``The most recent Winter Olympic Games took place in'' & (1) ``What city hosted the most recent Winter Olympics?'' \newline (2) ``Which city most recently hosted the Winter Olympics?'' \newline (3) ``In what city did the most recent Winter Olympic Games take place?'' \\
\addlinespace
Nobel Peace Prize \newline \textit{(Mohammadi)} & (1) ``The most recent Nobel Peace Prize was awarded to'' \newline (2) ``The most recent winner of the Nobel Peace Prize is'' \newline (3) ``The person who most recently received the Nobel Peace Prize is'' & (1) ``Who most recently won the Nobel Peace Prize?'' \newline (2) ``Who is the most recent Nobel Peace Prize laureate?'' \newline (3) ``Who was the most recent recipient of the Nobel Peace Prize?'' \\
\addlinespace
Nobel Physics \newline \textit{(Krausz)} & (1) ``The most recent Nobel Prize in Physics was awarded to'' \newline (2) ``The most recent winner of the Nobel Prize in Physics is'' \newline (3) ``The most recent Nobel Prize in Physics laureate is'' & (1) ``Who most recently won the Nobel Prize in Physics?'' \newline (2) ``Who is the most recent Nobel Prize in Physics winner?'' \newline (3) ``Who is the most recent Nobel Physics Prize laureate?'' \\
\addlinespace
Nobel Chemistry \newline \textit{(Bawendi)} & (1) ``The most recent Nobel Prize in Chemistry was awarded to'' \newline (2) ``The most recent winner of the Nobel Prize in Chemistry is'' \newline (3) ``The most recent Nobel Prize in Chemistry laureate is'' & (1) ``Who most recently won the Nobel Prize in Chemistry?'' \newline (2) ``Who is the most recent Nobel Prize in Chemistry winner?'' \newline (3) ``Who is the most recent Nobel Chemistry Prize laureate?'' \\
\addlinespace
Nobel Literature \newline \textit{(Fosse)} & (1) ``The most recent Nobel Prize in Literature was awarded to'' \newline (2) ``The most recent winner of the Nobel Prize in Literature is'' \newline (3) ``The most recent Nobel Prize in Literature laureate is'' & (1) ``Who most recently won the Nobel Prize in Literature?'' \newline (2) ``Who is the most recent Nobel Prize in Literature winner?'' \newline (3) ``Who is the most recent Nobel Literature Prize laureate?'' \\
\addlinespace
Nobel Medicine \newline \textit{(Kariko)} & (1) ``The most recent Nobel Prize in Physiology or Medicine was awarded to'' \newline (2) ``The most recent winner of the Nobel Prize in Physiology or Medicine is'' \newline (3) ``The most recent Nobel Prize in Physiology or Medicine laureate is'' & (1) ``Who most recently won the Nobel Prize in Physiology or Medicine?'' \newline (2) ``Who is the most recent Nobel Prize in Medicine winner?'' \newline (3) ``Who is the most recent Nobel Medicine Prize laureate?'' \\
\addlinespace
US President \newline \textit{(Biden)} & (1) ``The most recent US presidential election was won by'' \newline (2) ``The most recently inaugurated President of the United States is'' \newline (3) ``The most recent winner of the US presidential election is'' & (1) ``Who most recently won the US presidential election?'' \newline (2) ``Who is the most recently inaugurated President of the United States?'' \newline (3) ``What is the name of the most recent US presidential election winner?'' \\
\addlinespace
UK Prime Minister \newline \textit{(Sunak)} & (1) ``The most recently appointed Prime Minister of the United Kingdom is'' \newline (2) ``The most recent person to become Prime Minister of the United Kingdom is'' \newline (3) ``The most recent Prime Minister of the United Kingdom is'' & (1) ``Who is the most recently appointed UK Prime Minister?'' \newline (2) ``Who most recently became UK Prime Minister?'' \newline (3) ``Who is the most recent Prime Minister of the United Kingdom?'' \\
\addlinespace
Academy Award \newline \textit{(Everything)} & (1) ``The most recent Academy Award for Best Picture was won by the film'' \newline (2) ``The most recent winner of the Oscar for Best Picture is the film'' \newline (3) ``The film that most recently won the Academy Award for Best Picture is'' & (1) ``What film most recently won the Academy Award for Best Picture?'' \newline (2) ``What is the most recent Oscar Best Picture winner?'' \newline (3) ``Which film most recently won the Oscar for Best Picture?'' \\
\addlinespace
Time Person of the Year \newline \textit{(Swift)} & (1) ``The most recent Time Person of the Year is'' \newline (2) ``The most recent winner of Time magazine's Person of the Year award is'' \newline (3) ``Time magazine's most recent Person of the Year is'' & (1) ``Who is the most recent Time Person of the Year?'' \newline (2) ``Who most recently won Time's Person of the Year?'' \newline (3) ``Who is Time magazine's most recent Person of the Year?'' \\
\addlinespace
M-W Word of the Year \newline \textit{(authentic)} & (1) ``The most recent Merriam-Webster word of the year is'' \newline (2) ``The word most recently named Merriam-Webster's word of the year is'' \newline (3) ``Merriam-Webster's most recent word of the year is'' & (1) ``What is the most recent Merriam-Webster word of the year?'' \newline (2) ``What word was most recently named Merriam-Webster's word of the year?'' \newline (3) ``What is Merriam-Webster's most recent word of the year?'' \\
\addlinespace
Chinese Zodiac \newline \textit{(Rabbit)} & (1) ``The most recent Chinese Zodiac year is the Year of the'' \newline (2) ``The Chinese Zodiac animal for this year is the'' \newline (3) ``In the Chinese Zodiac, this year is the Year of the'' & (1) ``What is the most recent Chinese Zodiac animal?'' \newline (2) ``What is the Chinese Zodiac animal for this year?'' \newline (3) ``According to the Chinese Zodiac, what animal represents this year?'' \\
\addlinespace
FIFA World Cup \newline \textit{(Qatar)} & (1) ``The most recent FIFA World Cup was hosted by the country of'' \newline (2) ``The host country of the most recent FIFA World Cup was'' \newline (3) ``The most recent FIFA World Cup took place in the country of'' & (1) ``Which country most recently hosted the FIFA World Cup?'' \newline (2) ``What country most recently hosted the FIFA World Cup?'' \newline (3) ``In what country did the most recent FIFA World Cup take place?'' \\
\addlinespace
US Vice President \newline \textit{(Harris)} & (1) ``The most recently inaugurated Vice President of the United States is'' \newline (2) ``The most recent winner of the US vice presidential election is'' \newline (3) ``The most recent Vice President of the United States is'' & (1) ``Who is the most recently inaugurated Vice President of the United States?'' \newline (2) ``Who most recently became Vice President of the United States?'' \newline (3) ``Who is the most recent Vice President of the United States?'' \\
\addlinespace
German Chancellor \newline \textit{(Scholz)} & (1) ``The most recently elected Chancellor of Germany is'' \newline (2) ``The most recent person to become Chancellor of Germany is'' \newline (3) ``The most recent Chancellor of Germany is'' & (1) ``Who is the most recently elected Chancellor of Germany?'' \newline (2) ``Who most recently became Chancellor of Germany?'' \newline (3) ``Who is the most recent Chancellor of Germany?'' \\
\addlinespace
Italian PM \newline \textit{(Meloni)} & (1) ``The most recently appointed Prime Minister of Italy is'' \newline (2) ``The most recent person to become Prime Minister of Italy is'' \newline (3) ``The most recent Prime Minister of Italy is'' & (1) ``Who is the most recently appointed Prime Minister of Italy?'' \newline (2) ``Who most recently became Prime Minister of Italy?'' \newline (3) ``Who is the most recent Prime Minister of Italy?'' \\
\addlinespace
Brazilian President \newline \textit{(Lula)} & (1) ``The most recently elected President of Brazil is'' \newline (2) ``The most recent winner of the Brazilian presidential election is'' \newline (3) ``The most recent President of Brazil is'' & (1) ``Who is the most recently elected President of Brazil?'' \newline (2) ``Who most recently won the Brazilian presidential election?'' \newline (3) ``Who is the most recent President of Brazil?'' \\
\addlinespace
\end{longtable}
}

%% file: figures/factual_recency_ranges.tex
\begin{figure}[!htbp]
    \centering
    \includegraphics[width=\linewidth]{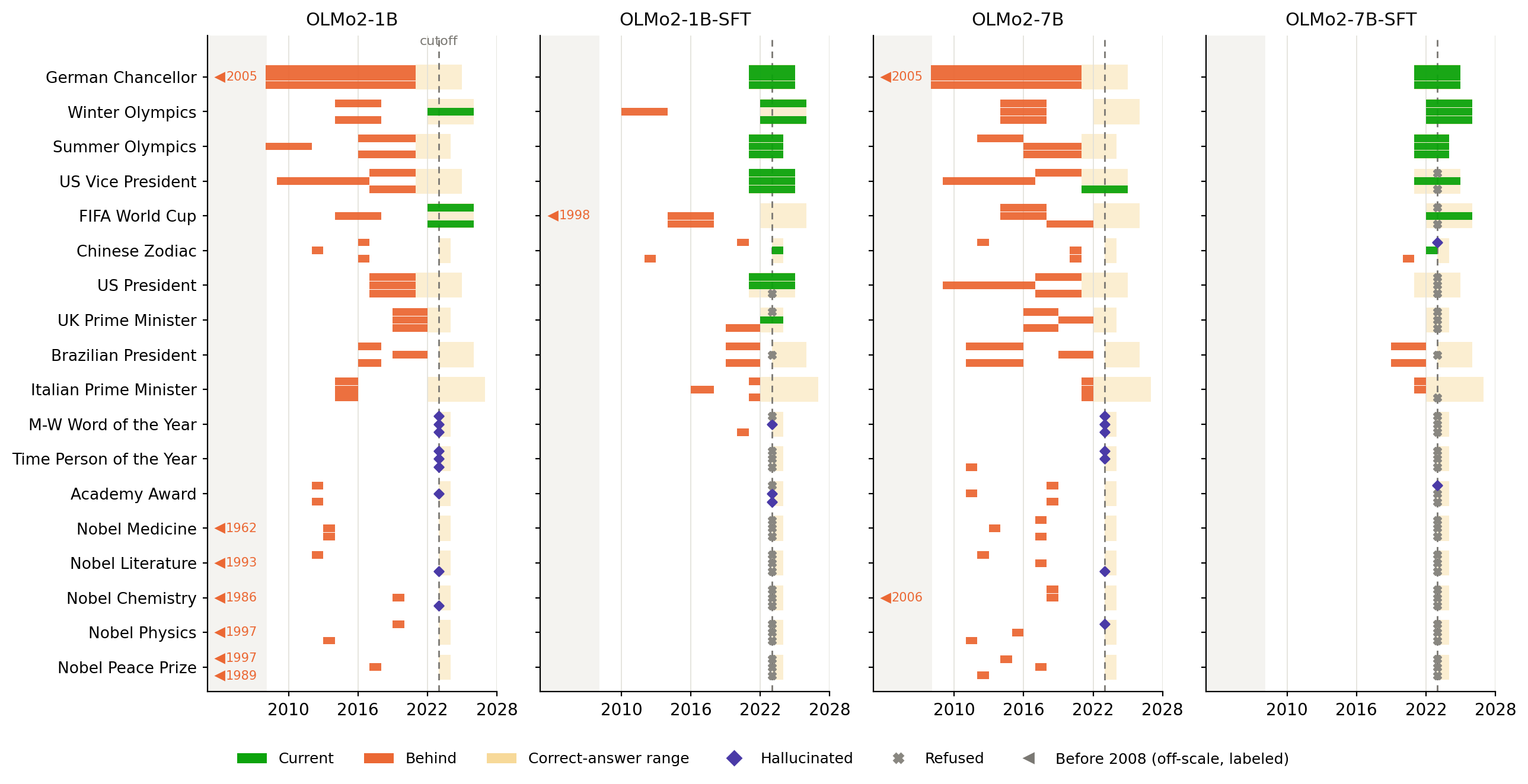}
    \caption{\textbf{Factual recency does not cleanly encode the current year.} Each row is one of 18 recurring facts; for its three phrasings we draw the year range over which the LM's named answer is correct; \emph{current} (green) if the range covers the $\sim$2023 cutoff (dashed line), \emph{behind} (orange) if the answer is real but older. Alternatively, we draw a marker for a \emph{hallucinated} ($\diamond$) or \emph{refused} ($\times$) response. The gold band is the correct-answer range; a left-pointing mark in the margin flags an answer from before 2008 (labeled with its year). Columns are the base and SFT versions of \texttt{OLMo2-1B} and \texttt{OLMo2-7B}.}
    \label{fig:factual_recency_ranges}
\end{figure}

%% file: figures/perplexity_predictions.tex
\begin{figure}[!htbp]
    \centering
    \includegraphics[width=0.9\linewidth]{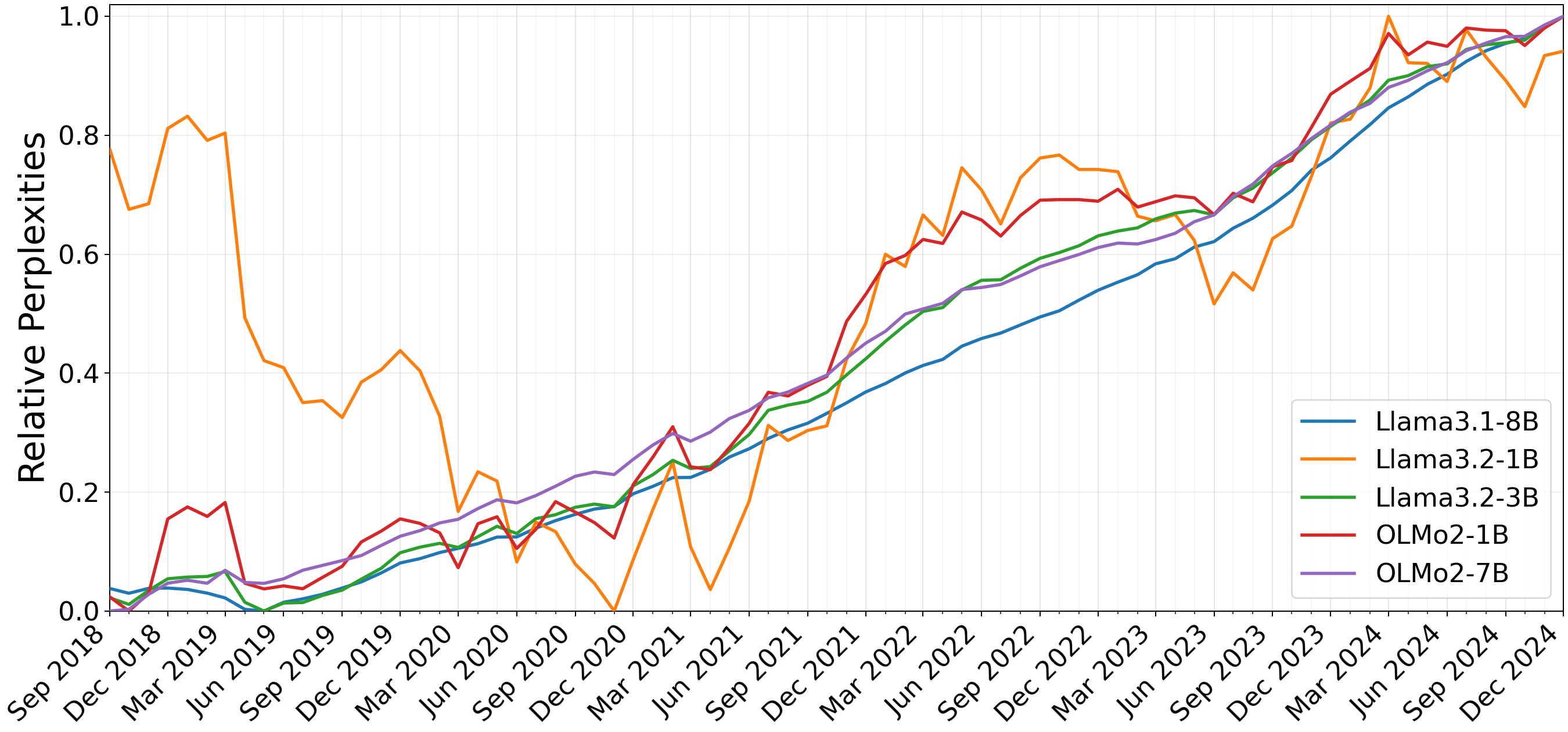}
    \caption{\textbf{Relative perplexities of LMs per month using the WIKISPAN dataset.}}
    \label{fig:perplexity_predictions}
\end{figure}

%% file: figures/associative_data_count_stage1.tex
\begin{figure}[!htbp]
    \centering
    \begin{subfigure}[b]{0.45\linewidth}
        \centering
        \includegraphics[width=\linewidth]{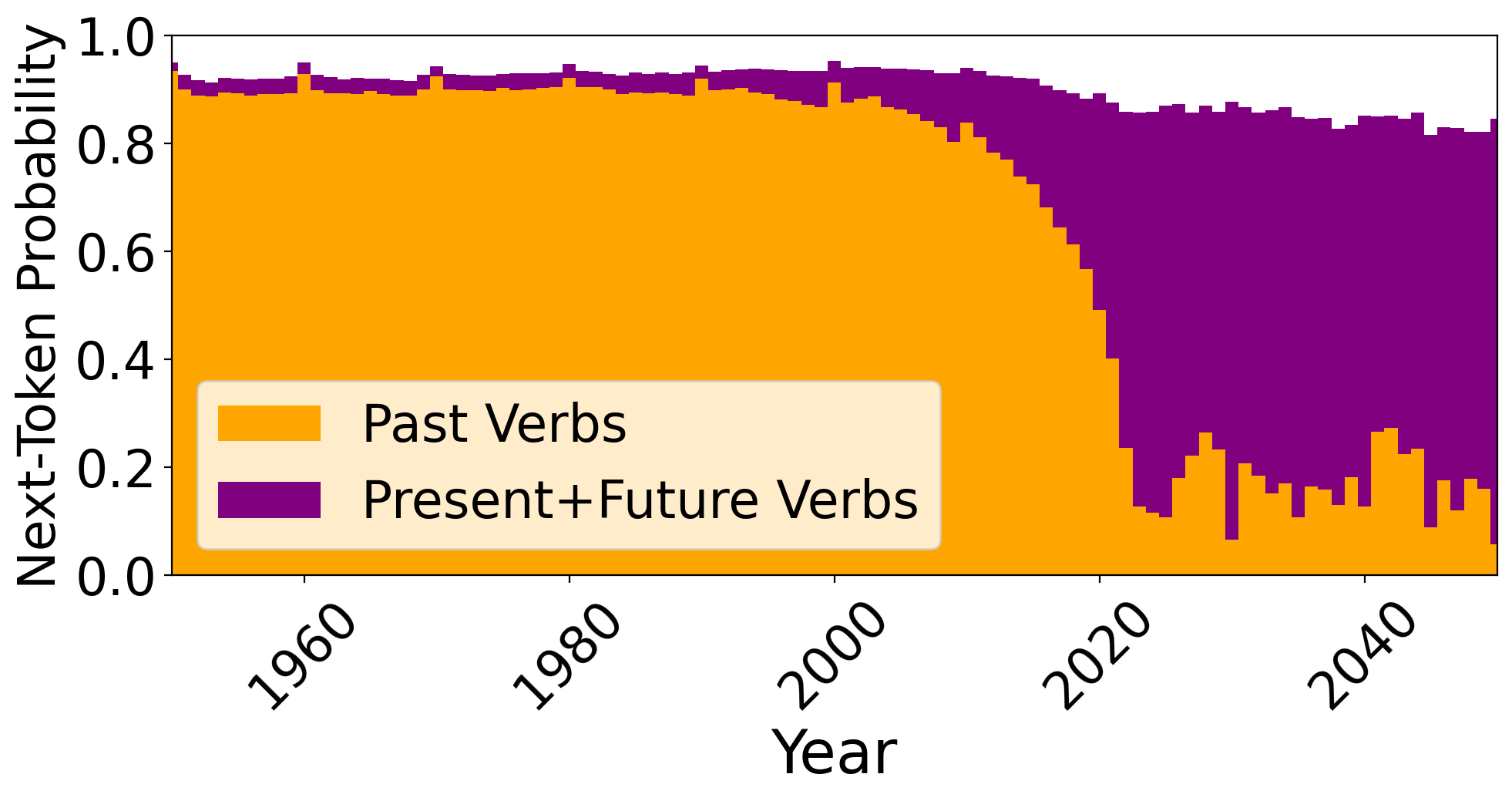}
        \caption{\texttt{OLMo2-1B} predictions}
        \label{fig:stage1_olmo1b}
    \end{subfigure}\hfill%
    \begin{subfigure}[b]{0.45\linewidth}
        \centering
        \includegraphics[width=\linewidth]{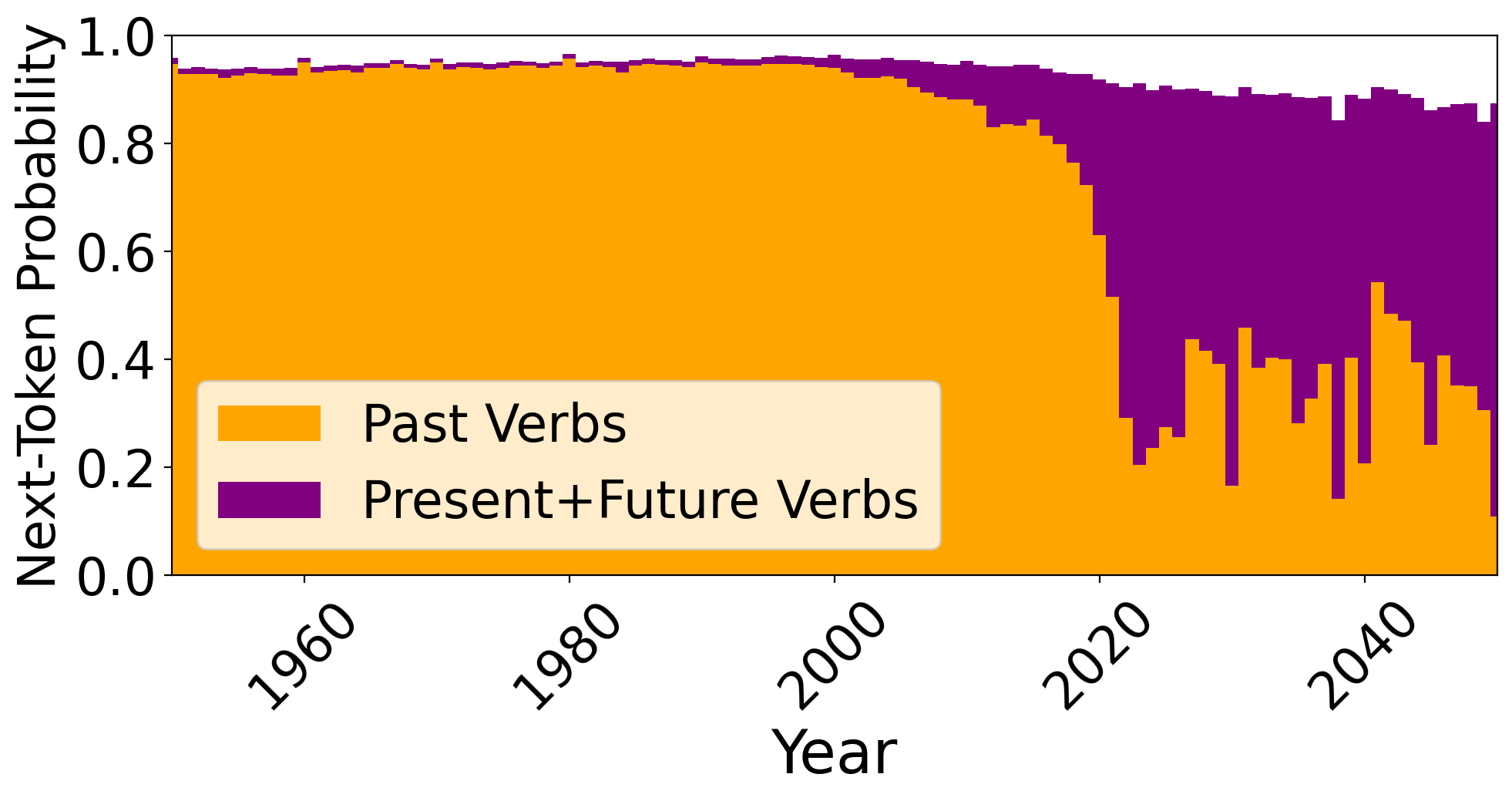}
        \caption{\texttt{OLMo2-7B} predictions}
        \label{fig:stage1_olmo7b}
    \end{subfigure}

    \vspace{1ex}

    \begin{subfigure}[b]{0.45\linewidth}
        \centering
        \includegraphics[width=\linewidth]{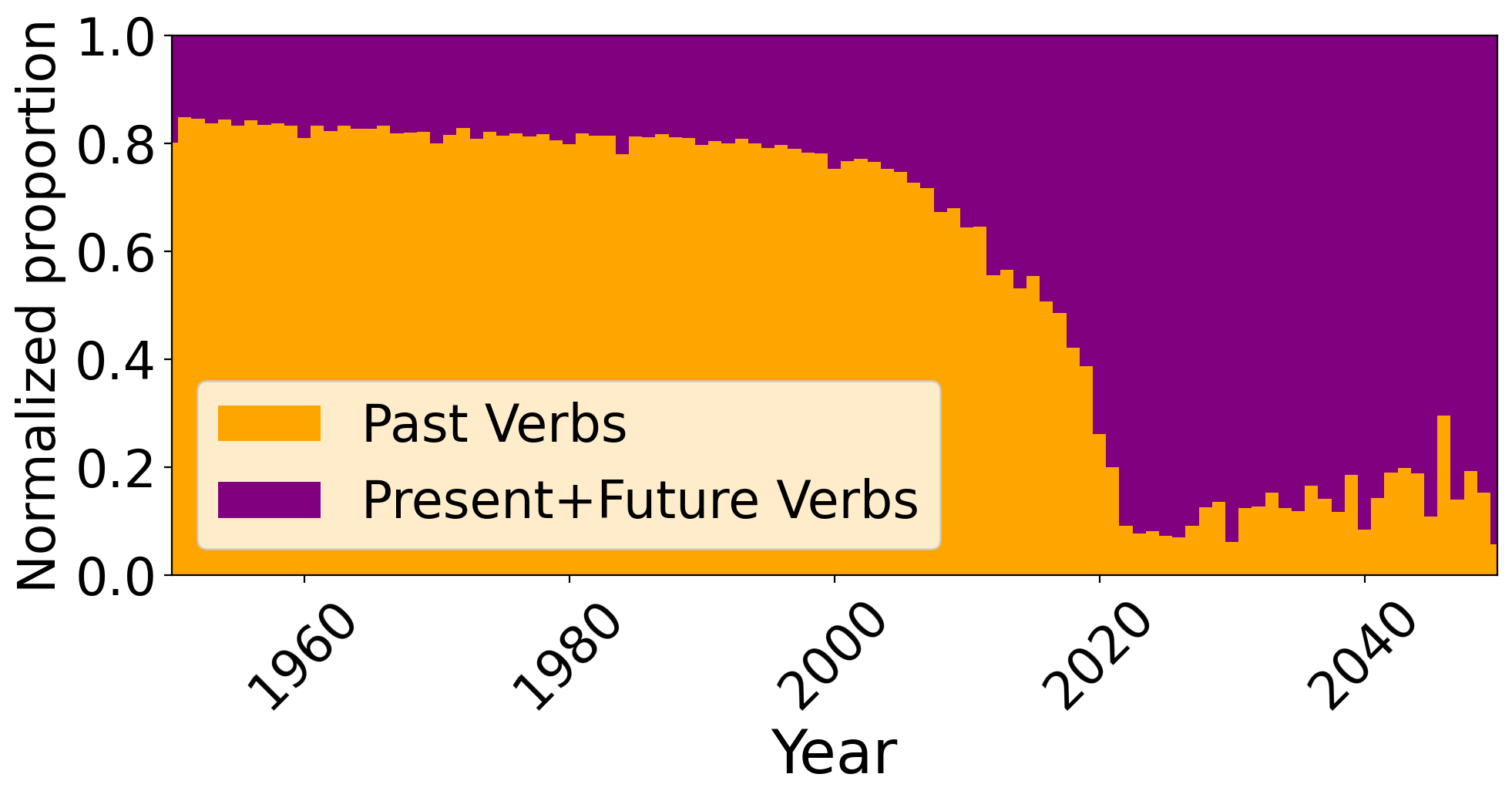}
        \caption{Co-occurrence predictions}
        \label{fig:stage1_cooccur}
    \end{subfigure}\hfill%
    \begin{subfigure}[b]{0.45\linewidth}
        \centering
        \includegraphics[width=\linewidth]{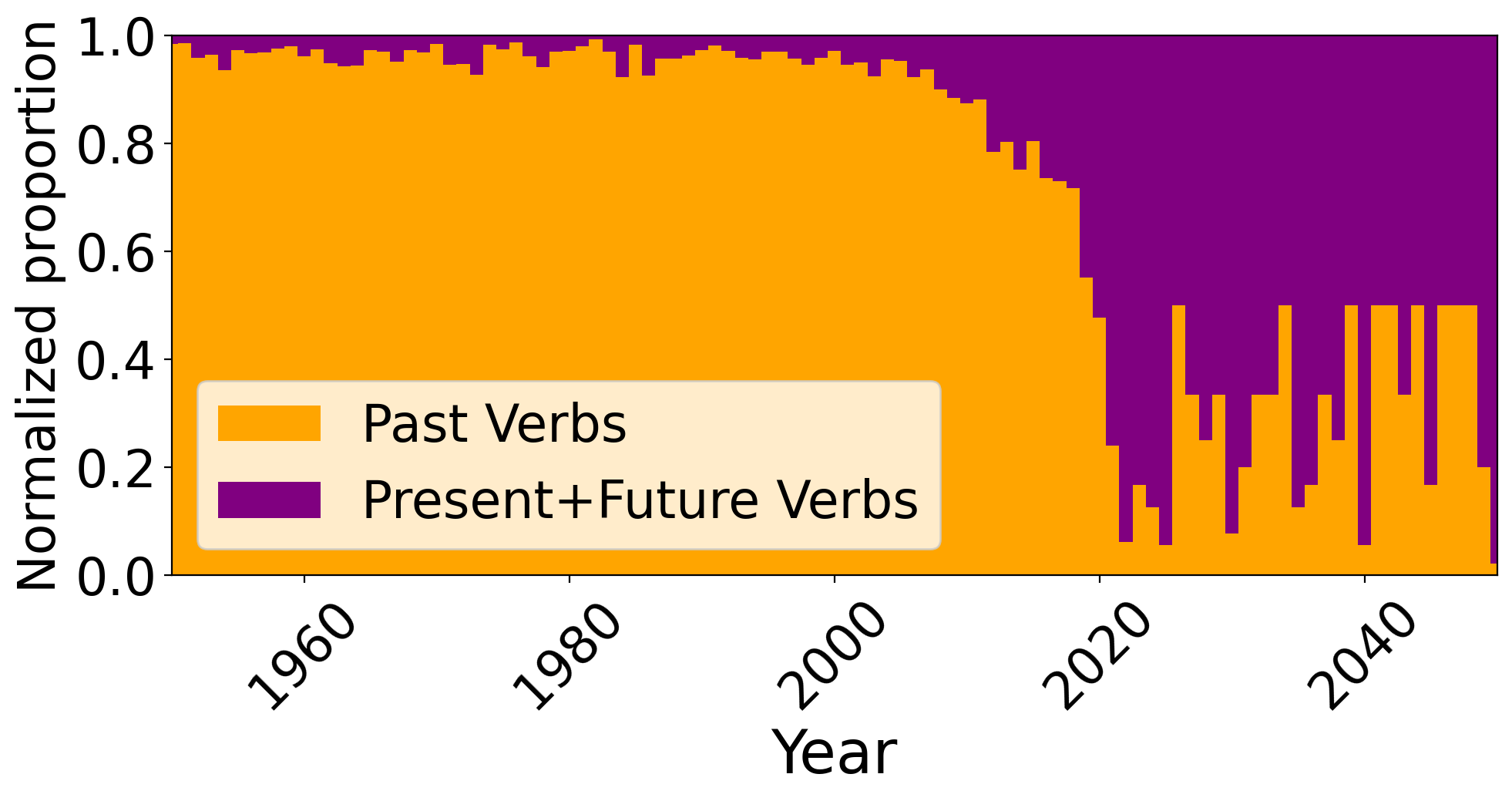}
        \caption{N-gram model predictions}
        \label{fig:stage1_ngram}
    \end{subfigure}
    \caption{\textbf{Next token predictions on \texttt{OLMo2} base stage1 LM.} \ref{fig:stage1_olmo1b} and~\ref{fig:stage1_olmo7b} show next-token predictions of the LM after stage2, \ref{fig:stage1_cooccur} shows the normalized co-occurrence model created from 10k steps of olmo-mix-1124, \ref{fig:stage1_ngram} shows the normalized n-gram model created from 10k steps of olmo-mix-1124. The associative task follows the training data distribution.}
    \label{fig:relative_comparison_stage1}
\end{figure}

%% file: figures/associative_data_count_stage2.tex
\begin{figure}[!htbp]
    \centering
    \begin{subfigure}[b]{0.45\linewidth}
        \centering
        \includegraphics[width=\linewidth]{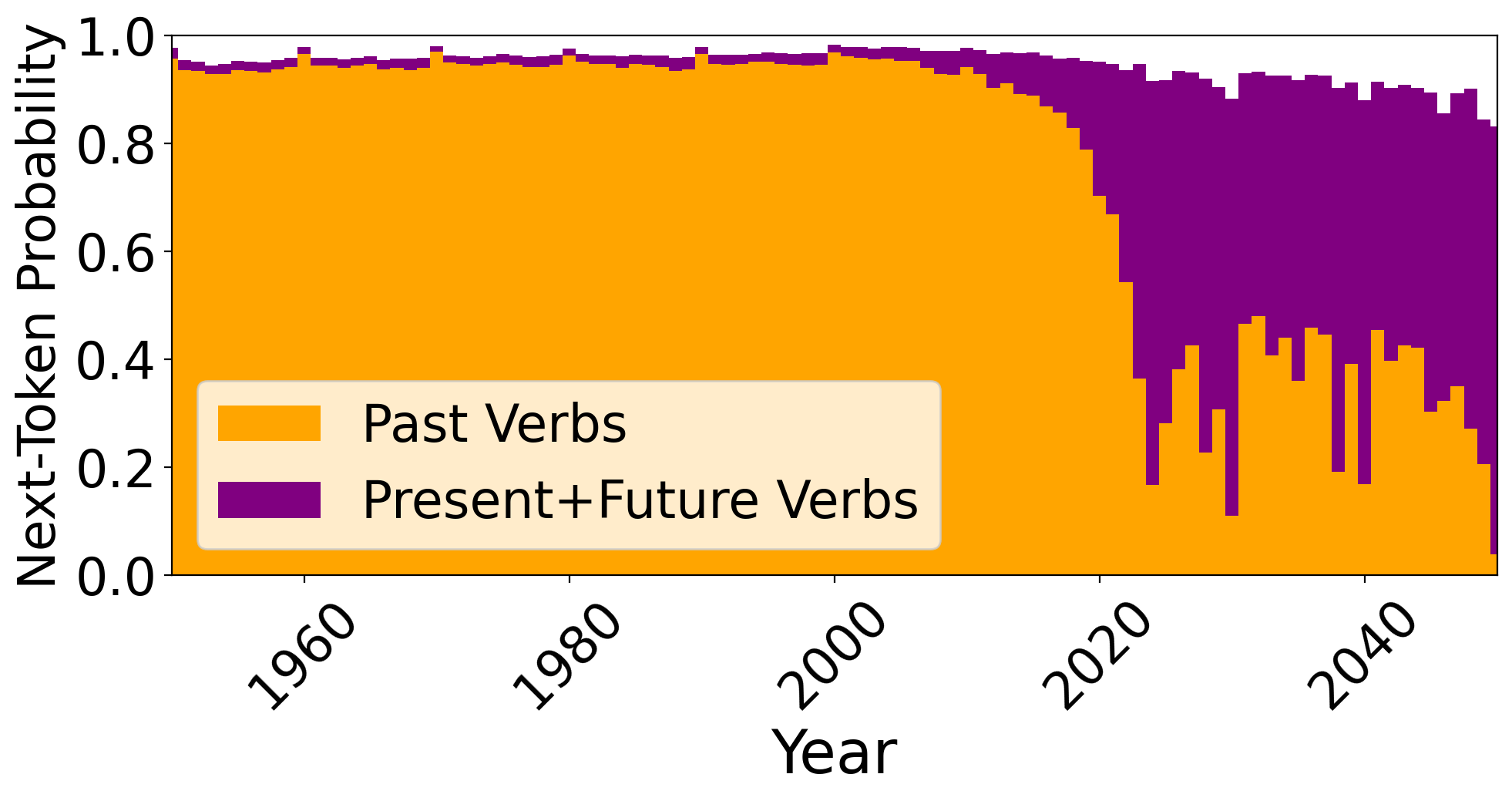}
        \caption{\texttt{OLMo2-1B} predictions}
        \label{fig:stage2_olmo1b}
    \end{subfigure}
    \begin{subfigure}[b]{0.45\linewidth}
        \centering
        \includegraphics[width=\linewidth]{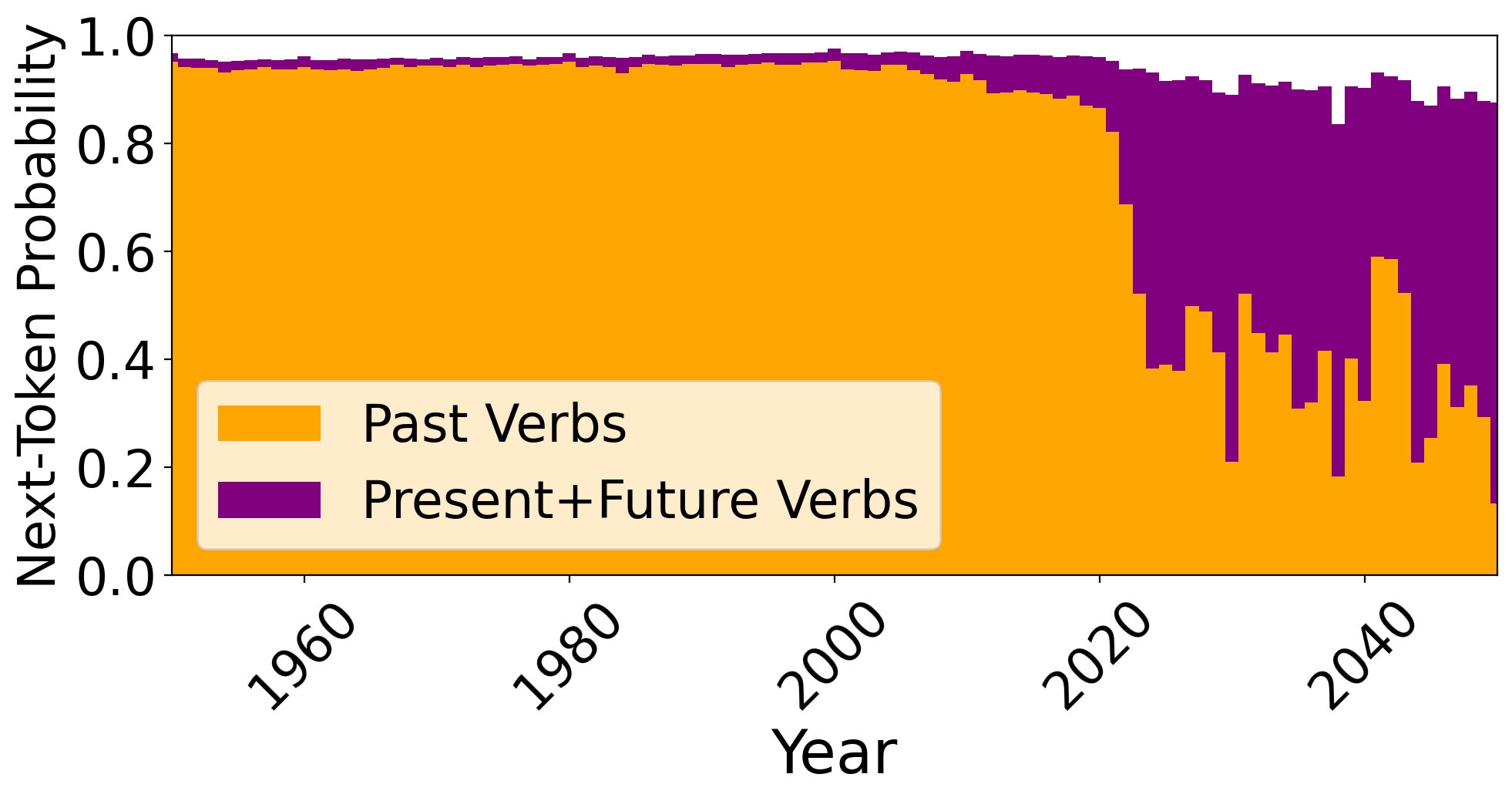}
        \caption{\texttt{OLMo2-7B} predictions}
        \label{fig:stage2_olmo7b}
    \end{subfigure}
    \begin{subfigure}[b]{0.45\linewidth}
        \centering
        \includegraphics[width=\linewidth]{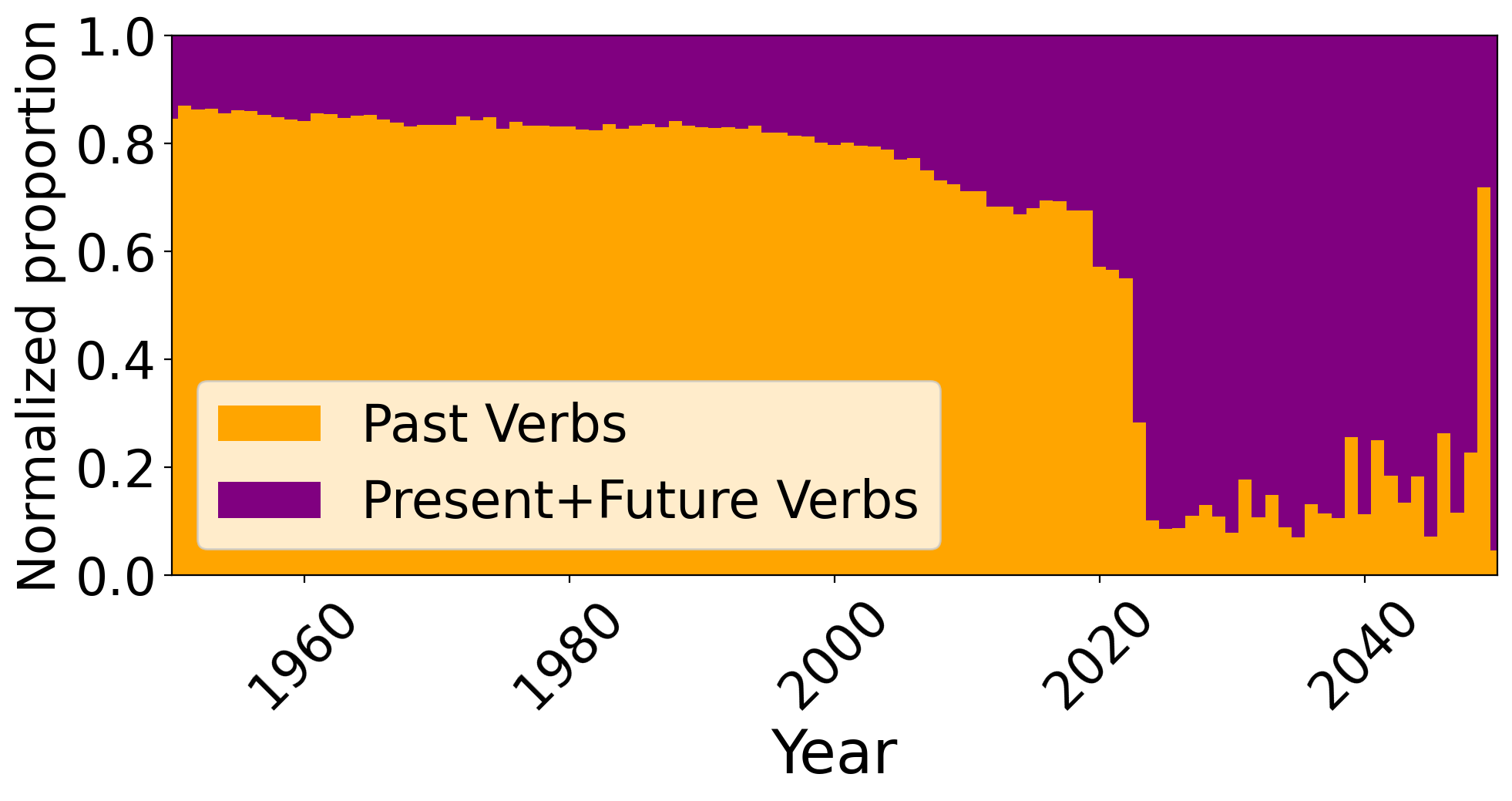}
        \caption{Co-occurrence predictions}
        \label{fig:stage2_cooccur}
    \end{subfigure}
    \begin{subfigure}[b]{0.45\linewidth}
        \centering
        \includegraphics[width=\linewidth]{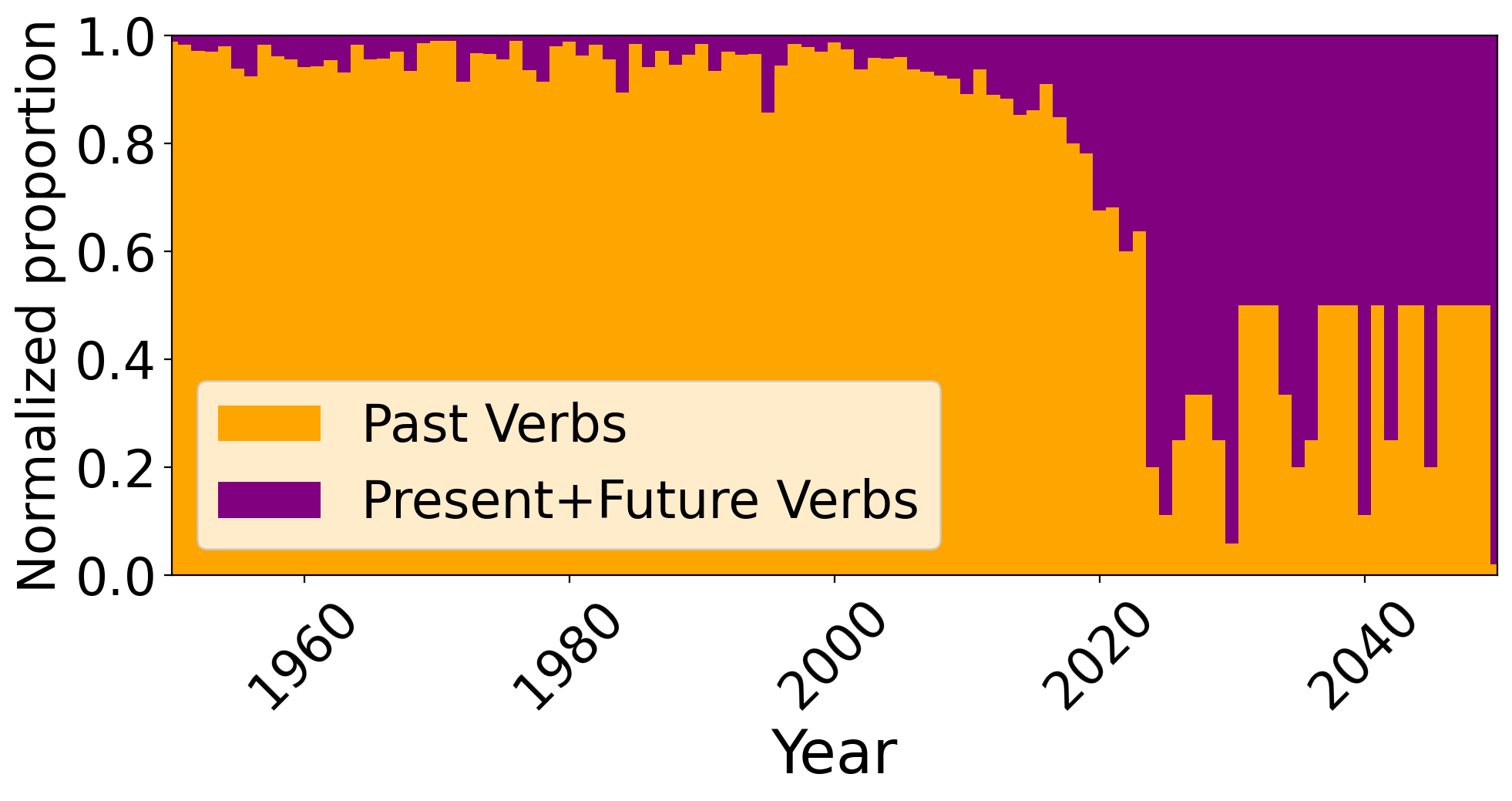}
        \caption{N-gram model predictions}
        \label{fig:stage2_ngram}
    \end{subfigure}
    \caption{\textbf{Next token predictions on \texttt{OLMo2} base stage2 LM.} \ref{fig:stage2_olmo1b} and~\ref{fig:stage2_olmo7b} show next-token predictions of the LM after stage2, \ref{fig:stage2_cooccur} shows the normalized co-occurrence model created from 10k steps of dolmino-mix-1124, \ref{fig:stage2_ngram} shows the normalized n-gram model created from 10k steps of dolmino-mix-1124. The associative task follows the training data distribution.}
    \label{fig:relative_comparison_stage2}
\end{figure}

%% file: tables/ngram_cooccur_CE.tex
\begin{table}[!htbp]
\centering
\small
\begin{tabular}{llcc}
\toprule
\textbf{Count model} & \textbf{LM} & \textbf{Stage 1} & \textbf{Stage 2} \\
\toprule
\multirow{2}{*}{N-gram}
  & OLMo2-1B & 0.34 \,($-$0.40) & 0.34 \,($-$0.19) \\
  & OLMo2-7B & 0.35 \,($-$0.28) & 0.36 \,($-$0.12) \\
\midrule
\multirow{2}{*}{Co-occurrence}
  & OLMo2-1B & 0.61 \,($-$0.42) & 0.68 \,($-$0.21) \\
  & OLMo2-7B & 0.78 \,($-$0.33) & 0.70 \,($-$0.16) \\
\bottomrule
\end{tabular}
\caption{\textbf{\texttt{OLMo2} LMs have lower cross-entropy with corpus-count models than baselines.} The first two sections show the CE difference against the N-gram and co-occurrence baselines, with the delta versus the corresponding random baseline in parentheses. The last two sections show the CE difference against random token selections drawn from N-gram and co-occurrence count distributions, respectively.}
\label{tab:cross_entropy_comparisons}
\end{table}

%% file: tables/detailed_subspace_analysis.tex
\begin{table}[!htbp]
\centering
\small
{\setlength{\tabcolsep}{6pt} 
\begin{tabular}{cccccc}
\toprule
\textbf{Layer} &
\shortstack{\textbf{Random}\\\textbf{sub-}\\\textbf{space}} &
\shortstack{\textbf{Randomly}\\\textbf{initialized}\\\textbf{model}} &
\shortstack{\textbf{Non-year}\\\textbf{non-number}\\\textbf{entities}} &
\shortstack{\textbf{Non-year}\\\textbf{num-}\\\textbf{bers}} &
\shortstack{\textbf{Different}\\\textbf{template}\\\textbf{structure}} \\
\midrule
31 & 0.14 & 0.00 & 0.00 & 0.00 & 0.24 \\
30 & 0.14 & 0.00 & 0.00 & 0.00 & 0.24 \\
28 & 0.14 & 0.00 & 0.00 & 0.08 & 0.28 \\
26 & 0.14 & 0.00 & 0.00 & 0.10 & 0.26 \\
24 & 0.14 & 0.00 & 0.00 & 0.15 & 0.28 \\
22 & 0.14 & 0.00 & 0.39 & 0.35 & 0.39 \\
20 & 0.14 & 0.00 & 0.14 & 0.35 & 0.39 \\
18 & 0.14 & 0.00 & 0.45 & 0.62 & 0.41 \\
16 & 0.14 & 0.00 & 0.46 & 0.72 & 0.51 \\
14 & 0.14 & 0.00 & 0.09 & 0.86 & 0.62 \\
12 & 0.14 & 0.00 & 0.02 & 0.87 & 0.61 \\
10 & 0.14 & 0.00 & 0.00 & 0.88 & 0.62 \\
8  & 0.14 & 0.00 & 0.01 & 0.91 & 0.67 \\
6  & 0.14 & 0.00 & 0.00 & 0.97 & 0.71 \\
4  & 0.14 & 0.00 & 0.00 & 1.00 & 0.74 \\
2  & 0.14 & 0.00 & 0.00 & 1.00 & 0.71 \\
1  & 0.14 & 0.00 & 0.00 & 1.00 & 0.67 \\
0  & 0.14 & 0.00 & 0.00 & 0.67 & 0.29 \\
\bottomrule
\end{tabular}
}

\caption{\textbf{The year-type subspace is non-trivial and specific to year--like representations.} IIA across layers on the second token of the year for various setups.}
\label{tab:detailed_subspace_analysis}
\end{table}

%% file: tables/associative_interchange_prompts.tex
\begin{table}[!htbp]
\centering
\small
\setlength{\tabcolsep}{3pt} 

\begin{subtable}[t]{\linewidth}
\centering
\begin{tabular}{c p{0.7\linewidth}}
\toprule
\textbf{\#} & \textbf{Prompt} \\
\midrule
\multicolumn{2}{l}{\textbf{Non-year non-number prompts}} \\
\midrule
1  & ``In summary there'' \\
2  & ``In Amsterdam there'' \\
3  & ``In water there'' \\
4  & ``In math there'' \\
5  & ``In theory there'' \\
6  & ``In general there'' \\
7  & ``In conclusion there'' \\
8  & ``In practice there'' \\
9  & ``In contrast there'' \\
10 & ``In reality there'' \\
11 & ``In essence there'' \\
\bottomrule
\end{tabular}
\caption{\textbf{Prompts used to interchange with non-numeric words.} Non-year, non-numeric words are substituted into the \textit{In year there} template.}
\end{subtable}

\vspace{1em}

\begin{subtable}[t]{\linewidth}
\centering
\begin{tabular}{c p{0.7\linewidth}}
\toprule
\textbf{\#} & \textbf{Prompt} \\
\midrule
\multicolumn{2}{l}{\textbf{Non-year number prompts}} \\
\midrule
1 & ``16 * 1850 + 1953 is'' \\
2 & ``4 - 3 + 1741 is'' \\
3 & ``100 + 200 + 1876 is'' \\
4 & ``2000 - 100 + 1492 is'' \\
5 & ``50 * 2 + 1066 is'' \\
6 & ``999 - 1 + 1789 is'' \\
7 & ``7 * 8 + 1945 is'' \\
8 & ``86828 + 4 - 2001 is'' \\
9 & ``3 * 63 + 1000 + 1998 is'' \\
\bottomrule
\end{tabular}
\caption{\textbf{Prompts used to interchange numbers that look like years but are not.} Arithmetic expressions containing year-like numbers in non-temporal contexts}
\end{subtable}
\caption{\textbf{All prompts used for the associative interchange robustness experiments.}}
\label{tab:associative_interchange_prompts}

\end{table}

%% file: figures/causal_tracing_associative.tex
\begin{figure}[!htbp]
    \centering
    \begin{subfigure}[b]{0.7\linewidth}
        \centering
        \includegraphics[width=\linewidth]{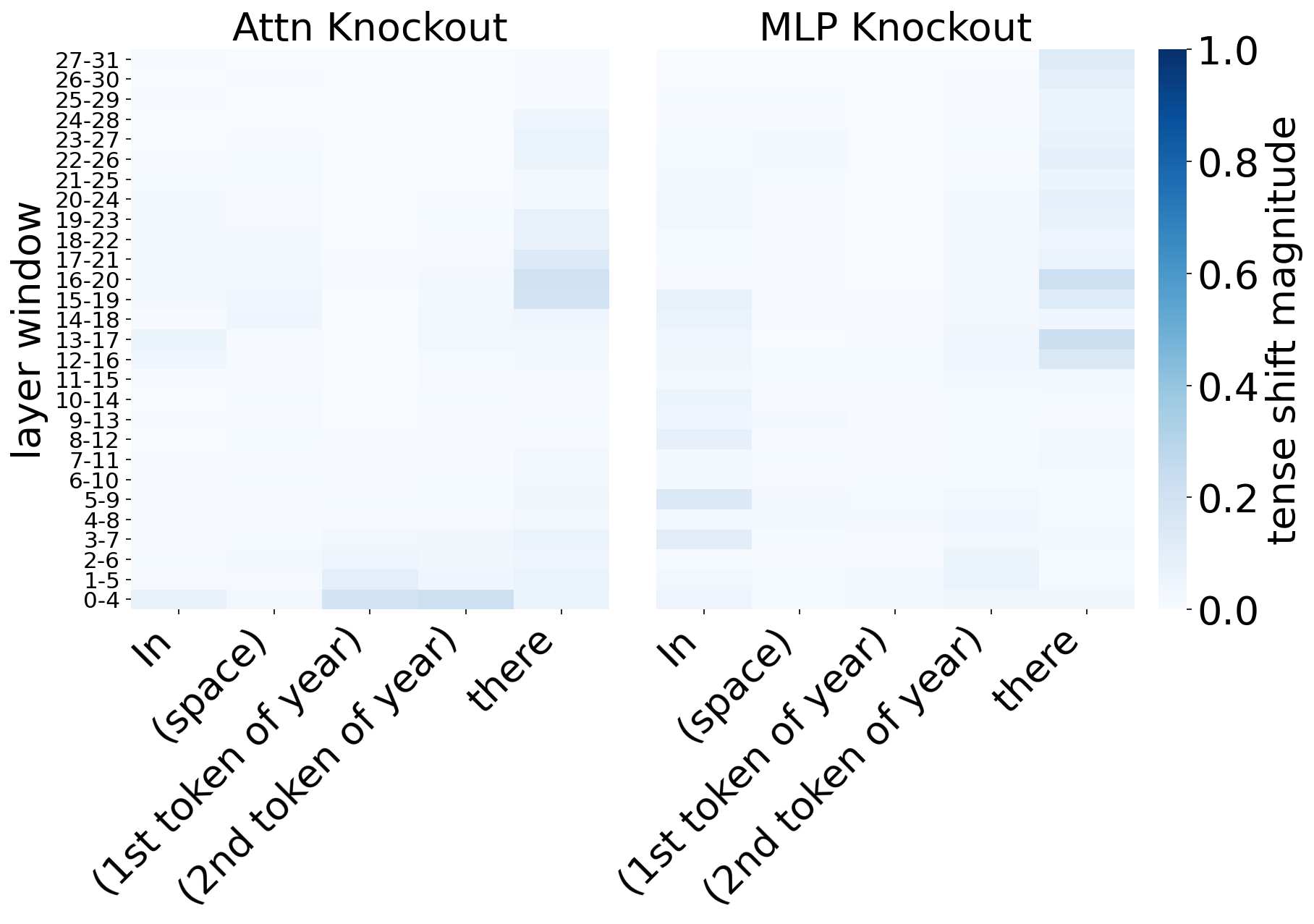}
        \caption{Past years}
        \label{fig:past-assoc}
    \end{subfigure}
    \begin{subfigure}[b]{0.7\linewidth}
        \centering
        \includegraphics[width=\linewidth]{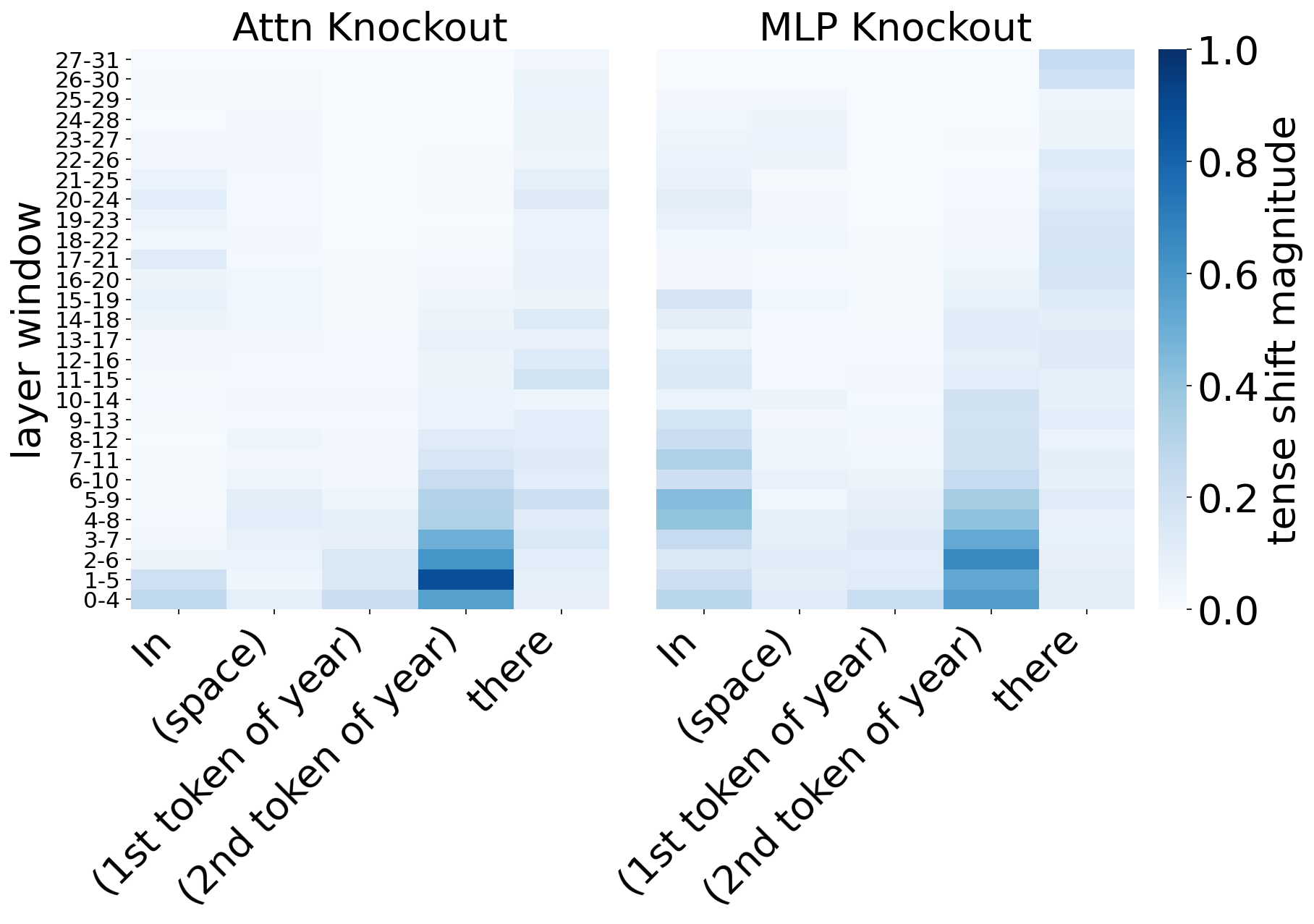}
        \caption{Future years}
        \label{fig:pres-fut-assoc}
    \end{subfigure}

    \caption{\textbf{Attention and MLP knockout corroborate associative task's causal abstraction results.} We show results for the associative task, measured by tense shift magnitude (TSM), averaged across 35 past-year (top) and 35 future-year (bottom) prompts. We see large TSM in early layers on year tokens and late layers on the final token, the same as in causal abstraction. Future-year prompts show stronger dependence on the second year token (last digit) than past-year prompts.}
    
    \label{fig:mech_analysis_associative}
    \vspace{-3ex}
\end{figure}

%% file: tables/declr_prompts.tex
\begin{table}[!htbp]
\centering
\small
\begin{tabular}{c p{0.7\linewidth}}
\toprule
\textbf{\#} & \textbf{Prompt} \\
\midrule
\multicolumn{2}{c}{\textbf{Continuation}} \\
\midrule
1 & ``Finish the sentence: The current year (in YYYY format) is'' \\
2 & ``What day of the week is Valentines day this year? For the year'' \\
3 & ``Today, the year is'' \\
4 & ``Write a story about the current year. Story: The year is'' \\
5 & ``Write a yearly summary, log the current year at the top. Year:'' \\
6 & ``Write a story, starting with the current year. Year:'' \\
7 & ``Complete the sentence: The year is'' \\
8 & ``What day of the week is New Years this year? For the year'' \\
9 & ``Write a short summary, log the current year at the top. Year:'' \\
10 & ``Write a story. Story: The year is'' \\
\midrule
\multicolumn{2}{c}{\textbf{Instruction}} \\
\midrule
11 & ``Answer the question: What is the current year (YYYY)?'' \\
12 & ``What year is it, and what day of the week is Valentines day this year?'' \\
13 & ``As of today, what is the year?'' \\
14 & ``Write a story about the current year, starting with the current year.'' \\
15 & ``Write a yearly summary, log the current year at the top.'' \\
16 & ``Write a story, starting with the current year.'' \\
17 & ``Respond to the prompt: What's the year?'' \\
18 & ``What year is it, and what day of the week is New Years this year?'' \\
19 & ``Write a short summary, log the current year at the top.'' \\
20 & ``Write a story, where the story starts with the current year.'' \\
\bottomrule
\end{tabular}
\caption{\textbf{Declarative prompts}. Prompts grouped by continuation and instruction. These prompts were used in Section~\ref{section:declarative}.}
\label{tab:declarative_prompts}
\end{table}

%% file: figures/in_year_td_counts.tex
\begin{figure}[!htbp]
    \centering
    \begin{subfigure}[b]{0.7\linewidth}
        \centering
        \includegraphics[width=\linewidth]{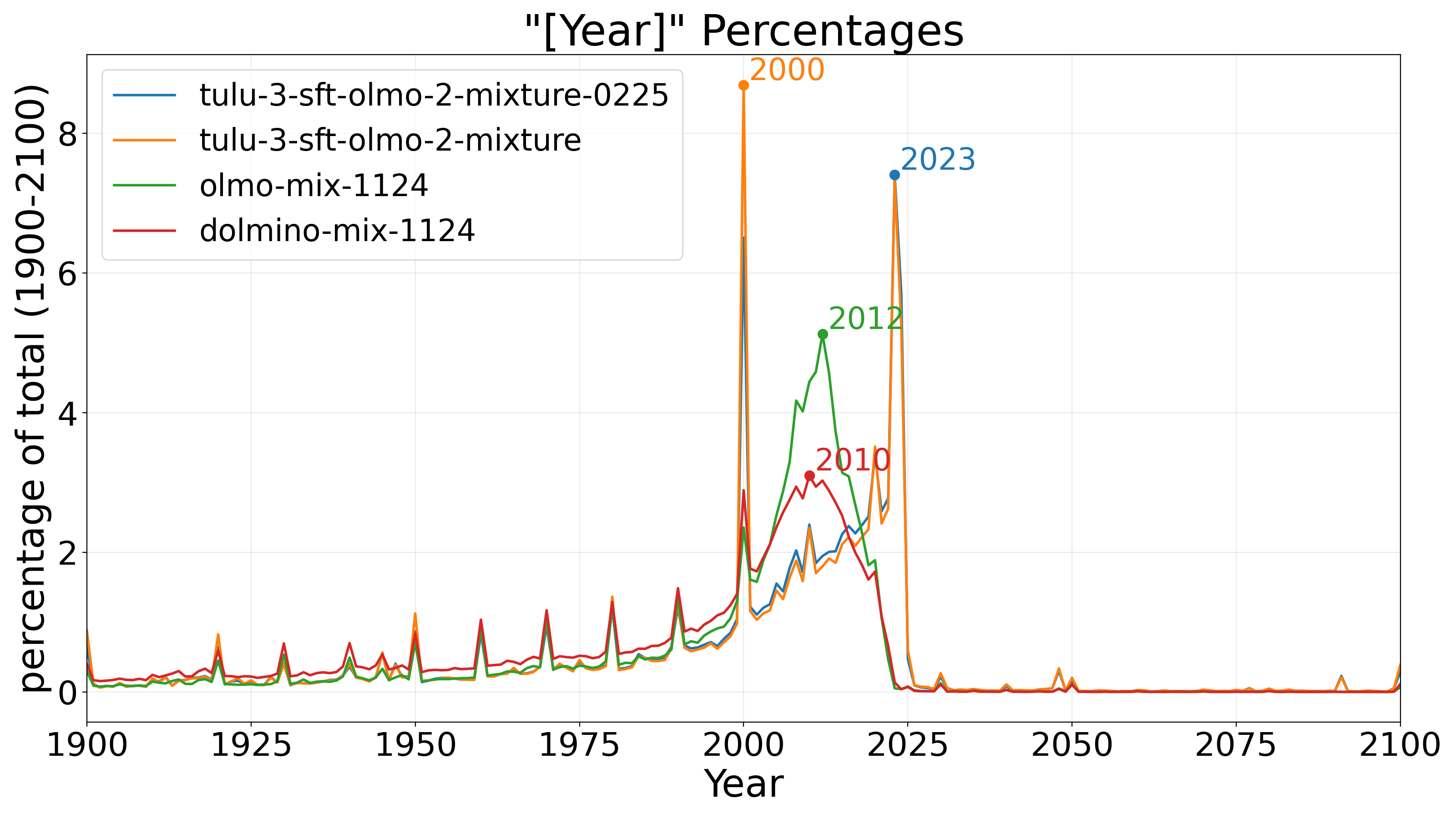}
        \caption{``[year]" counts (\%)}
    \end{subfigure}
    \hfill
    \begin{subfigure}[b]{0.7\linewidth}
        \centering
        \includegraphics[width=\linewidth]{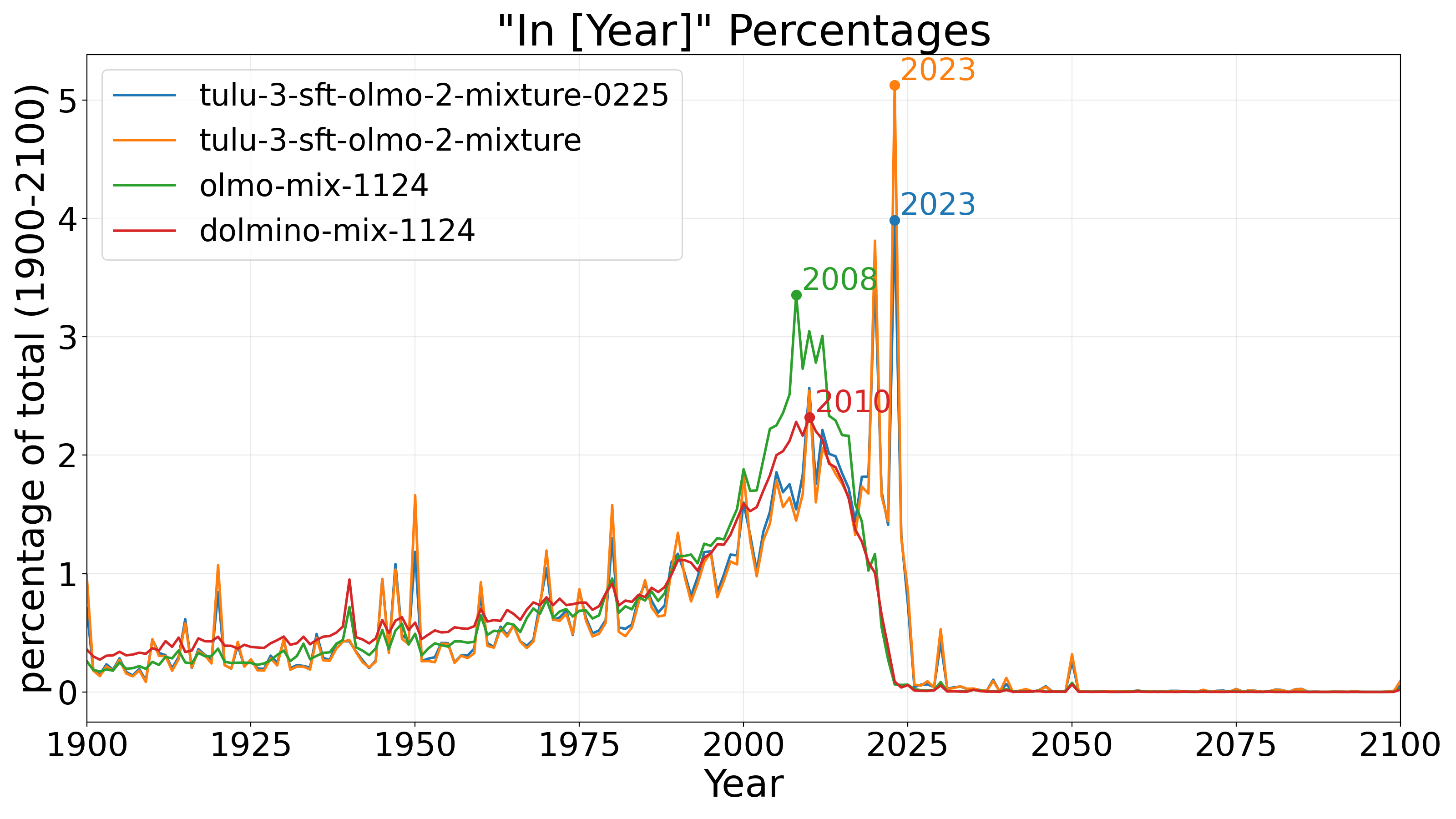}
        \caption{``In [year]" counts (\%)}
    \end{subfigure}

    \caption{\textbf{Year counts for training datasets olmo-mix, dolmino, and tulu3.} Normalized year frequency in pre-training (olmo-mix, dolmino-mix) and SFT (tulu3) datasets, counted as raw 4-digit occurrences (top) and in the \textit{In [year]} context (bottom). The \textit{In [year]} counts are used as the primary measure since they reflect temporal usage better than arbitrary 4-digit numbers.}
    \label{fig:combined_year_counts}
\end{figure}

%% file: figures/causal_tracing_declarative_averages.tex
\begin{figure}[!htbp]
    \centering

    \begin{subfigure}[b]{0.95\linewidth}
        \centering
        \includegraphics[width=\linewidth]{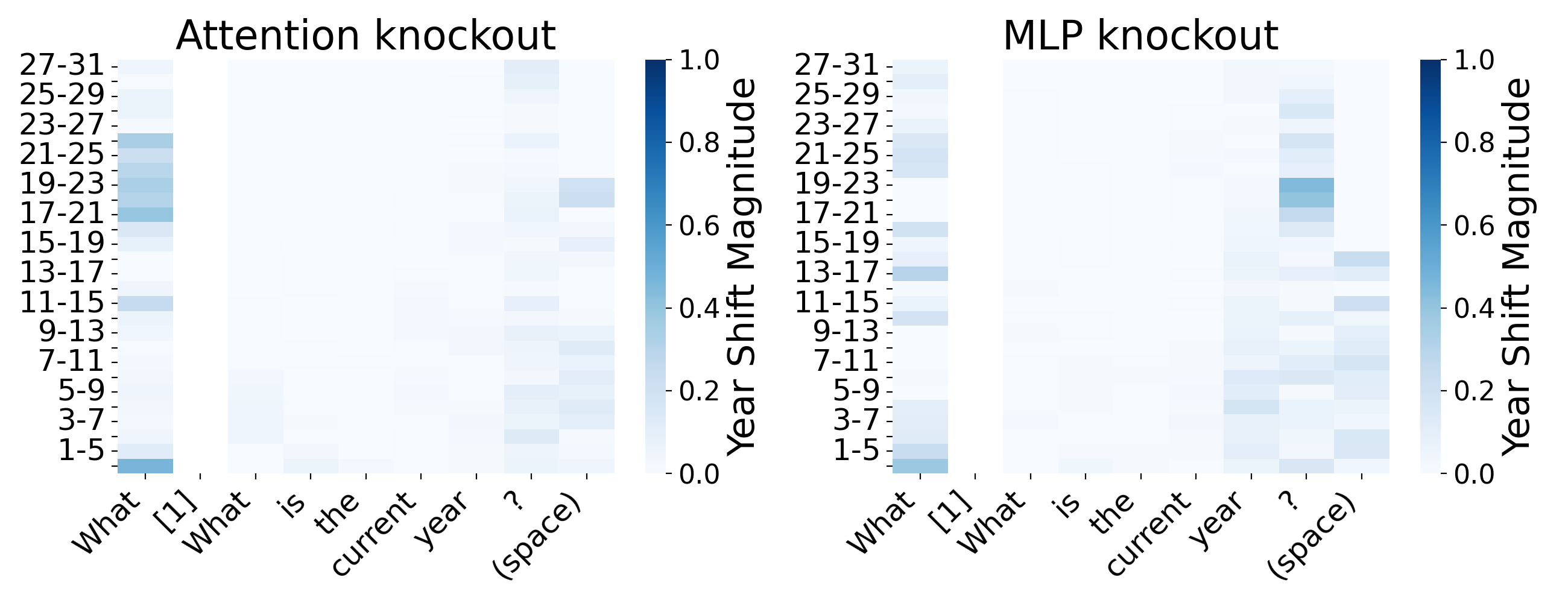}
        \caption{Mean. [1] is \textit{is the current month?[month]. What is the current day?[day].}}
    \end{subfigure}%
    \hfill
    \begin{subfigure}[b]{0.95\linewidth}
        \centering
        \includegraphics[width=\linewidth]{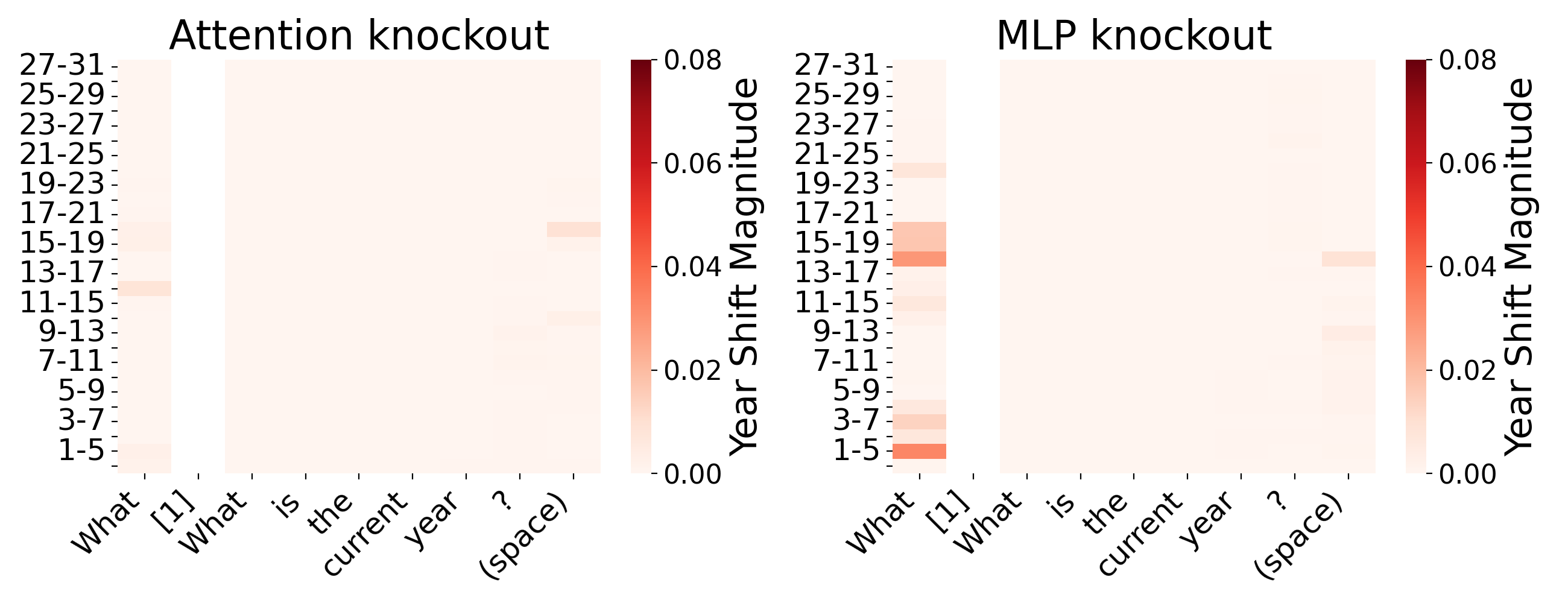}
        \caption{Variance. [1] is \textit{is the current month?[month]. What is the current day?[day].}}
    \end{subfigure}

\caption{\textbf{Mean and variance of causal tracing magnitude shift for declarative prompts} \ldots (continued)}
    \label{fig:mech_analysis_declarative_mean_var}
\end{figure}

\begin{figure}[!htbp]
    \ContinuedFloat
    \centering

    \begin{subfigure}[b]{0.95\linewidth}
        \centering
        \includegraphics[width=\linewidth]{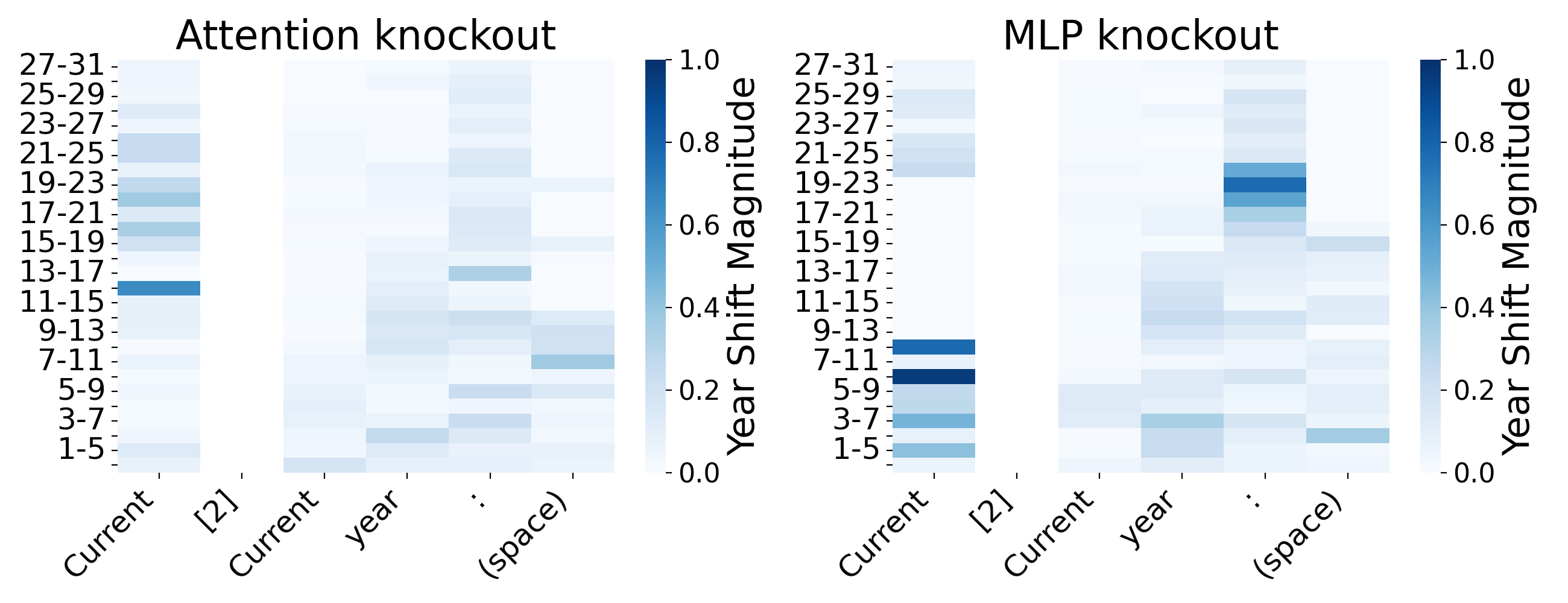}
        \caption{Mean. [2] is \textit{month:[month]. Current day:[day].}}
    \end{subfigure}
    \hfill
    \begin{subfigure}[b]{0.95\linewidth}
        \centering
        \includegraphics[width=\linewidth]{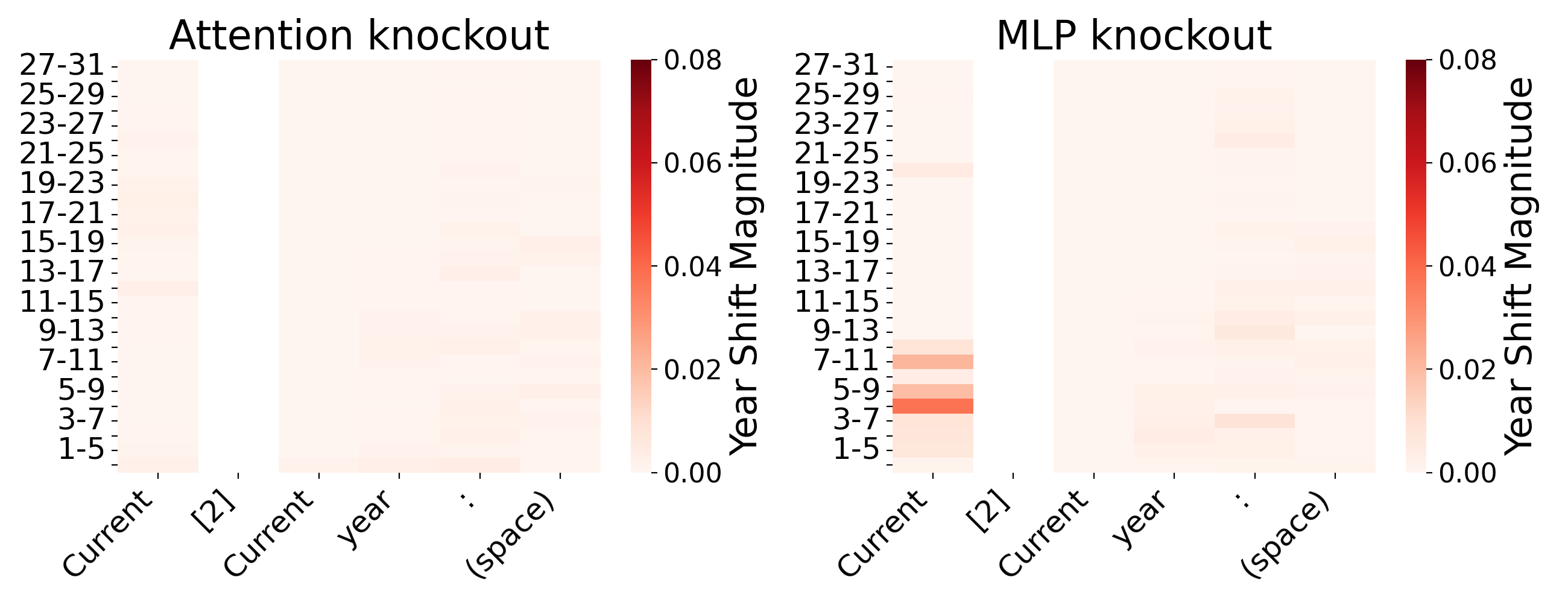}
        \caption{Variance. [2] is \textit{month:[month]. Current day:[day].}}
    \end{subfigure}

\caption{\textbf{Mean and variance of causal tracing magnitude shift for declarative prompts} \ldots (continued)}
\end{figure}

\begin{figure}[!htbp]
    \ContinuedFloat
    \centering

    \begin{subfigure}[b]{0.95\linewidth}
        \centering
        \includegraphics[width=\linewidth]{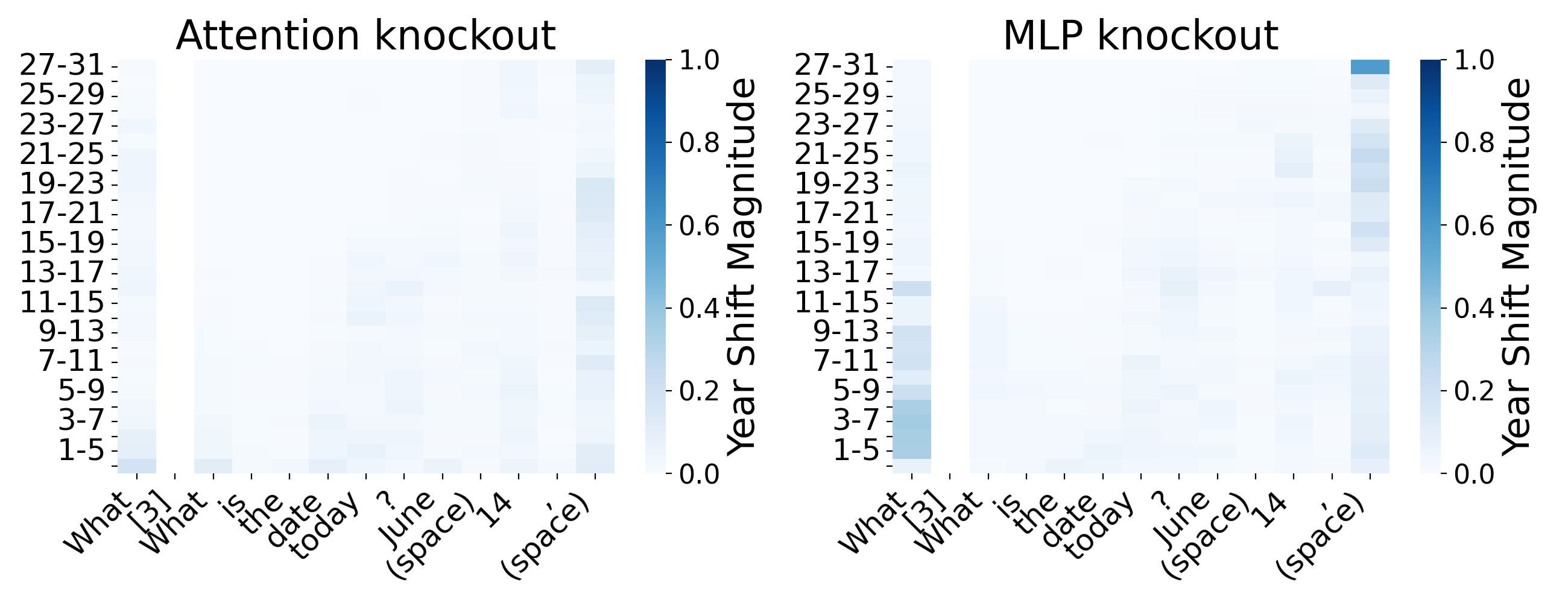}
        \caption{Mean. [3] is \textit{is the date[event1]?[date1]. What is the date[event2]?[date2].}}
    \end{subfigure}%
    \hfill
    \begin{subfigure}[b]{0.95\linewidth}
        \centering
        \includegraphics[width=\linewidth]{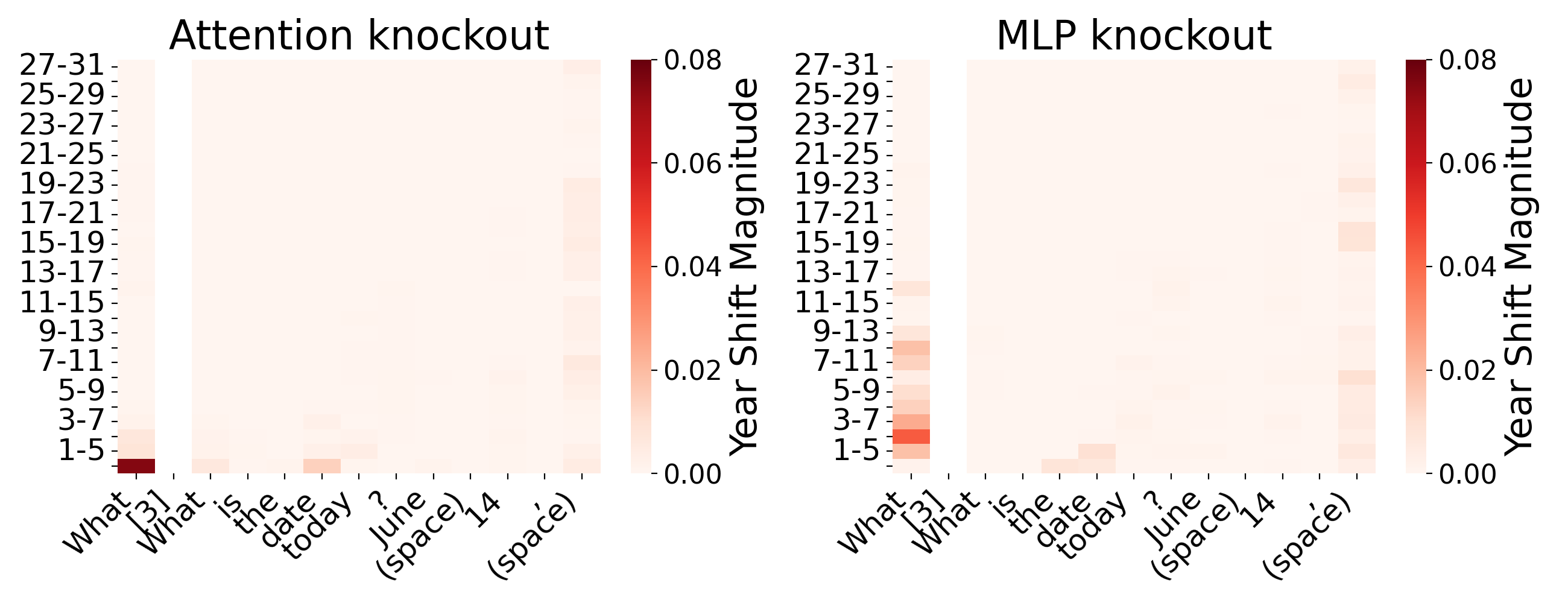}
        \caption{Variance. [3] is \textit{is the date[event1]?[date1]. What is the date[event2]?[date2].}}
    \end{subfigure}

\caption{\textbf{Mean and variance of causal tracing magnitude shift for declarative prompts} \ldots (continued)}

\end{figure}

\begin{figure}[!htbp]
    \ContinuedFloat
    \centering

    \begin{subfigure}[b]{0.95\linewidth}
        \centering
        \includegraphics[width=\linewidth]{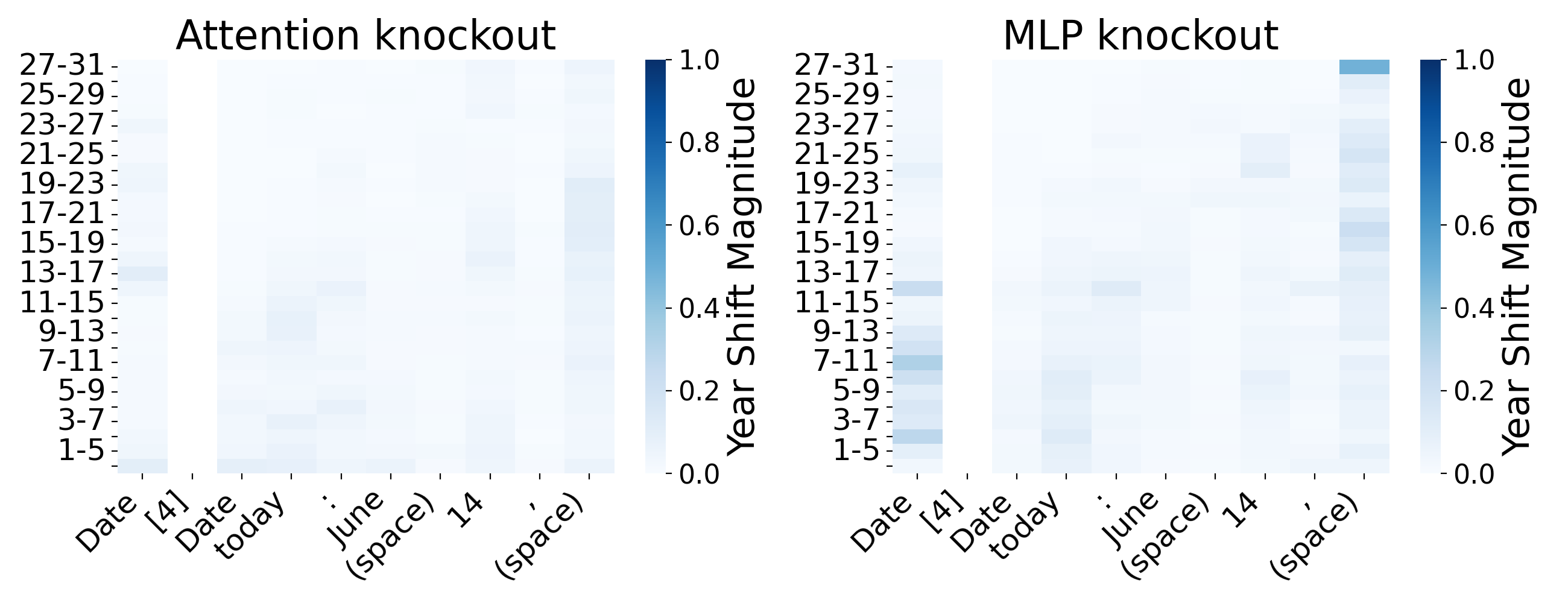}
        \caption{Mean. [4] is \textit{[event1]:[date1]. Date[event2]:[date2]}}
    \end{subfigure}%
    \hfill
    \begin{subfigure}[b]{0.95\linewidth}
        \centering
        \includegraphics[width=\linewidth]{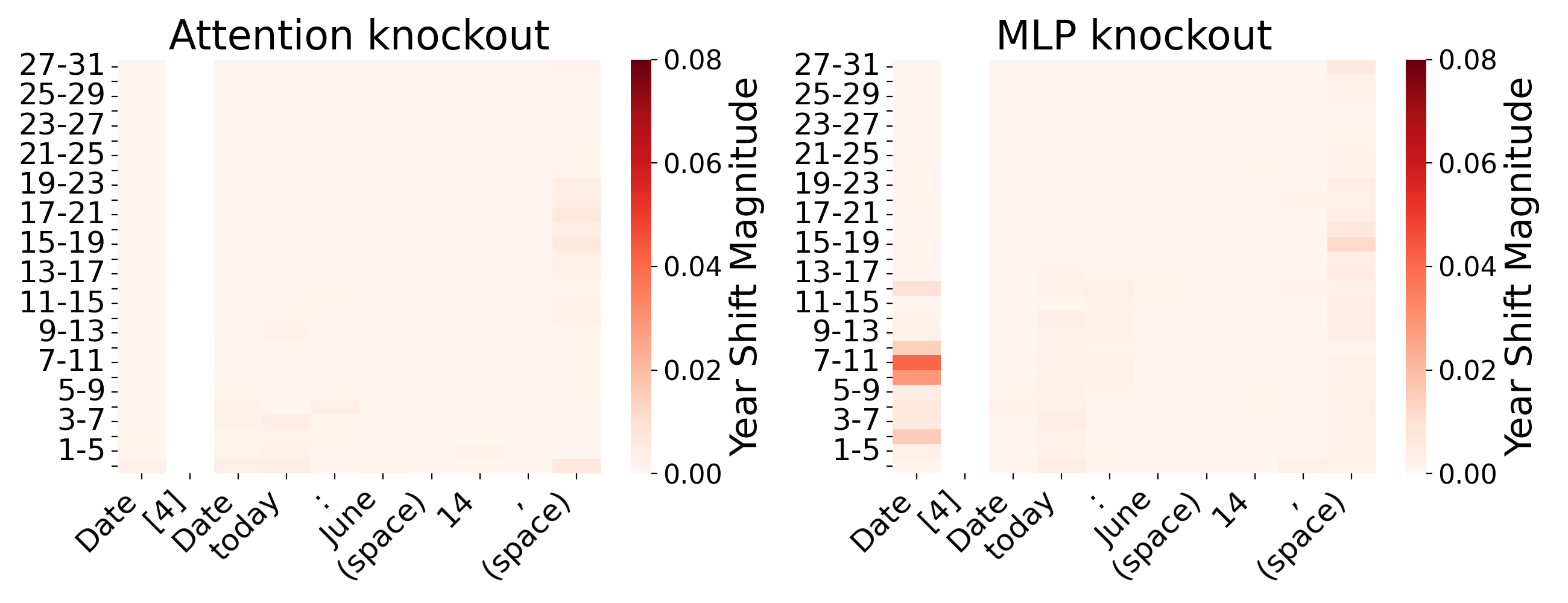}
        \caption{Variance. [4] is \textit{[event1]:[date1]. Date[event2]:[date2]}}
    \end{subfigure}

\caption{\textbf{Mean and variance of causal tracing magnitude shift for declarative prompts}, shown across four prompt types ([1]–[4]) in question and continuation formats. Unlike the associative task, there is no consistent layer or token position across prompt types, reflecting the lack of a stable causal pathway for the declarative current year. [1], [2], [3], and [4] are used to refer to the varying few-shot examples in the prompt, which differ in token length and therefore cannot be shown in the plots.}

\end{figure}

%% file: tables/declr_prompts_structured.tex
\begin{table}[!htbp]
\centering
\small
\setlength{\tabcolsep}{2pt} 

\begin{subtable}[t]{\linewidth}
\centering
\begin{tabular}{c p{0.7\linewidth}}
\toprule
\textbf{\#} & \textbf{Prompt} \\
\midrule
\multicolumn{2}{l}{\textbf{current\_year\_question}} \\
\midrule
1  & ``What is the current month? May. What is the current day? Sunday. What is the current year?'' \\
2  & ``What is the current month? October. What is the current day? Monday. What is the current year?'' \\
3  & ``What is the current month? January. What is the current day? Friday. What is the current year?'' \\
4  & ``What is the current month? February. What is the current day? Tuesday. What is the current year?'' \\
5  & ``What is the current month? March. What is the current day? Thursday. What is the current year?'' \\
6  & ``What is the current month? April. What is the current day? Saturday. What is the current year?'' \\
7  & ``What is the current month? June. What is the current day? Wednesday. What is the current year?'' \\
8  & ``What is the current month? July. What is the current day? Monday. What is the current year?'' \\
9  & ``What is the current month? August. What is the current day? Sunday. What is the current year?'' \\
10 & ``What is the current month? December. What is the current day? Tuesday. What is the current year?'' \\
\bottomrule
\end{tabular}
\end{subtable}

\vspace{0.2em}

\begin{subtable}[t]{\linewidth}
\centering
\begin{tabular}{c p{0.7\linewidth}}
\toprule
\textbf{\#} & \textbf{Prompt} \\
\midrule
\multicolumn{2}{l}{\textbf{current\_year\_stated}} \\
\midrule
1  & ``Current month: May. Current day: Sunday. Current year:'' \\
2  & ``Current month: October. Current day: Monday. Current year:'' \\
3  & ``Current month: January. Current day: Friday. Current year:'' \\
4  & ``Current month: February. Current day: Tuesday. Current year:'' \\
5  & ``Current month: March. Current day: Thursday. Current year:'' \\
6  & ``Current month: April. Current day: Saturday. Current year:'' \\
7  & ``Current month: June. Current day: Wednesday. Current year:'' \\
8  & ``Current month: July. Current day: Monday. Current year:'' \\
9  & ``Current month: August. Current day: Sunday. Current year:'' \\
10 & ``Current month: December. Current day: Tuesday. Current year:'' \\
\bottomrule
\end{tabular}
\end{subtable}

\caption{\textbf{Declarative prompts used for current-year tasks.} We use few-shot prompts to elicit a current year with both a question and continuation format. Month and day vary across prompts to prevent the LM from pattern-matching on a fixed date.}
\label{tab:declr_prompts_structured_current_year}
\end{table}

\begin{table}[!htbp]
\centering
\small
\renewcommand{\arraystretch}{0.92}
\setlength{\tabcolsep}{2pt} 

\begin{subtable}[t]{\linewidth}
\centering
\begin{tabular}{c p{0.7\linewidth}}
\toprule
\textbf{\#} & \textbf{Prompt} \\
\midrule
\multicolumn{2}{l}{\textbf{today\_question}} \\
\midrule
1  & ``What is the date the Apollo 11 Moon landing occurred? July 20, 1969. What is the date the iPhone was announced? January 9, 2007. What is the date today? June 14, '' \\
2  & ``What is the date the Berlin Wall fell? November 9, 1989. What is the date the September 11 attacks occurred? September 11, 2001. What is the date today? June 14, '' \\
3  & ``What is the date the Titanic sank? April 15, 1912. What is the date the Wright brothers made the first powered flight? December 17, 1903. What is the date today? June 14, '' \\
4  & ``What is the date the Sputnik 1 launch occurred? October 4, 1957. What is the date the World Wide Web was publicly announced at CERN? August 6, 1991. What is the date today? June 14, '' \\
5  & ``What is the date the U.S. Declaration of Independence was adopted? July 4, 1776. What is the date the U.S. Constitution was signed? September 17, 1787. What is the date today? June 14, '' \\
6  & ``What is the date the 'I Have a Dream' speech was delivered? August 28, 1963. What is the date the Nelson Mandela release from prison occurred? February 11, 1990. What is the date today? June 14, '' \\
7  & ``What is the date the attack on Pearl Harbor occurred? December 7, 1941. What is the date the first human spaceflight occurred? April 12, 1961. What is the date today? June 14, '' \\
8  & ``What is the date the Beatles appeared on The Ed Sullivan Show? February 9, 1964. What is the date the Woodstock festival began? August 15, 1969. What is the date today? June 14, '' \\
9  & ``What is the date the 19th Amendment was ratified? August 18, 1920. What is the date the Nixon resignation occurred? August 8, 1974. What is the date today? June 14, '' \\
10 & ``What is the date the start of World War II occurred? September 1, 1939. What is the date the end of World War II occurred? September 2, 1945. What is the date today? June 14, '' \\
\bottomrule
\end{tabular}
\end{subtable}

\vspace{0.2em}

\begin{subtable}[t]{\linewidth}
\centering
\begin{tabular}{c p{0.7\linewidth}}
\toprule
\textbf{\#} & \textbf{Prompt} \\
\midrule
\multicolumn{2}{l}{\textbf{today\_stated}} \\
\midrule
1  & ``Date the Apollo 11 Moon landing occurred: July 20, 1969. Date the iPhone was announced: January 9, 2007. Date today: June 14, '' \\
2  & ``Date the Berlin Wall fell: November 9, 1989. Date the September 11 attacks occurred: September 11, 2001. Date today: June 14, '' \\
3  & ``Date the Titanic sank: April 15, 1912. Date the Wright brothers made the first powered flight: December 17, 1903. Date today: June 14, '' \\
4  & ``Date Sputnik 1 launched: October 4, 1957. Date the World Wide Web was publicly announced at CERN: August 6, 1991. Date today: June 14, '' \\
5  & ``Date the U.S. Declaration of Independence was adopted: July 4, 1776. Date the U.S. Constitution was signed: September 17, 1787. Date today: June 14, '' \\
6  & ``Date the 'I Have a Dream' speech was delivered: August 28, 1963. Date Nelson Mandela was released from prison: February 11, 1990. Date today: June 14, '' \\
7  & ``Date the attack on Pearl Harbor occurred: December 7, 1941. Date the first human went to space: April 12, 1961. Date today: June 14, '' \\
8  & ``Date the Beatles appeared on The Ed Sullivan Show: February 9, 1964. Date Woodstock began: August 15, 1969. Date today: June 14, '' \\
9  & ``Date women's suffrage in the U.S. was ratified: August 18, 1920. Date Watergate ended with Nixon's resignation: August 8, 1974. Date today: June 14, '' \\
10 & ``Date the start of World War II occurred: September 1, 1939. Date World War II ended: September 2, 1945. Date today: June 14, '' \\
\bottomrule
\end{tabular}
\end{subtable}

\caption{\textbf{Declarative prompts used for today-date tasks.} We use few-shot prompts to elicit a current year with both a question and continuation format. We fix the month and day as June 14 and vary the historical examples.}
\label{tab:declr_prompts_structured_today_date}
\end{table}

%% file: figures/7b_training_dynamics.tex
\begin{figure}[!htbp]
    \centering
    \includegraphics[width=0.9\linewidth]{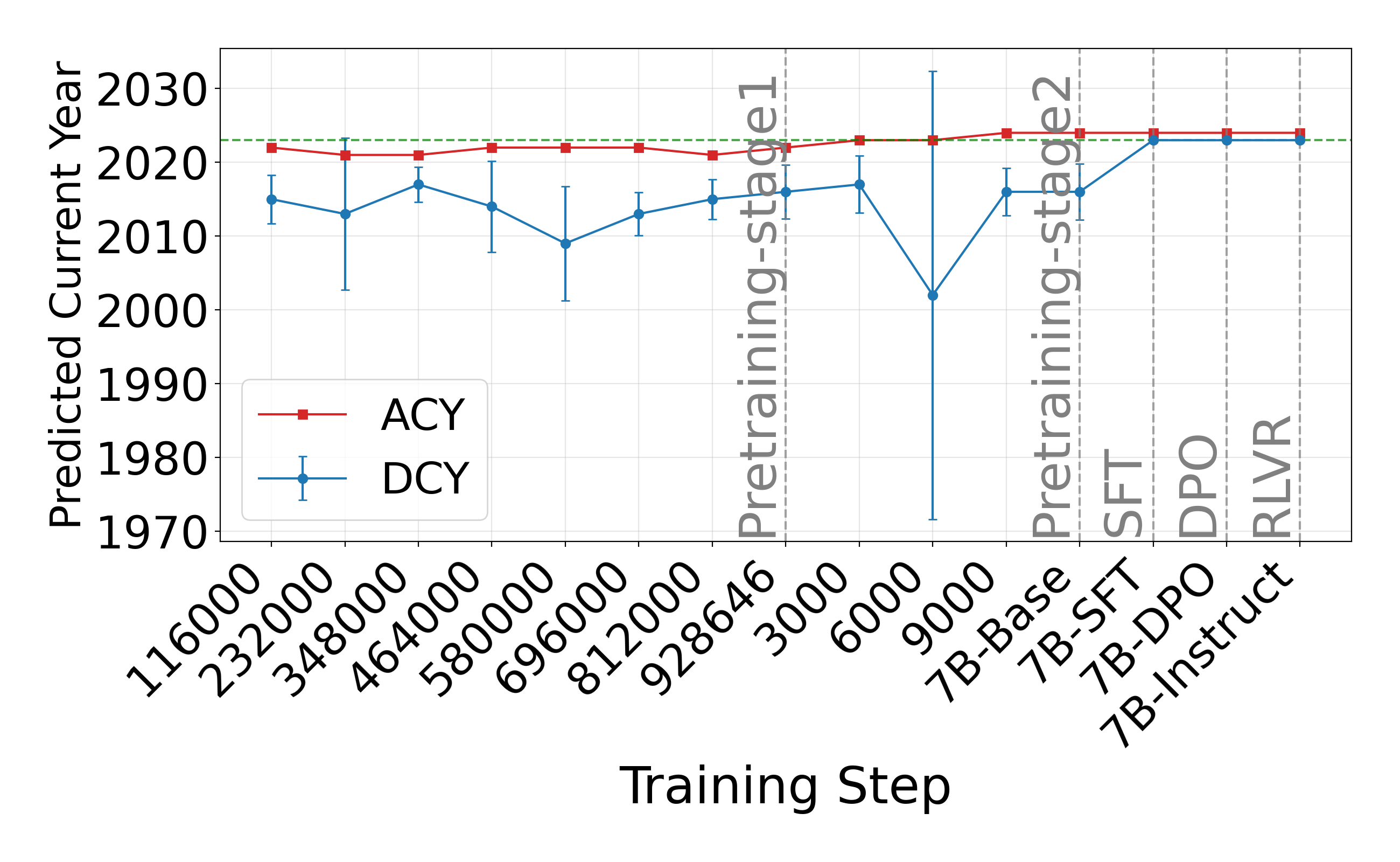}
    \caption{\textbf{Training dynamics across training checkpoints for \texttt{OLMo2-7B}.} Points represent the declarative year, and bars represent the standard deviation. The horizontal line indicates the data cutoff year. This is the \texttt{OLMo2-7B} counterpart of Figure~\ref{fig:training_dynamics_1b}.}
    \label{fig:training_dynamics_7b}
\end{figure}